\documentclass{article} 
\usepackage{collas2024_conference,times}
\usepackage{easyReview}

\usepackage{hyperref}
\hypersetup{
    colorlinks=true,
    linkcolor=red,
    filecolor=magenta,
    urlcolor=blue,
    citecolor=purple,
    pdftitle={Overleaf Example},
    pdfpagemode=FullScreen,
    }

\usepackage{graphicx}
\usepackage{amsmath}
\usepackage{amssymb}
\usepackage{mathtools}
\usepackage{amsthm}
\usepackage{algorithm} 
\usepackage{algorithmic}
\usepackage{color}
\usepackage{caption,subcaption}
\usepackage{bbm}
\usepackage{mdframed}
\usepackage{array}
\usepackage{tikz}
\usepackage{booktabs}
\usepackage{arydshln}
\usepackage{xcolor}
\usepackage{xspace}

\newcommand{\rebuttal}[1]{#1}

\title{Maintaining Plasticity in Continual Learning via Regenerative Regularization}

\author{Saurabh Kumar\thanks{Denotes equal contribution} \\
Department of Computer Science\\
Stanford University \\
\texttt{\{szk\}@stanford.edu} \\
\And 
Henrik Marklund$^*$  \\
Department of Computer Science \\
Stanford University \\
\texttt{\{marklund\}@stanford.edu} \\
\AND 
Benjamin Van Roy \\
Department of Electrical Engineering \\
Department of Management Science \& Engineering \\
Stanford University \\
\texttt{\{bvr\}@stanford.edu}
}

\collasfinalcopy 

\definecolor{Baseline}{HTML}{0072B2}           
\definecolor{L2Init}{HTML}{000000}             
\definecolor{L2}{HTML}{E69F00}                
\definecolor{LayerNorm}{HTML}{808080}         
\definecolor{ContinualBackprop}{HTML}{56B4E9} 
\definecolor{ShrinkAndPerturb}{HTML}{CC79A7}
\definecolor{ConcatReLU}{HTML}{7FC97F}
\definecolor{ReDO}{HTML}{D55E00}

\definecolor{L1Init}{HTML}{E69F00}  
\definecolor{L2InitResample}{HTML}{CC79A7} 
\definecolor{HuberInit}{HTML}{2CA02C}

\definecolor{lambda_zero}{HTML}{0173B2}
\definecolor{lambda_1e-5}{HTML}{DE8F05}
\definecolor{lambda_1e-4}{HTML}{029E73}
\definecolor{lambda_1e-3}{HTML}{D55E00}
\definecolor{lambda_1e-2}{HTML}{CC78BC}
\definecolor{lambda_1e-1}{HTML}{666666}

\definecolor{PyTorch_Default}{HTML}{0173B2}
\definecolor{Kaiming_Normal}{HTML}{DE8F05}
\definecolor{Kaiming_Uniform}{HTML}{029E73}
\definecolor{Xavier_Normal}{HTML}{D55E00}
\definecolor{Xavier_Uniform}{HTML}{CC78BC}

\newcommand{\methodname}{\text{L$2$ Init}\xspace}
\newcommand{\task}{{\rm{T}}}

\begin{document}

\maketitle

\begin{abstract}
In continual learning, plasticity refers to the ability of an agent to quickly adapt to new information. Neural networks are known to lose plasticity when processing non-stationary data streams. In this paper, we propose \textit{\methodname}, a simple approach for maintaining plasticity by incorporating in the loss function L2 regularization toward initial parameters.  This is very similar to standard L$2$ regularization (L$2$), the only difference being that L$2$ regularizes toward the origin.  \methodname is simple to implement and requires selecting only a single hyper-parameter. The motivation for this method is the same as that of methods that reset neurons or parameter values. Intuitively, when recent losses are insensitive to particular parameters, these parameters should drift toward their initial values. This prepares parameters to adapt quickly to new tasks. On problems representative of different types of nonstationarity in continual supervised learning, we demonstrate that \methodname most consistently mitigates plasticity loss compared to previously proposed approaches. 
\end{abstract}

\section{Introduction}

In continual learning, an agent must continually adapt to an ever-changing data stream. Previous studies have shown that in non-stationary problems, neural networks tend to lose their ability to adapt over time (see e.g., \citet{achille2017critical, ash2020warm, dohare2021continual}). This is known as \textit{loss of plasticity}. Methods proposed to mitigate this issue include those which continuously or periodically reset some subset of weights~\citep{dohare2021continual,sokar2023dormant},
add regularization to the training objective~\citep{ash2020warm}, or add architectural changes to the neural network~\citep{ba2016layer,lyle2023understanding,nikishin2023deep}. 

However, these approaches either fail on a broader set of problems or can be quite complicated to implement, with multiple moving parts or hyper-parameters to tune. In this paper, we draw inspiration from methods that effectively maintain plasticity in continual learning, such as Continual Backprop~\citep{dohare2021continual}, to propose a simpler regularization-based alternative. Our main contribution is a simple approach for maintaining plasticity that we call \textit{\methodname}. Our approach manifests as a simple modification to L2 regularization which is used throughout the deep learning literature. Rather than regularizing toward zero, \methodname regularizes toward the initial parameter values. Specifically, our proposed regularization term is the squared L2 norm of the difference between the network's current parameter values and the initial values. \methodname is a simple method to implement that only requires one additional hyper-parameter.

The motivation for this approach is the same as that of methods that reset neurons or parameters, such as Continual Backprop. Intuitively, by ensuring that some parameter values are close to initialization, there are always parameters that can be recruited for rapid adaption to a new task. There are multiple reasons why having parameters close to initialization may increase plasticity, including maintaining smaller weight magnitudes, avoiding dead ReLU units, and preventing weight rank from collapsing. 

To study \methodname, we perform an empirical study on continual supervised learning problems, each exhibiting one of two types of non-stationarity: input distribution shift and target function (or concept) shift. We find that \methodname most consistently retains high plasticity on both types of non-stationarity relative to other methods. To better understand the mechanism by which \methodname maintains plasticity, we study how the average weight magnitude and feature rank evolve throughout training. While both \methodname and standard L$2$ regularization reduce weight magnitude, \methodname maintains high feature rank, a property that is sometimes correlated with retaining plasticity~\citep{kumar2020implicit}. Finally, in an ablation, we find that regularizing toward the fixed initial parameters rather than a random set of parameters is an important component of the method. Further, we find that using the L1 distance instead of L2 distance when regularizing towards initial parameters also significantly mitigates plasticity loss, but overall performance is slightly worse compared to L2 Init.

\section{Related Work}

Over the past decade, there has been emerging evidence that neural networks lose their capacity to learn over time  when faced with nonstationary data streams~\citep{ash2020warm, dohare2021continual}. This phenomenon was first identified for deep learning in the context of pre-training~\citep{achille2017critical,zilly2020negative, ash2020warm}. 
For instance, \citet{achille2017critical} demonstrated that training a neural network on blurred CIFAR images significantly reduced its ability to subsequently learn on the original CIFAR images. Since then, the deterioration of neural networks' learning capacity over time has been identified under various names, including the negative pre-training effect \citep{zilly2020negative}, intransigence \citep{chaudhry2018riemannian}, critical learning periods \citep{achille2017critical}, the primacy bias \citep{nikishin2022primacy}, dormant neuron phenomenon \citep{sokar2023dormant}, implicit under-parameterization \citep{kumar2020implicit}, capacity loss \citep{lyle2022understanding}, and finally, the all-encompassing term, loss of plasticity (or plasticity loss) \citep{lyle2023understanding}. In this section, we review problem settings in which plasticity loss has been studied, potential causes of plasticity loss, and methods previously proposed to mitigate this issue. 

\subsection{Problem Settings}
We first review two problem settings in which plasticity loss has been studied: continual learning and reinforcement learning.

\textbf{Continual Learning.} In this paper, we aim to mitigate plasticity loss in the continual learning setting, and in particular, continual supervised learning. While the continual learning literature has primarily focused on reducing catastrophic forgetting~\citep{goodfellow2013empirical, kirkpatrick2017overcoming}, more recently, the issue of plasticity loss has gained significant attention~\citep{dohare2021continual, dohare2023maintaining, abbas2023loss}. \citet{dohare2021continual} demonstrated that loss of plasticity sometimes becomes evident only after training for long sequences of tasks. 
Therefore, in continual learning,
mitigating plasticity loss becomes especially important as agents encounter many tasks, or more generally a non-stationary data stream, over a long lifetime.

\textbf{Reinforcement Learning.} Plasticity loss has also gained significant attention in the deep reinforcement learning (RL) literature~\citep{igl2020transient,kumar2020implicit,nikishin2022primacy,lyle2022understanding,gulcehre2022empirical,sokar2023dormant,nikishin2023deep, lyle2023understanding}. In RL, the input data stream exhibits two sources of non-stationarity. First, observations are significantly correlated over time and are influenced by the agent's policy which is continuously evolving. Second, common RL methods using temporal difference learning bootstrap off of the predictions of a periodically updating target network~\citep{mnih2013playing}. The changing regression target introduces an additional source of non-stationarity.

\subsection{Causes of plasticity loss}

While there are several hypotheses for why neural networks lose plasticity, this issue remains poorly understood. Proposed causes include inactive ReLU units, feature or weight rank collapse, and divergence due to large weight magnitudes~\citep{lyle2023understanding, sokar2023dormant, dohare2023maintaining, kumar2020implicit}. \citet{dohare2021continual} suggest that using the Adam optimizer makes it difficult to update weights with large magnitude since updates are bounded by the step size. \citet{zilly2021plasticity} propose that when both the incoming and outgoing weights of a neuron are close to zero, they are ``mutually frozen" and will be very slow to update, which can result in reduced plasticity. 
However, both \citet{lyle2023understanding} and \citet{gulcehre2022empirical} show that many of the previously suggested mechanisms for loss of plasticity are insufficient to explain plasticity loss. While the causes of plasticity loss remain unclear, we believe it is possible to devise methods to mitigate the issue, drawing inspiration from the fact that initialized neural networks have high plasticity.

\subsection{Mitigating plasticity loss}
There have been about a dozen methods proposed for mitigating loss of plasticity. We categorize them into four main types: resetting, regularization, architectural, and optimizer solutions. 

\textbf{Resetting.} This paper draws inspiration from resetting methods, which reinitialize subsets of neurons or parameters~\citep{zilly2020negative,dohare2021continual,nikishin2022primacy,nikishin2023deep,sokar2023dormant, dohare2023maintaining}. For instance, Continual Backprop \citep{dohare2021continual} tracks a utility measure for each neuron, \rebuttal{ranks neurons based on utility and resets the $k$ lowest utility neurons. The value of $k$ is determined by combination of a hyper-paramter called replacement rate and how recently the neuron was reset.} This procedure involves multiple hyper-parameters, including the maturity threshold, the replacement rate, and the utility decay rate. \citet{sokar2023dormant} propose a similar but simpler idea. Instead of tracking utilities for each neuron, they periodically compute the activations on a batch of data. 
A neuron is reset if it has small average activation relative to other neurons in the corresponding layer of the neural network. A related solution to resetting individual neurons is to keep a replay buffer and train a newly initialized neural network from scratch on data in the buffer~\citep{igl2020transient}, either using the original labels or using the current network's outputs as targets. This is a conceptually simple but computationally very expensive method. Inspired by these approaches, the aim of this paper is to develop a simple regularization method that implicitly, and smoothly, resets weights with low utility.

\textbf{Regularization.} A number of methods have been proposed that regularize neural network parameters~\citep{ash2020warm,kumar2020implicit,lyle2022understanding}. The most similar approach to our method is L2 regularization, which regularizes parameters towards zero. While L2 regularization reduces parameter magnitudes which helps mitigate plasiticy loss, regularizing toward the origin is likely to collapse the ranks of the weight matrices as well as lead to so-called mutually frozen weights \citep{zilly2021plasticity}, both of which may have adverse effects on plasticity. In contrast, our regenerative regularization approach avoids these issues. Another method similar to ours is Shrink \& Perturb \citep{ash2020warm} which is a two-step procedure applied at regular intervals. The weights are first shrunk by multiplying with a scalar and then perturbed by adding random noise. The shrinkage and noise scale factors are hyper-parameters. %
In Appendix~\ref{appendix:shrink-and-perturb-connection}, we discuss the relationship between Shrink \& Perturb and the regenerative regularization we propose. Additional regularization methods to mitigate plasticity loss include those proposed by \citet{lyle2022understanding}, which regularizes a neural network's output towards earlier predictions, and \citet{kumar2020implicit}, which maximizes feature rank. 

Lastly, we discuss Elastic Weight Consolidation (EWC) 
\citep{kirkpatrick2017overcoming} which was designed for mitigating catastrophic forgetting. EWC is similar our method in that it regularizes towards previous parameters. An important difference, however, is that EWC does not regularize towards initial parameters, but rather towards parameters at the end of each previous task. Thus, while EWC is designed to remember information about previous tasks, our method is designed to maintain plasticity. \rebuttal{In effect, our method could be considered as form of `remembering how to learn'.}

\textbf{Architectural.} %
Layer normalization \citep{ba2016layer}, which is a common technique used throughout deep learning, has been shown to mitigate plasticity loss~\citep{lyle2023understanding}. 
A second solution aims to reduce the number of neural network features which consistently output zero by modifying the ReLU activation function~\citep{shang2016understanding, abbas2023loss}. In particular, applying Concatenated ReLU ensures that each neuron is always activated and therefore has non-zero gradient. However, Concatenated ReLU comes at the cost of doubling the total number of parameters. In particular, each hidden layer output is concatenated with the negative of the output values before applying the ReLU activation, which doubles the number of inputs to the next layer. In our experiments in Section~\ref{sec:experiments}, we modify the neural network architecture of Concat ReLU such that it has the same parameter count as all other agents.

\textbf{Optimizer.} The Adam optimizer in its standard form is ill-suited for the continual learning setting. In particular, Adam tracks estimates of the first and second moments of the gradient, and these estimates can become inaccurate when the incoming data distribution changes rapidly. When training value-based RL agents, \citet{lyle2023understanding} evaluates the effects of resetting the optimizer state when the target network is updated. This alone did not mitigate plasticity loss. Another approach they evaluate is tuning Adam hyper-parameters such that second moment estimates are more rapidly updated and sensitivity to large gradients is reduced. While this significantly improved performance on toy RL problems, some plasticity loss remained. An important benefit of the method we propose is that it is designed to work with any neural network architecture and optimizer.

\section{Regenerative Regularization}\label{sec:method}

In this section, we propose a simple method for maintaining plasticity, which we call \methodname. Our approach draws inspiration from prior works which demonstrate the benefits of selectively reinitializing parameters for retaining plasticity. The motivation for these approaches is that reinitialized parameters can be recruited for new tasks, and dormant or inactive neurons can regain their utility~\citep{dohare2021continual,nikishin2022primacy,sokar2023dormant}. While these methods have enjoyed success across different problems, they often involve multiple additional components or hyper-parameters. In contrast, \methodname is simple to implement and introduces a single hyper-parameter.

Given neural network parameters $\theta$, \methodname augments a standard training loss function $\mathcal{L}_{\text{train}}(\theta)$ with a regularization term. Specifically, \methodname performs L$2$ regularization toward initial parameter values $\theta_0$ at every time step for which a gradient update occurs. The augmented loss function is
\begin{align*}
    \mathcal{L}_{\text{reg}}(\theta) = \mathcal{L}_{\text{train}}(\theta) + \lambda ||\theta - \theta_0||_2^2,
\end{align*}
where $\lambda$ is the regularization strength and $\theta_0$ is the vector of parameter values at time step $0$.

Our regularization term is similar to standard L$2$ regularization, with the difference that \methodname regularizes toward the initial parameter values instead of the origin. While this is a simple modification, we demonstrate in Section \ref{sec:experiments} that it significantly reduces plasticity loss relative to standard L$2$ regularization in continual learning settings. 

\methodname is similar in spirit to resetting methods such as Continual Backprop~\citep{dohare2021continual}, which explicitly computes a utility measure for each neuron and then resets neurons with low utility. Rather than resetting full neurons, \methodname works on a per-weight basis, and encourages weights with low utility to reset. Intuitively, when the training loss $\mathcal{L}_{\text{train}}$ becomes insensitive to particular parameters, these parameters drift toward their initial values, preparing them to adapt quickly to future tasks. Thus, \methodname can be thought of as implicitly and smoothly resetting low-utility weights. We use the term \textit{regenerative regularization} to characterize regularization which rejuvenates parameters that are no longer useful.

\section{Continual Supervised Learning}\label{sec:continual_supervised_learning}

In this paper, we study plasticity loss in the continual supervised learning setting. In the continual supervised learning problems we consider, an agent is presented with a sequence $\{\task_i\}_{i=1}^K$ of $K$ tasks. Each task $\task_i$ corresponds to a unique dataset $\mathcal{D}_{\task_i}$ of (image, label) data pairs, and the agent receives a batch of samples from this dataset at each timestep, for a fixed duration of $M$ timesteps.

\subsection{Evaluation Protocol}\label{sec:eval_protocol}

To measure agents' performance as well as their ability to retain plasticity, we measure the average online accuracy on each task.
In particular, for each task $\task_i$, we compute 
$$\text{Avg Online Task Accuracy}(\task_i) = \frac{1}{M} \sum_{j = t_i}^{t_i + M - 1} a_j$$ 
where $t_i$ is the starting time step of task $T_i$ and $a_j$ is the average accuracy on the $j$th batch of samples. We refer to this metric as the \textit{average online task accuracy}. This metric captures how quickly the agent is able to learn to do well on the task, which is a measure of its plasticity. If average online task accuracy goes down over time, we say that there is plasticity loss, assuming all tasks are of equal difficulty. 

To perform model selection, we additionally compute each agent's average online accuracy over all data seen in the agent's lifetime. This is a common metric used in online continual learning~\citep{cai2021online,ghunaim2023real,prabhu2023online} and is computed as follows: 
$$\text{Total Avg Online Accuracy} = \frac{1}{MK} \sum_{t = 0}^{MK} a_t$$ 

To distinguish from average online task accuracy, we will refer to this metric as the \textit{total average online accuracy}.

Plasticity loss encapsulates two related but distinct phenomena. First, it encompasses the reduction in a neural network's capacity to fit incoming data. For instance, \citet{lyle2023understanding} show how a neural network trained using Adam optimizer significantly loses its ability to fit a dataset of MNIST images with randomly assigned labels. Second, plasticity loss also includes a reduction in a neural network's capacity to generalize to new data \citep{igl2020transient, liu2020bad}. \rebuttal{In environments where each data point is seen only once}, the two metrics above will be sensitive to both of these phenomena. \rebuttal{However, if data points are seen more than once, the metric will be less sensitive to generalization the more times each data points are seen.}

\subsection{Problems}\label{sec:problems}
In our experiments in Section~\ref{sec:experiments}, we evaluate methods on five continual image classification problems. Three of the problems, Permuted MNIST, 5+1 CIFAR, and Continual ImageNet exhibit input distribution shift, where different tasks have different inputs. The remaining problems, Random Label MNIST and Random Label CIFAR, exhibit concept shift, where different tasks have the exact same inputs but different labels assigned to each input. All continual image classification problems we consider consist of a sequence of supervised learning tasks. The agent is presented with batches of (image, label) data pairs from a task for a fixed number of timesteps, after which the next task arrives. The agent is trained incrementally to minimize cross-entropy loss on the batches it receives. While there are discrete task boundaries, the agent is not given any indication when a task switches. 

\textbf{Permuted MNIST.} The first problem we consider is Permuted MNIST, a common benchmark from the continual learning literature~\citep{goodfellow2013empirical}. In our Permuted MNIST setup, we randomly sample 10,000 images from the MNIST training dataset. A Permuted MNIST task is characterized by applying a fixed randomly sampled permutation to the input pixels of all 10,000 images. The agent is presented with these 10,000 images in a sequence of batches, equivalent to training for $1$ epoch through the task's dataset. After all samples have been seen once, the next task arrives, and the process repeats. In our Permuted MNIST experiments, we train agents for $500$ tasks.

\textbf{Random Label MNIST.} Our second problem is Random Label MNIST, a variation of the problem in~\citet{lyle2023understanding}. We randomly sample $1200$ images from the MNIST dataset. A Random Label MNIST task is characterized by randomly assigning a label to each individual image in this subset. In contrast to Permuted MNIST, we train the agent for $400$ epochs such that the neural network learns to memorize the labels for the images. After $400$ epochs are complete, the next task arrives, and the process repeats. We train agents for $50$ tasks. 

\textbf{Random Label CIFAR.} The third problem is Random Label CIFAR, which is equivalent to the setup of Random Label MNIST except that data is sampled from the CIFAR $10$ training dataset. For Permuted MNIST, Random Label MNIST, and Random Label CIFAR, data arrives in batches of size $16$.

\textbf{5+1 CIFAR.} In our fourth problem, 5+1 CIFAR, tasks have varying difficulty. Specifically, every even task is ``hard" while every odd task is ``easy." Data is drawn from the CIFAR 100 dataset, and a hard task is characterized by seeing (image, label) data pairs of $5$ CIFAR 100 classes, whereas in an easy task, data from from only a single class arrives. Each hard task consists of $2500$ data pairs ($500$ from each class), while each easy tasks consists of $500$ data pairs from a single class. In particular, the tasks which have a single class are characterized as ``easy" since all labels are the same. Each task has a duration of $780$ timesteps which corresponds to $10$ epochs through the hard task datasets when using a batch size of $32$. This problem is designed to reflect continual learning scenarios with varying input distributions, as agents receive data with varying levels of diversity at different times. In this problem, we measure agents' performance specifically on the hard tasks since all agents do well on the easy tasks. \rebuttal{Note that this is a highly synthetic environment designed to stress test methods which mitigate plasticity loss.}

\textbf{Continual ImageNet.} The fifth problem is a variation of Continual ImageNet \citep{dohare2023maintaining}, where each task is to distinguish between two ImageNet classes. Each task draws from a dataset of $1200$ images, $600$ from each of two classes. We train agents for $10$ epochs on each task using batch size $100$. In line with \citep{dohare2023maintaining}, the images are downsized to 32 x 32 to save computation. In both 5+1 CIFAR and Continual ImageNet, each individual class does not occur in more than one task. Additional details of all problems are in Appendix \ref{appendix:environments}.

\section{Experiments}\label{sec:experiments}
The goal of our experiments is to determine whether \methodname mitigates plasticity loss in continual supervised learning. To this end, we evaluate \methodname and a selection of prior approaches on  continual image classification problems introduced in Section~\ref{sec:problems}, most of which have been previously used to study plasticity loss~\cite{dohare2021continual,lyle2023understanding}. We select methods which have shown good performance in previous work studying continual learning and which are representative of three different method types: resetting, regularization, and architectural solutions. These methods we consider are the following:
\begin{itemize}
\item Resetting: Continual Backprop \citep{dohare2021continual}, ReDO \citep{sokar2023dormant}
\item Regularization: L$2$ Regularization (L$2$), Shrink \& Perturb \citep{ash2020warm}
\item Architectural: Concatenated ReLU (Concat ReLU) \citep{shang2016understanding, abbas2023loss}, Layer Normalization (Layer Norm) \citep{ba2016layer} 
\end{itemize} 

\renewcommand\thesubfigure{\hspace{0.5cm}(\alph{subfigure})}
\captionsetup[subfigure]{labelformat=simple, labelsep=none}
\begin{figure*}[!htb]

    \centering
        
        \begin{subfigure}{0.3\textwidth}
            \caption*{Permuted MNIST} %
            \includegraphics[width=\linewidth]{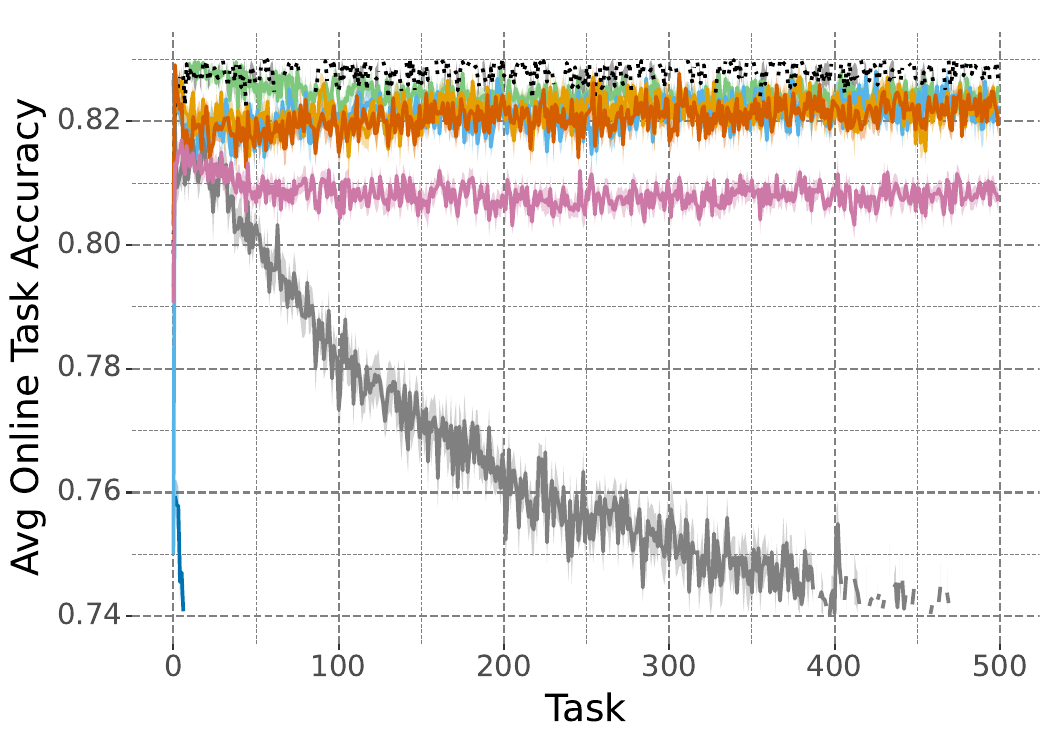}
            \vspace{-0.5cm} %
        \end{subfigure} 
        \begin{subfigure}{0.3\textwidth}
            \caption*{Random Label MNIST} %
            \includegraphics[width=\linewidth]{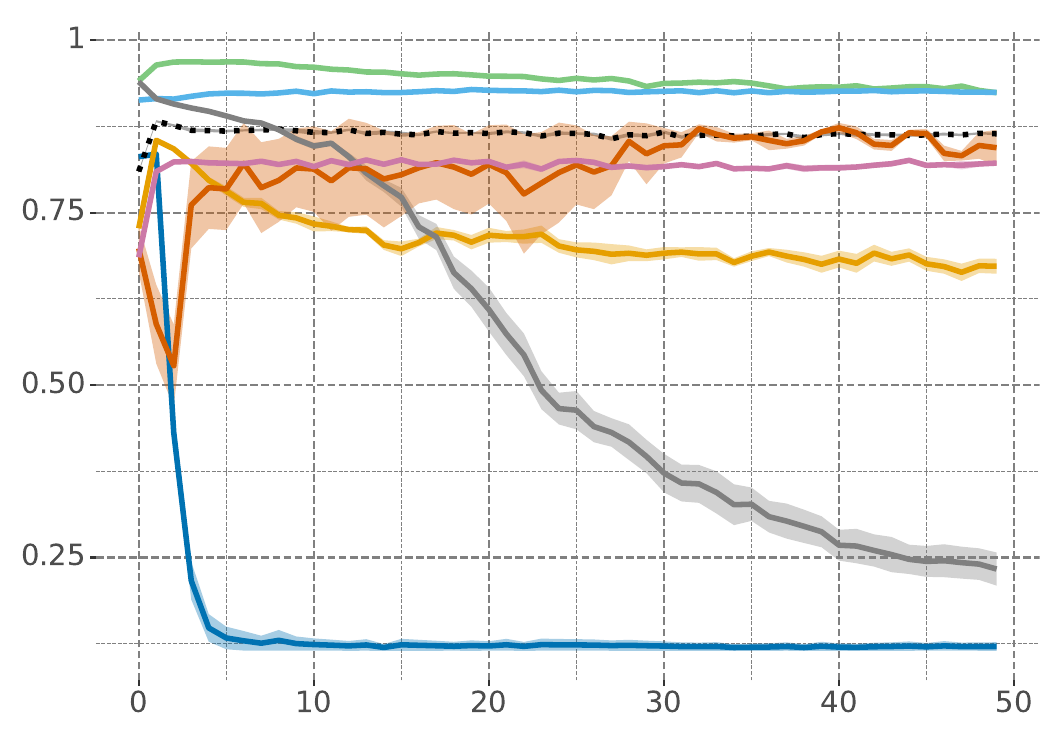}
            \vspace{-0.5cm} %
        \end{subfigure} 
        \begin{subfigure}{0.3\textwidth}
            \caption*{Random Label CIFAR} %
            \includegraphics[width=\linewidth]{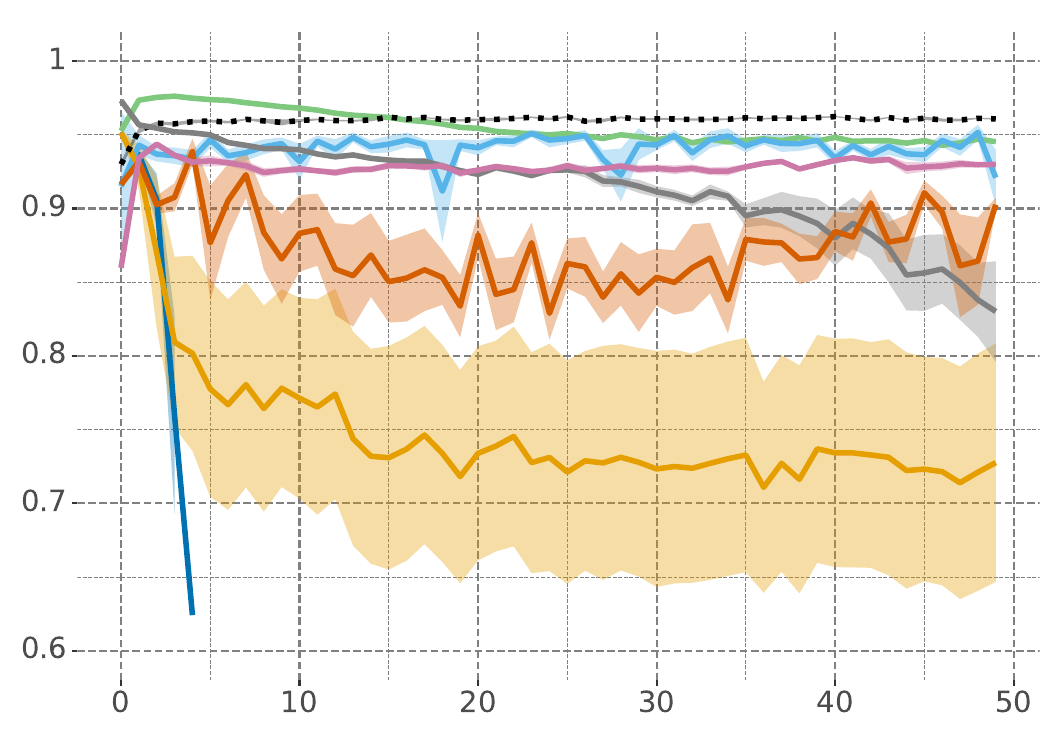}
            \vspace{-0.5cm} %
        \end{subfigure} \\

        \begin{subfigure}{0.3\textwidth}
            \caption*{5+1 CIFAR} %
            \includegraphics[width=\linewidth]{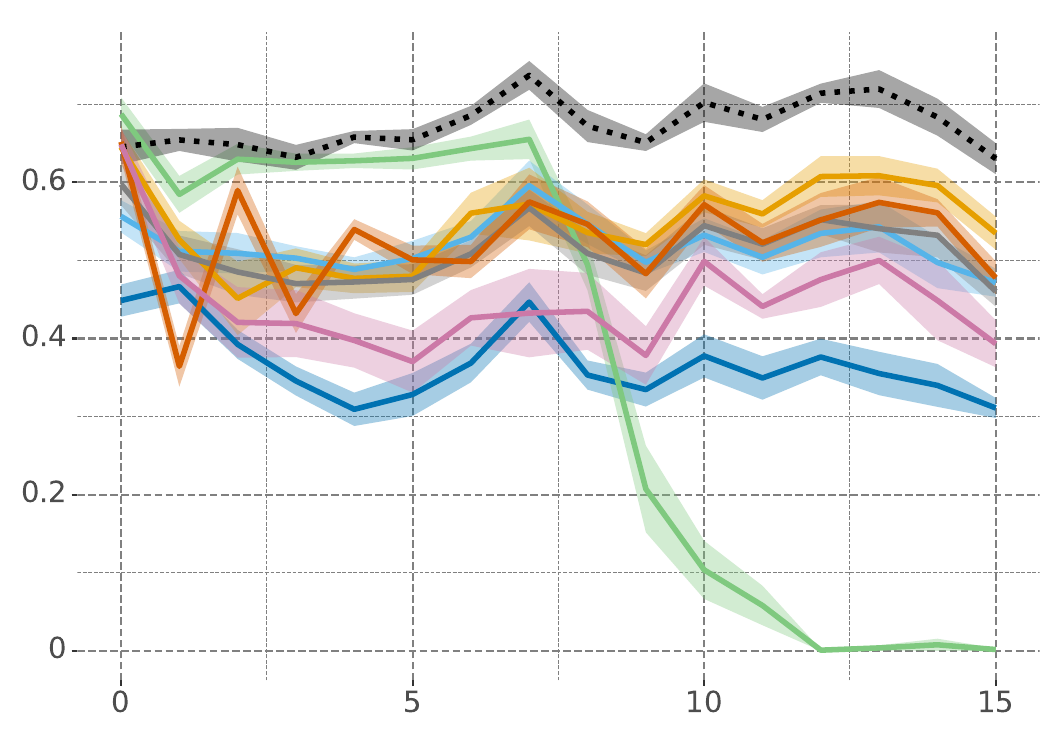}
            \vspace{-0.1cm} %
        \end{subfigure} 
        \begin{subfigure}{0.3\textwidth}
            \caption*{Continual ImageNet} %
            \includegraphics[width=\linewidth]{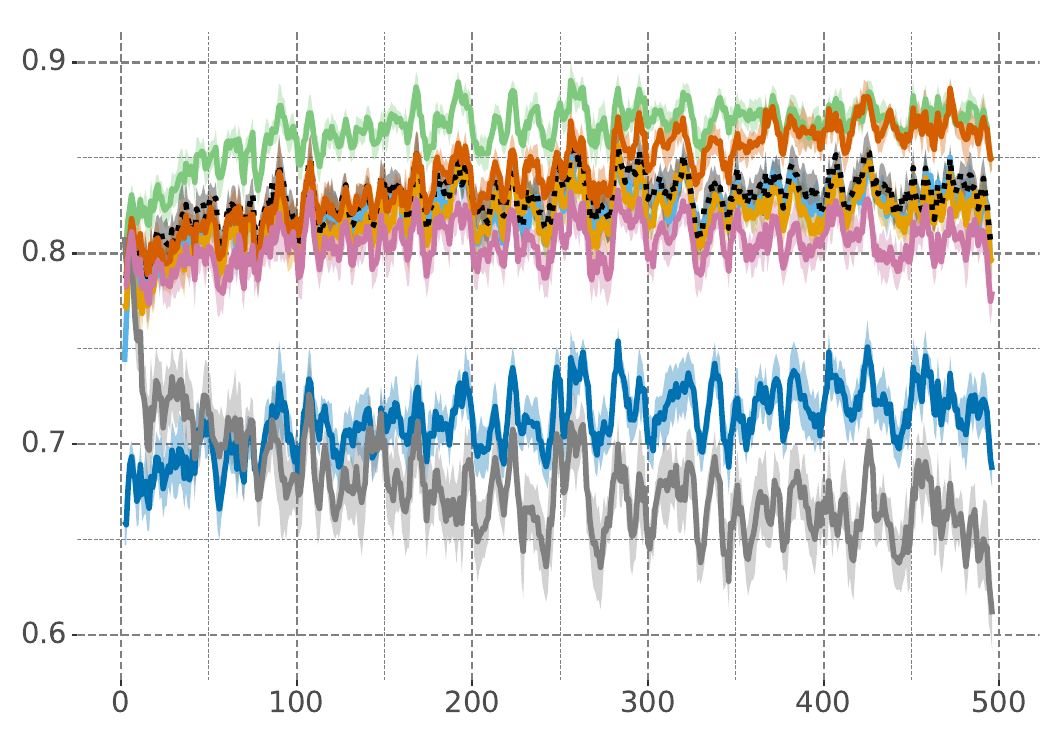}
            \vspace{-0.1cm} %
        \end{subfigure}

\begin{tikzpicture}

    \def\firstRowY{0.5}
    \draw [Baseline, thick, line width=2pt] (-2,\firstRowY) -- (-1.5,\firstRowY);
    \node[anchor=west] at (-1.5,\firstRowY) {Baseline};
    
    \draw [LayerNorm, thick, line width=2pt] (1,\firstRowY) -- (1.5,\firstRowY);
    \node[anchor=west] at (1.5,\firstRowY) {Layer Norm};

    \draw [ShrinkAndPerturb, thick, line width=2pt] (4,\firstRowY) -- (4.5,\firstRowY);
    \node[anchor=west] at (4.5,\firstRowY) {Shrink \& Perturb};

    \draw [ReDO, thick, line width=2pt] (8,\firstRowY) -- (8.5,\firstRowY);
    \node[anchor=west] at (8.5,\firstRowY) {ReDO};

    \draw [L2Init, thick, dotted, line width=2pt] (-2,0) -- (-1.5,0);
    \node[anchor=west] at (-1.5,0) {L2 Init};
    
    \draw [L2, thick, line width=2pt] (1,0) -- (1.5,0);
    \node[anchor=west] at (1.5,0) {L2};
    
    \draw [ContinualBackprop, thick, line width=2pt] (4,0) -- (4.5,0); 
    \node[anchor=west] at (4.5,0) {Continual Backprop};

    \draw [ConcatReLU, thick, line width=2pt] (8,0) -- (8.5,0);
    \node[anchor=west] at (8.5,0) {Concat ReLU};
\end{tikzpicture}
    
    \caption{Comparison of average online task accuracy across all five problems when using the Adam optimizer. \methodname consistently maintains plasticity. While L$2$ mitigates plasticity loss completely on Permuted MNIST and Continual ImageNet, this method performs poorly on Random Label MNIST, Random Label CIFAR, and 5+1 CIFAR. Concat ReLU generally performs very well, except on 5+1 CIFAR where it suffers a sharp drop in performance.
    }
    \label{fig:adam-performance-comparison}
\end{figure*}

\begin{figure*}[!htbp]
    \centering
    {\fontsize{6.5pt}{7.7pt}\selectfont
    \setlength{\tabcolsep}{0pt}
    \begin{tabular}{r}
        L2 Init \\
        Concat ReLU \\
        Continual Backprop \\
        ReDO \\
        L2 \\
        Shrink and Perturb \\
        Layer Norm \\
        Baseline \\
        \\
        \vspace{-0.19cm}
    \end{tabular}
    }
    \hspace{0.001cm}  %
    \begin{minipage}{0.25\textwidth}
        \caption*{\small Permuted MNIST}
        \vspace{-0.35cm}
        \includegraphics[width=\textwidth]{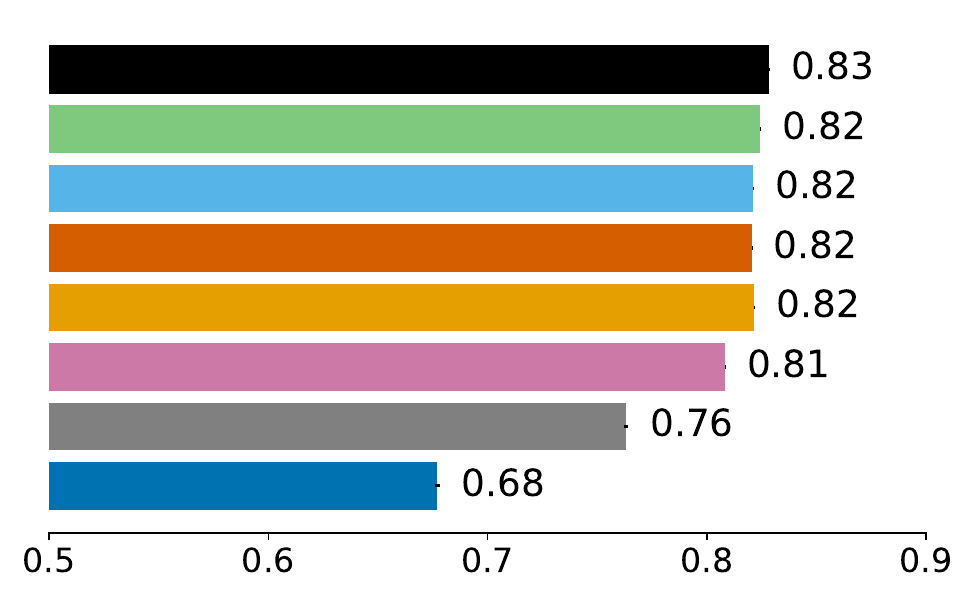}
        \vspace{0.1cm}
    \end{minipage} 
    \begin{minipage}{0.25\textwidth}
        \caption*{\small Random Label MNIST}
        \vspace{-0.35cm}
        \includegraphics[width=\textwidth]{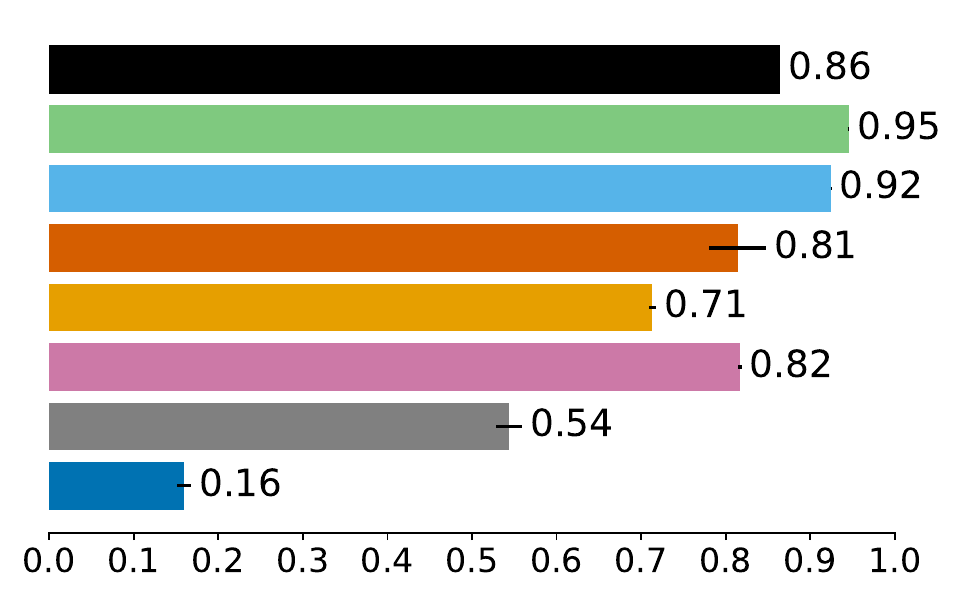}
        \vspace{0.1cm}
    \end{minipage} 
    \begin{minipage}{0.25\textwidth}
        \caption*{\small Random Label CIFAR}
        \vspace{-0.35cm}
        \includegraphics[width=\textwidth]{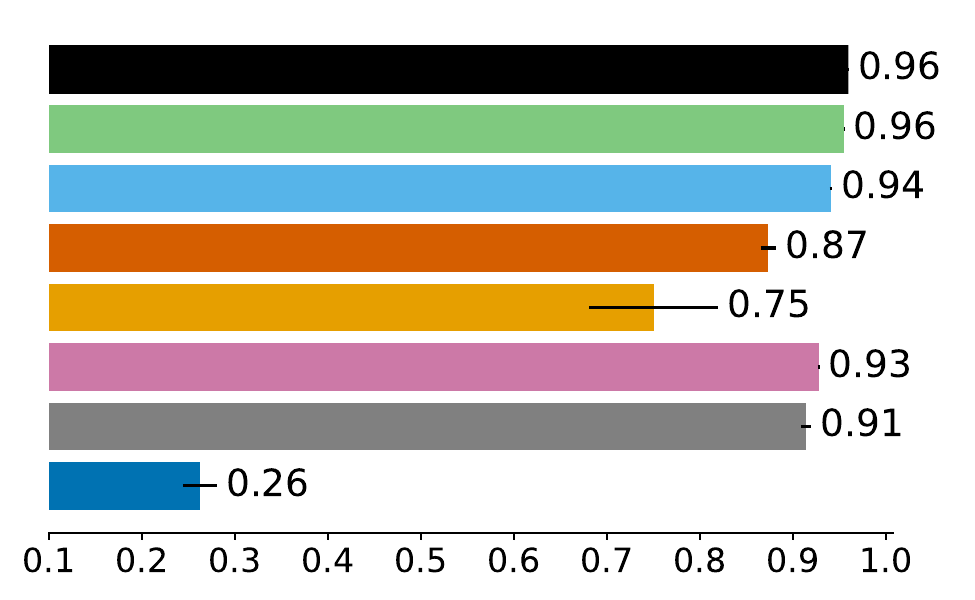}
        \vspace{0.1cm}
    \end{minipage}
     \\
    {\fontsize{6.5pt}{7.7pt}\selectfont
    \setlength{\tabcolsep}{0pt}
    \begin{tabular}{r}
        \\
        L2 Init \\
        Concat ReLU \\
        Continual Backprop \\
        ReDO \\
        L2 \\
        Shrink and Perturb \\
        Layer Norm \\
        Baseline \\
        \vspace{-0.15cm}
    \end{tabular}
    }
    \hspace{0.001cm}  %
    \begin{minipage}{0.25\textwidth}
        \caption*{\small 5+1 CIFAR}
        \vspace{-0.35cm}
        \includegraphics[width=\textwidth]{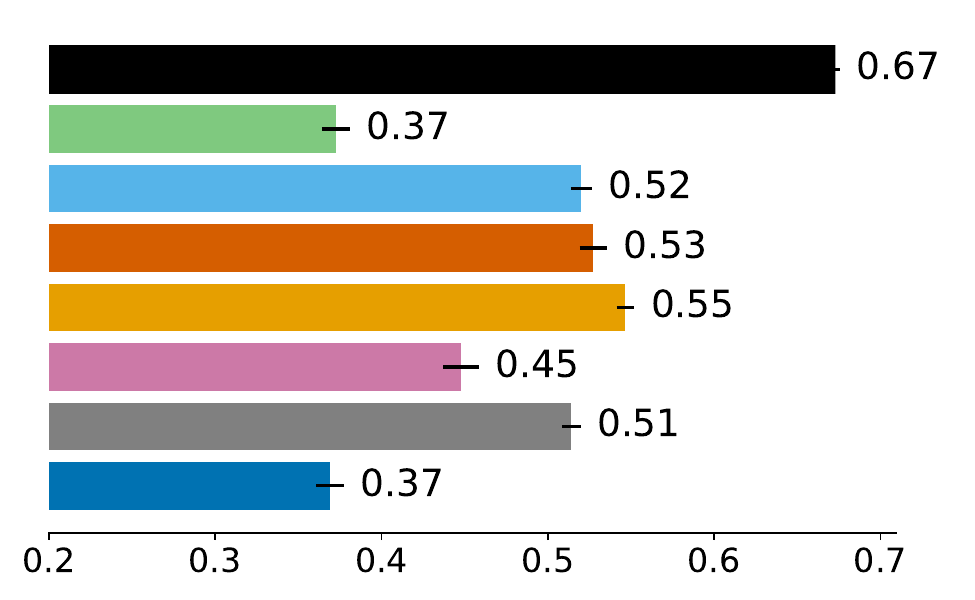}
    \end{minipage} 
    \begin{minipage}{0.25\textwidth}
        \caption*{\small Continual ImageNet}
        \vspace{-0.35cm}
        \includegraphics[width=\textwidth]{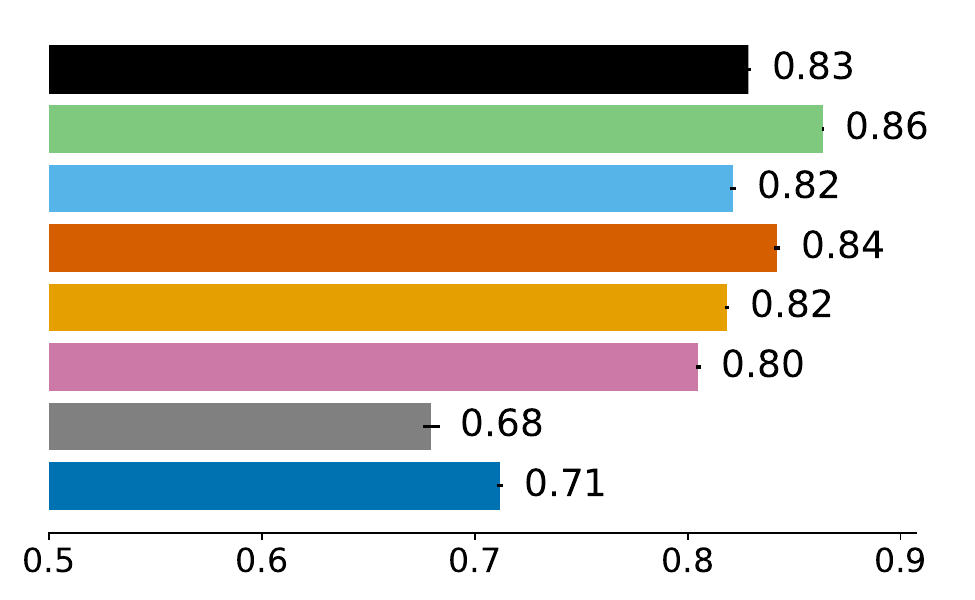}
    \end{minipage}
    
    \caption{Comparison of total average online accuracy across all five problems when using the Adam optimizer. \methodname performs in the top 3 in each of the five environments. On 5+1 CIFAR, it significantly outperforms all other methods. Concat ReLU does well on all problems except for 5+1 CIFAR.} 
    \label{fig:adam-total-avg-acc}
\end{figure*}

\textbf{Evaluation.} On all problems, we perform a hyper-parameter sweep for each method and average results over $3$ seeds. For each method, we select the configuration that resulted in the largest total average online accuracy. We then run the best configuration for each method on $10$ additional seeds, which produces the results in Figures~\ref{fig:adam-performance-comparison}-\ref{fig:adam-ablation}. In addition to determinining the initialization of the neural network, each seed also determines the problem parameters, such as the data comprising each task, and the sequence of sampled (data, label) pairs from each task's dataset. For instance, on Permuted MNIST, the seed determines a unique sequence of permutations applied to the images resulting in a unique task sequence, as well as how the task data is shuffled. As another example, on Continual ImageNet, it determines the pairs of classes that comprise each task, the sequence of tasks, and the sequence of batches in each task. For all problems, the seed determines a unique set of tasks and sequence of those tasks.

\textbf{Hyper-parameters.} 
We train all agents with the Adam optimizer. Since recent work has argued against the use of Adam in continual learning \citep{ashley2021does}, we additionally train agents with SGD and include results in Appendix \ref{appendix:sgd_results}. For all agents, we sweep over stepsizes $\alpha \in \{ 1\mathrm{e}{-3}, 1\mathrm{e}{-4} \}$ when using Adam. For L$2$ and \methodname, we sweep over regularization strength $\lambda \in \{1\mathrm{e}{-2}, 1\mathrm{e}{-3}, 1\mathrm{e}{-4}, 1\mathrm{e}{-5} \}$. For Shrink \& Perturb, we perform a grid search over shrinkage parameter $p \in \{1\mathrm{e}{-2}, 1\mathrm{e}{-3}, 1\mathrm{e}{-4}, 1\mathrm{e}{-5} \}$ and noise scale $\sigma \in \{1\mathrm{e}{-2}, 1\mathrm{e}{-3}, 1\mathrm{e}{-4}, 1\mathrm{e}{-5} \}$. For Continual Backprop, we sweep over the replacement rate \rebuttal{$r \in \{1\mathrm{e}{-1}, 1\mathrm{e}{-2}, 1\mathrm{e}{-3}, 1\mathrm{e}{-4}, 1\mathrm{e}{-5}, 1\mathrm{e}{-6}\}$} and use the values reported in~\citet{dohare2023maintaining} for other hyperparameters. For ReDO, we sweep over the recycle period by recycling neurons either every $1$, $2$, or $5$ tasks, and we sweep over the recycle threshold in the set $\{ 0, 0.01, 0.1 \}$. Finally, as a baseline method, we run Adam, using the PyTorch default hyperparameters other than the stepsize. Additional training details, including neural network architectures and hyper-parameter settings, are in Appendix \ref{appendix:agents}.

\textbf{\rebuttal{Parameter Initialization.}}
\rebuttal{For all agents, neural networks are initialized using PyTorch default initialization. In each layer $l$ of the neural network, each weight and bias is sampled from the uniform distribution $\mathcal{U}(\frac{-1}{\sqrt{\text{fan\_in}(l)}}, \frac{1}{\sqrt{\text{fan\_in}(l)}})$ where $\text{fan\_in}(l)$ is the input dimension for layer $l$. In fully-connected layers, this is the number of incoming weights to each neuron, and in convolutional layers, this is the number of input channels. The initial parameters that L2 Init regresses towards are the neural network weights drawn from this distribution at the beginning of training.}

\subsection{Comparative Evaluation}

We plot the average online task accuracy and the total average online task accuracy for all methods when using Adam in Figures \ref{fig:adam-performance-comparison} and \ref{fig:adam-total-avg-acc}. On all five problems, the Baseline method either significantly loses plasticity over time or performs poorly overall. Because we select hyperparameters based on total average online accuracy, the Baseline method is sometimes run with a smaller learning rate which results in low plasticity loss but still relatively poor performance. Importantly, \methodname consistently retains high plasticity across problems and maintains high average online task accuracy throughout training. \methodname has comparable performance to the two resetting methods Continual Backprop and ReDO. Specifically, it performs as well as or better than Continual Backprop on four out of the five problems. The same roughly holds true when comparing to the performance of ReDO. 

Concat ReLU performs well on all problems except 5+1 CIFAR on which it loses plasticity completely. Concat ReLU loses some plasticity on Random Label MNIST and Random Label CIFAR, but the overall performance is still quite high. While L2 significantly mitigates plasticity loss on Permuted MNIST, there is still large plasticity loss on Random Label MNIST, Random Label CIFAR, and 5+1 CIFAR as compared to \methodname. Shrink \& Perturb does mitigate plasticity loss on all problems, but overall performance is consistently lower than that of \methodname. Finally, Layer Norm mitigates only some plasticity loss. 

\subsection{Looking inside the network}
\begin{figure*}

    \centering
    \setlength{\tabcolsep}{2pt}
    
    \begin{tabular}{>
    {\centering\arraybackslash}m{0.23\textwidth}>
    {\centering\arraybackslash}m{0.23\textwidth}>
    {\centering\arraybackslash}m{0.23\textwidth}>
    {\centering\arraybackslash}m{0.23\textwidth}
    }
    
        \multicolumn{2}{c}{\hspace{0.2cm} Permuted MNIST} 
        & \multicolumn{2}{c}{\hspace{0.2cm} Random Label MNIST} \\
        \hspace{0.4cm}\footnotesize{Weight Magnitude} &
        \hspace{0.4cm}\footnotesize{Feature SRank} &
        \hspace{0.4cm}\footnotesize{Weight Magnitude} &
        \hspace{0.4cm}\footnotesize{Feature SRank} \\
        
        \begin{subfigure}{0.23\textwidth}
            \includegraphics[width=\linewidth]{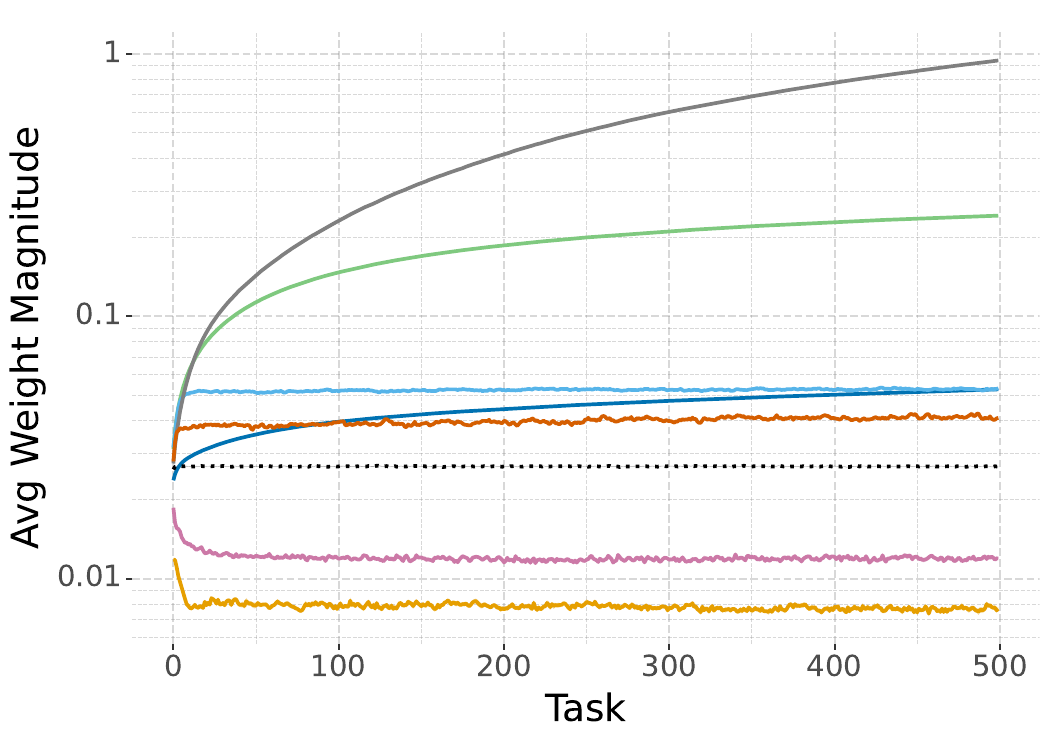}
            \vspace{-0.7cm}
            \caption*{} %
        \end{subfigure} &
        \begin{subfigure}{0.23\textwidth}
            \includegraphics[width=\linewidth]{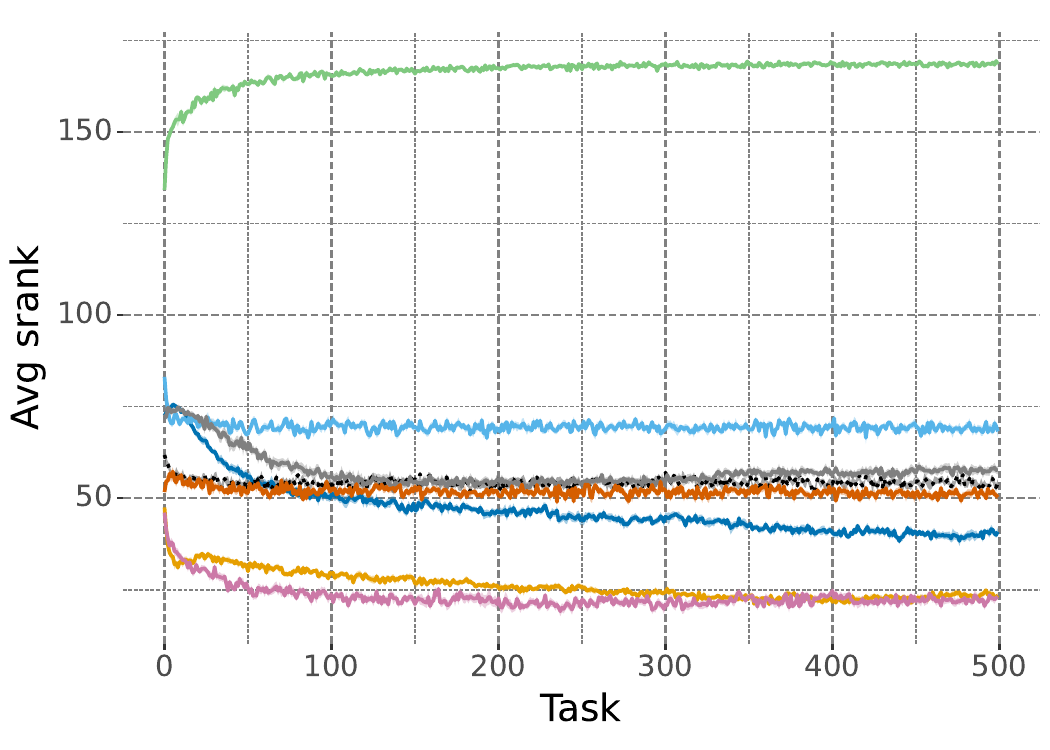}
            \vspace{-0.7cm}
            \caption*{} %
        \end{subfigure} &
        \begin{subfigure}{0.23\textwidth}
            \includegraphics[width=\linewidth]{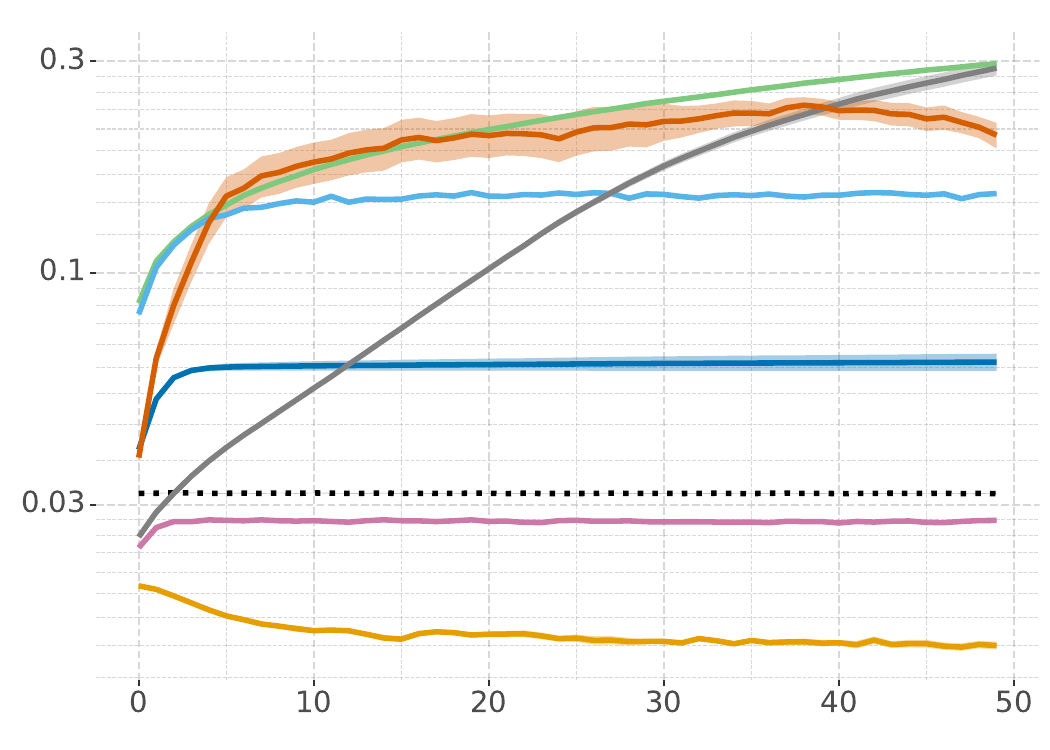}
            \vspace{-0.7cm}
            \caption*{} %
        \end{subfigure} &
        \begin{subfigure}{0.23\textwidth}
            \includegraphics[width=\linewidth]{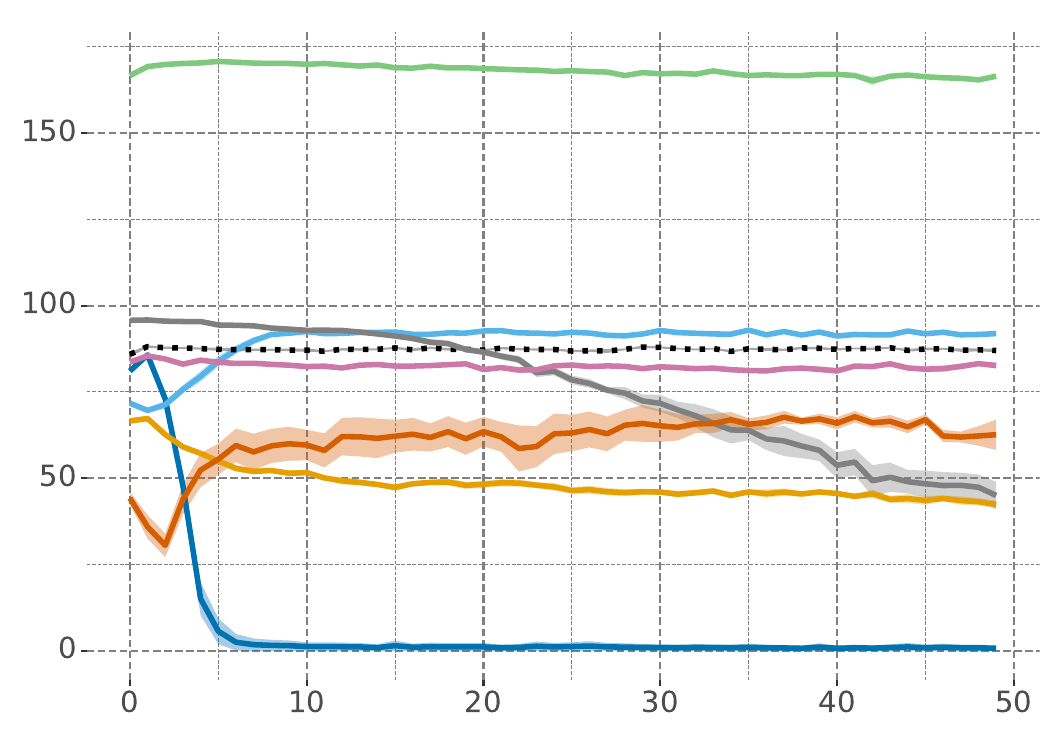}
            \vspace{-0.7cm}
            \caption*{} %
        \end{subfigure} \\

        \multicolumn{2}{c}{\hspace{0.2cm} 5+1 CIFAR} 
        & \multicolumn{2}{c}{\hspace{0.2cm} Continual ImageNet} \\
        \hspace{0.4cm}\footnotesize{Weight Magnitude} &
        \hspace{0.4cm}\footnotesize{Feature SRank} &
        \hspace{0.4cm}\footnotesize{Weight Magnitude} &
        \hspace{0.4cm}\footnotesize{Feature SRank} \\

        \begin{subfigure}{0.23\textwidth}
            \includegraphics[width=\linewidth]{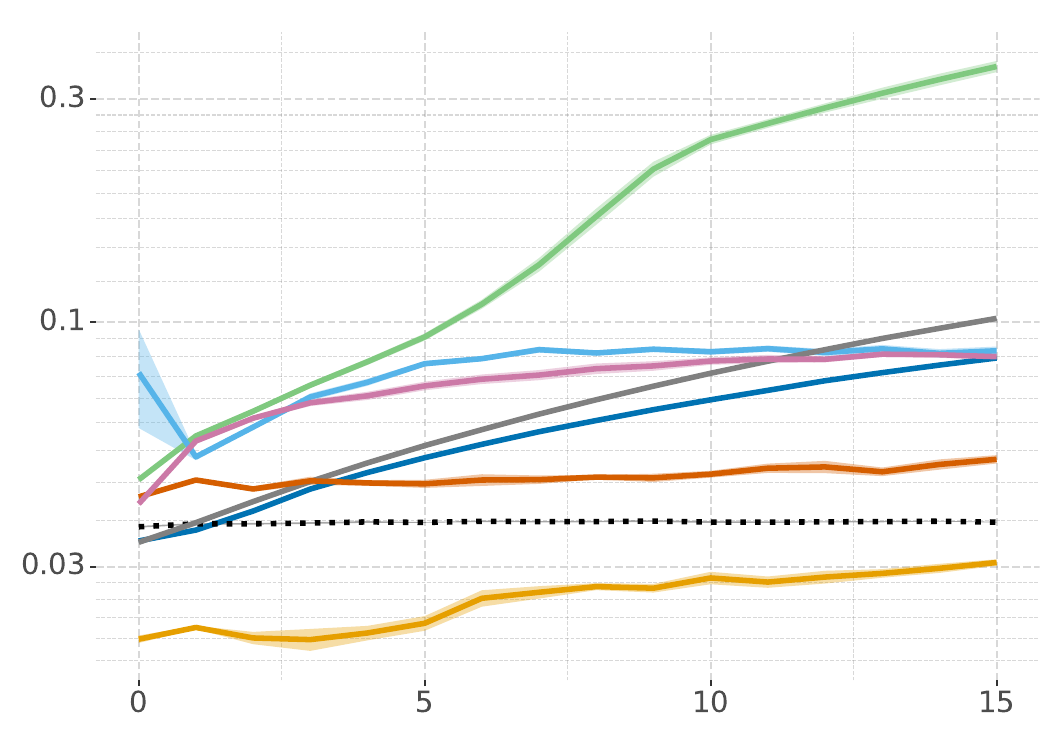}
            \vspace{-0.7cm}
            \caption*{} %
        \end{subfigure} &
        \begin{subfigure}{0.23\textwidth}
            \includegraphics[width=\linewidth]{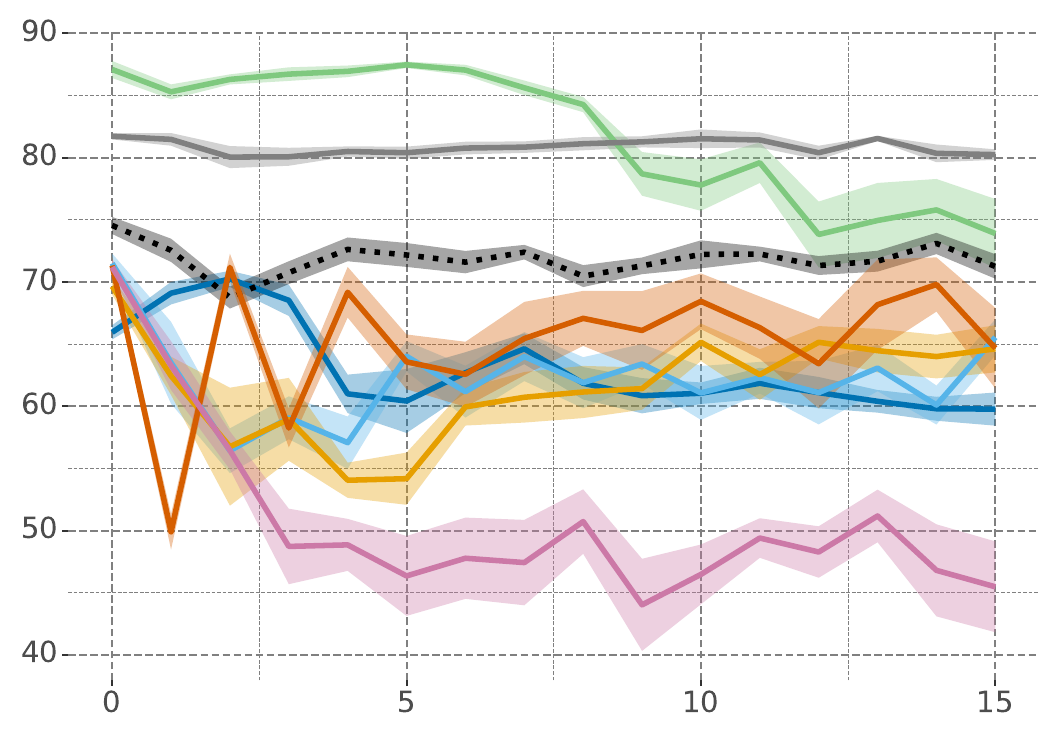}
            \vspace{-0.7cm}
            \caption*{} %
        \end{subfigure} &
        \begin{subfigure}{0.23\textwidth}
            \includegraphics[width=\linewidth]{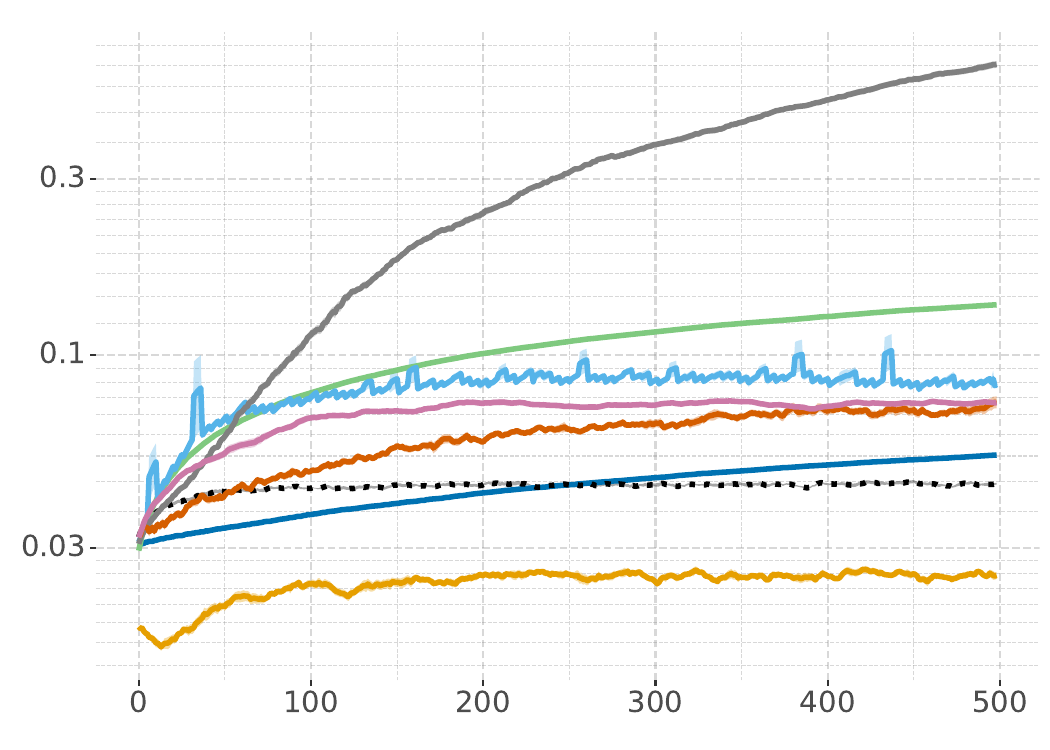}
            \vspace{-0.7cm}
            \caption*{} %
        \end{subfigure} &
        \begin{subfigure}{0.23\textwidth}
            \includegraphics[width=\linewidth]{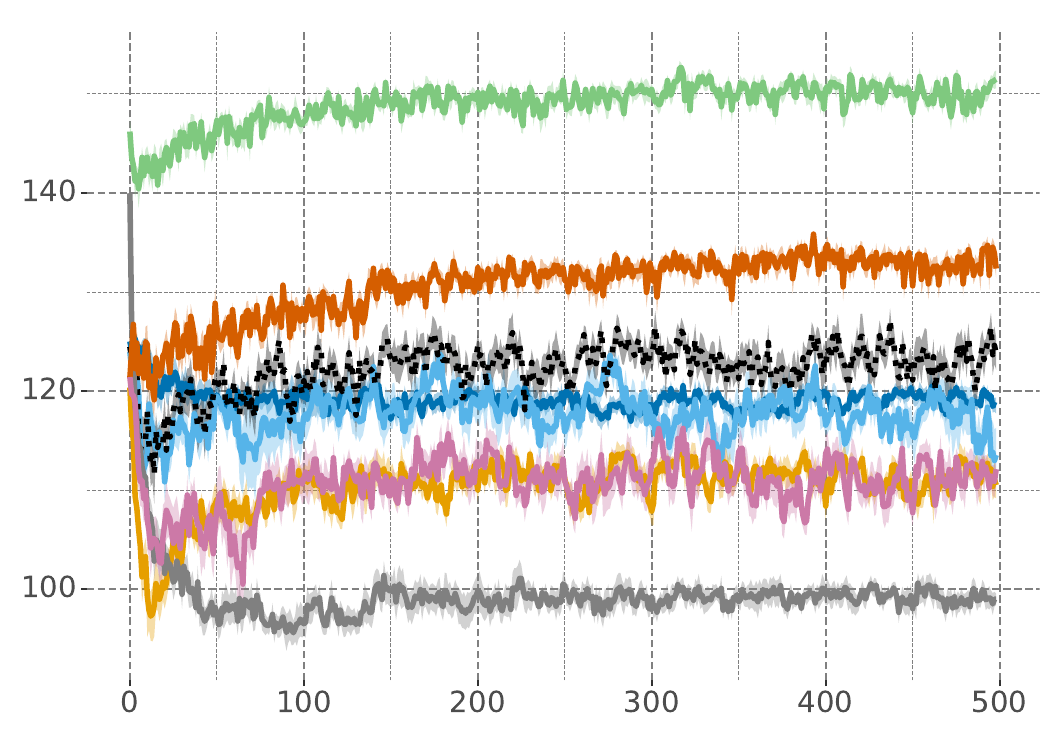}
            \vspace{-0.7cm}
            \caption*{} %
        \end{subfigure}\\
        
    \end{tabular}

    \begin{tikzpicture}

    \def\firstRowY{0.5}
    \draw [Baseline, thick, line width=2pt] (-2,\firstRowY) -- (-1.5,\firstRowY);
    \node[anchor=west] at (-1.5,\firstRowY) {Baseline};
    
    \draw [LayerNorm, thick, line width=2pt] (1,\firstRowY) -- (1.5,\firstRowY);
    \node[anchor=west] at (1.5,\firstRowY) {Layer Norm};

    \draw [ShrinkAndPerturb, thick, line width=2pt] (4,\firstRowY) -- (4.5,\firstRowY);
    \node[anchor=west] at (4.5,\firstRowY) {Shrink \& Perturb};

    \draw [ReDO, thick, line width=2pt] (8,\firstRowY) -- (8.5,\firstRowY);
    \node[anchor=west] at (8.5,\firstRowY) {ReDO};

    \draw [L2Init, thick, dotted, line width=2pt] (-2,0) -- (-1.5,0);
    \node[anchor=west] at (-1.5,0) {L2 Init};
    
    \draw [L2, thick, line width=2pt] (1,0) -- (1.5,0);
    \node[anchor=west] at (1.5,0) {L2};
    
    \draw [ContinualBackprop, thick, line width=2pt] (4,0) -- (4.5,0); 
    \node[anchor=west] at (4.5,0) {Continual Backprop};

    \draw [ConcatReLU, thick, line width=2pt] (8,0) -- (8.5,0);
    \node[anchor=west] at (8.5,0) {Concat ReLU};
\end{tikzpicture}

    \caption{Average weight magnitude and feature rank over time when training all agents using Adam. \methodname retains a relatively small average weight magnitude and high feature rank.}
    \label{fig:adam-network-metrics}
\end{figure*}

While the causes of plasticity loss remain unclear, it is likely that large parameter magnitudes as well as a reduction in feature rank can play a role. For instance, ReLU units that stop activating regardless of input will have zero gradients and will not be updated, therefore potentially not adapting to future tasks. To understand how \methodname affects neural network dynamics, we plot the average weight magnitude (L1 norm) as well as the average feature rank computed at the end of each task on four problems when training using the Adam optimizer (Figure~\ref{fig:adam-network-metrics}).  

A measure of the effective rank of a matrix, that \citet {kumar2020implicit} call \textit{srank}, is computed from the singular values of the matrix. Specifically, using the ordered set of singular values $\sigma_1 > \sigma_2,... \sigma_n$, we compute the srank as
\begin{align*}
    \text{srank} = \min_k \frac{\sum_{i=1}^k \sigma_i }{\sum_{j=1}^n \sigma_j} \geq 1 - \delta
\end{align*}
using the threshold $\delta=0.01$ following \cite{kumar2020implicit}. Thus, in this case, the srank is how many singular values you need to sum up to make up $99\%$ of the total sum of singular values.

In Figure \ref{fig:adam-network-metrics}, we see that both \methodname and L2 reduce the average weight magnitude relative to the Baseline. As pointed out by \citet{dohare2021continual}, this is potentially important when using the Adam optimizer. Since the updates with Adam are bounded by the global stepsize or a small multiple of the global stepsize, when switching to a new task, the relative change in these weights may be small. However, agents which perform quite well, such as Continual Backprop and Concat ReLU, result in surprisingly large average weight magnitude, making any clear takeaway lacking. However, on 5+1 CIFAR the weight magnitude of Concat ReLU is very large relative to other methods, potentially explaining its sharp drop in performance in Figure~\ref{fig:adam-performance-comparison}.

When using L$2$, the effective feature rank is smaller than it is when applying \methodname. This is to be expected since \methodname is regularizing towards a set of full-rank matrices, and could potentially contribute to the increased plasticity we see with \methodname. Notably, Concat ReLU enjoys high feature rank across problems (with the exception of 5+1 CIFAR) which is potentially contributing to its high performance.

\subsection{Ablation Study of Regenerative Regularization}\label{sec:ablation}

\begin{figure*}

    \centering
    \setlength{\tabcolsep}{2pt}
    
\begin{tabular}{>{\centering\arraybackslash}m{0.2cm}>{\centering\arraybackslash}m{0.3\textwidth}>{\centering\arraybackslash}m{0.3\textwidth}>{\centering\arraybackslash}m{0.3\textwidth}} \\

         \rotatebox[origin=c]{90}{} \vspace{0.5cm} & 
        \begin{subfigure}{0.28\textwidth}
            \caption*{Permuted MNIST} %
            \includegraphics[width=\linewidth]{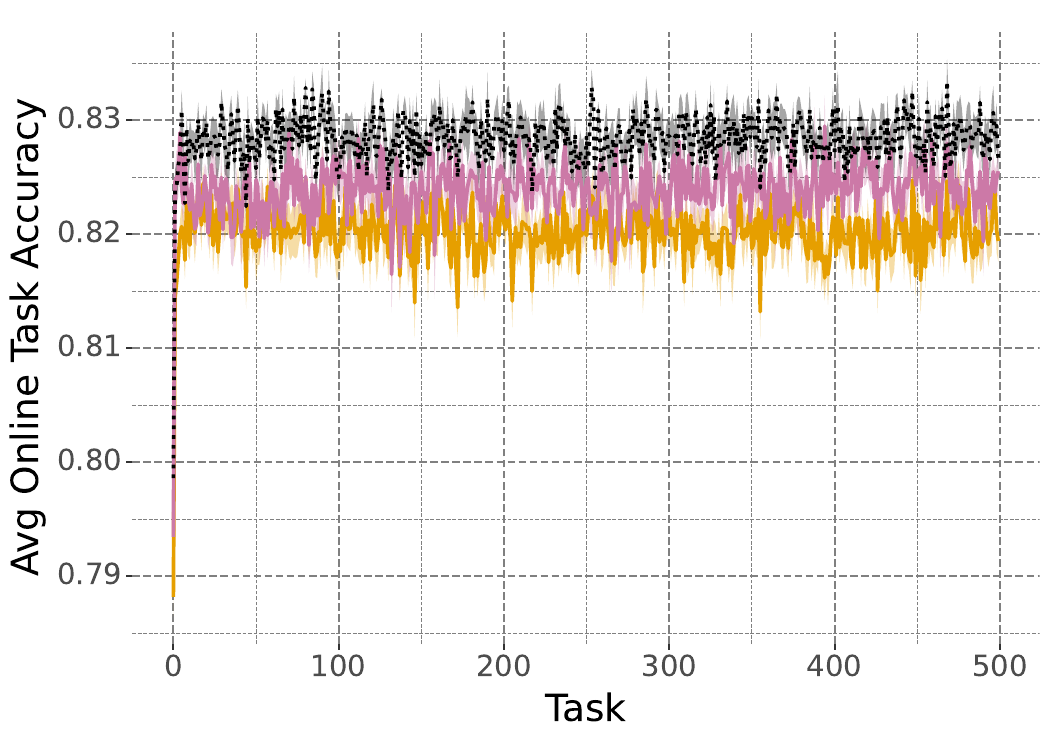}
                \vspace{-0.3cm} %
        \end{subfigure} &
        \begin{subfigure}{0.28\textwidth}
            \caption*{Random Label MNIST} %
            \includegraphics[width=\linewidth]{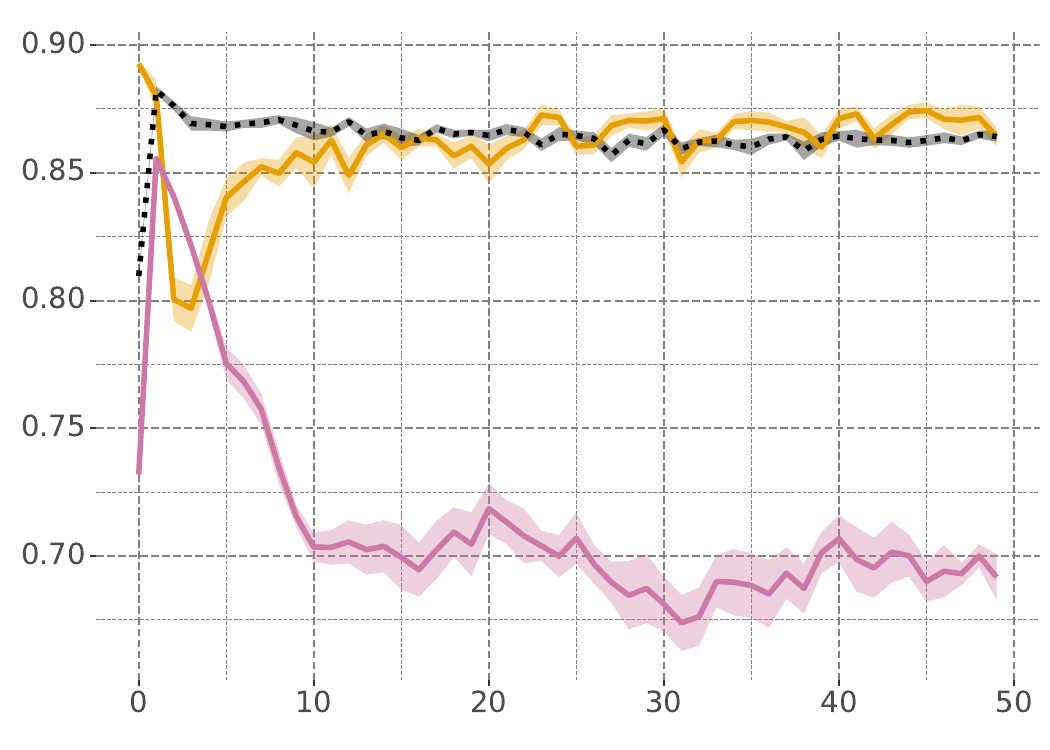}
                \vspace{-0.3cm} %
        \end{subfigure} &
        \begin{subfigure}{0.28\textwidth}
            \caption*{5+1 CIFAR} %
            \includegraphics[width=\linewidth]{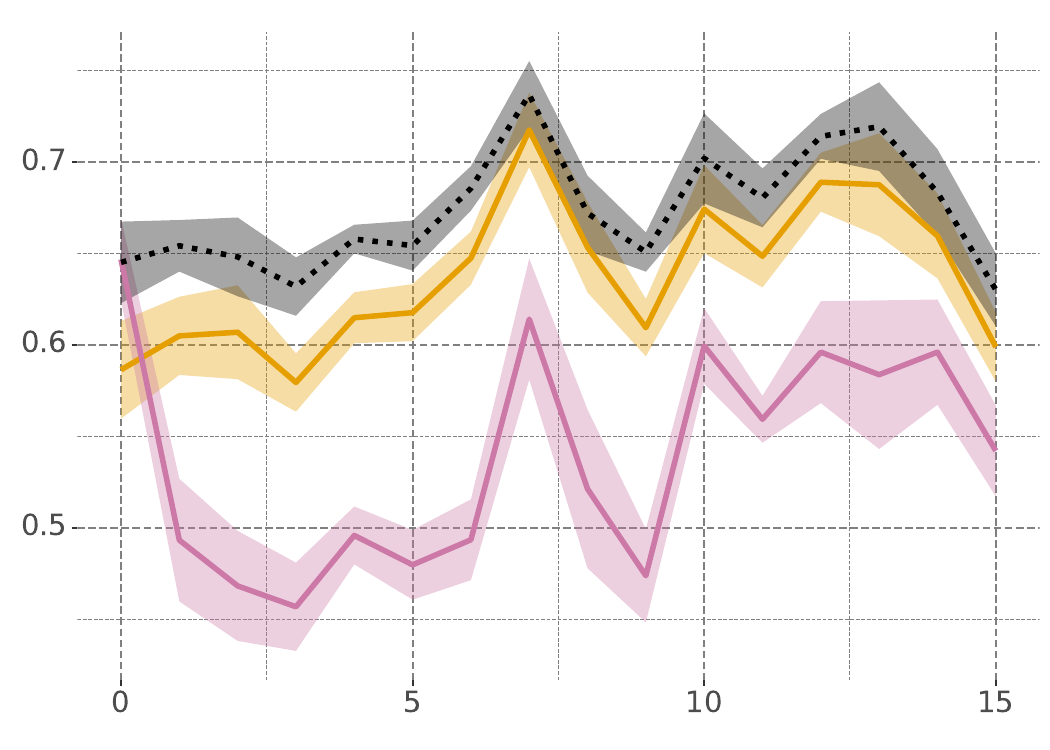}
                \vspace{-0.3cm} %
        \end{subfigure}
    \end{tabular}

\begin{tikzpicture}
    \draw [L2Init, thick, dotted, line width=2pt] (-2,0) -- (-1.5,0);
    \node[anchor=west] at (-1.5,0) {L2 Init};

    \draw [L2InitResample, thick, line width=2pt] (1,0) -- (1.5,0);
    \node[anchor=west] at (1.5,0) {L2 Init + Resample};
    
    \draw [L1Init, thick, line width=2pt] (6,0) -- (6.5,0);
    \node[anchor=west] at (6.5,0) {L1 Init};


\end{tikzpicture}
    
    \caption{Comparison of L2 Init, L2 Init + Resample, and L1 Init on three problems when using Adam. L2 Init + Resample performs poorly on all environments, especially on Random Label MNIST and 5+1 CIFAR where it loses plasticity. L1 Init matches the performance of L2 Init on Random Label MNIST and performs slightly worse on Permuted MNIST and 5+1 CIFAR. 
    }
    \label{fig:adam-ablation}
\end{figure*}

\textbf{Regularizing toward random parameters.} With \methodname, we regularize toward the specific fixed parameters $\theta_0$ sampled at initialization. Following a procedure more similar to Shrink \& Perturb, we could alternatively sample a new set of parameters at each time step. That is, we could sample $\phi_t$ from the same distribution that $\theta_0$ was sampled from and let the regularization term be $||\theta_t - \phi_t||_2^2$ instead. In Figure \ref{fig:adam-ablation}, we compare the performance between \methodname and this variant (L$2$ Init + Resample) on Permuted MNIST, Random Label MNIST, and 5+1 CIFAR when using the Adam optimizer. We select the best regularization strength for each method using the same hyper-parameter sweep used for \methodname. We find that regularizing towards the initial parameters rather than sampling a new set of parameters at each time step performs much better.

\textbf{Choice of norm.} While \methodname uses the L2 norm, we could alternatively use the L1 norm of the difference between the parameters and their initial values. We call this approach \textit{L1 Init}, which uses the following loss function:
\begin{align*}
    \mathcal{L}_{\text{reg}}(\theta) = \mathcal{L}_{\text{train}}(\theta) + \lambda ||\theta - \theta_0||_1
\end{align*}


We compare the performance of \methodname and L1 Init on Permuted MNIST, Random Label MNIST, and 5+1 CIFAR when using the Adam optimizer (see Figure~\ref{fig:adam-network-metrics}). We find that while L1 Init mitigates plasticity loss, the performance is worse on Permuted MNIST and 5+1 CIFAR. 

\subsection{Robustness to Network Width}\label{sec:depth_and_width}
To determine whether L2 Init remains effective when using a wider network, we evaluate L2 Init's performance on a subset of problems when using a network with additional neurons in each hidden layer. We use the same neural network architectures as described in Section \ref{appendix:agents} but increase the number neurons in each layer by $4$x. 
We find that L2 Init's effectiveness is not diminished by increased network width, as shown in Figure~\ref{fig:width}. We repeat this study with increasing network depth in Appendix~\ref{appendix:depth}.

\renewcommand\thesubfigure{\hspace{0.5cm}(\alph{subfigure})}
\captionsetup[subfigure]{labelformat=simple, labelsep=none, font=footnotesize}

\renewcommand\thesubfigure{\hspace{0.5cm}(\alph{subfigure})}
\captionsetup[subfigure]{labelformat=simple, labelsep=none, font=footnotesize}
\begin{figure*}

    \centering
    \setlength{\tabcolsep}{2pt}
    
\begin{tabular}{>{\centering\arraybackslash}m{0.2cm}>{\centering\arraybackslash}m{0.32\textwidth}>{\centering\arraybackslash}m{0.32\textwidth}>{\centering\arraybackslash}m{0.32\textwidth}} \\
         \rotatebox[origin=c]{90}{} \vspace{0.5cm} & 
        \begin{subfigure}{0.28\textwidth}
            \caption*{Permuted MNIST} %
            \includegraphics[width=\linewidth]{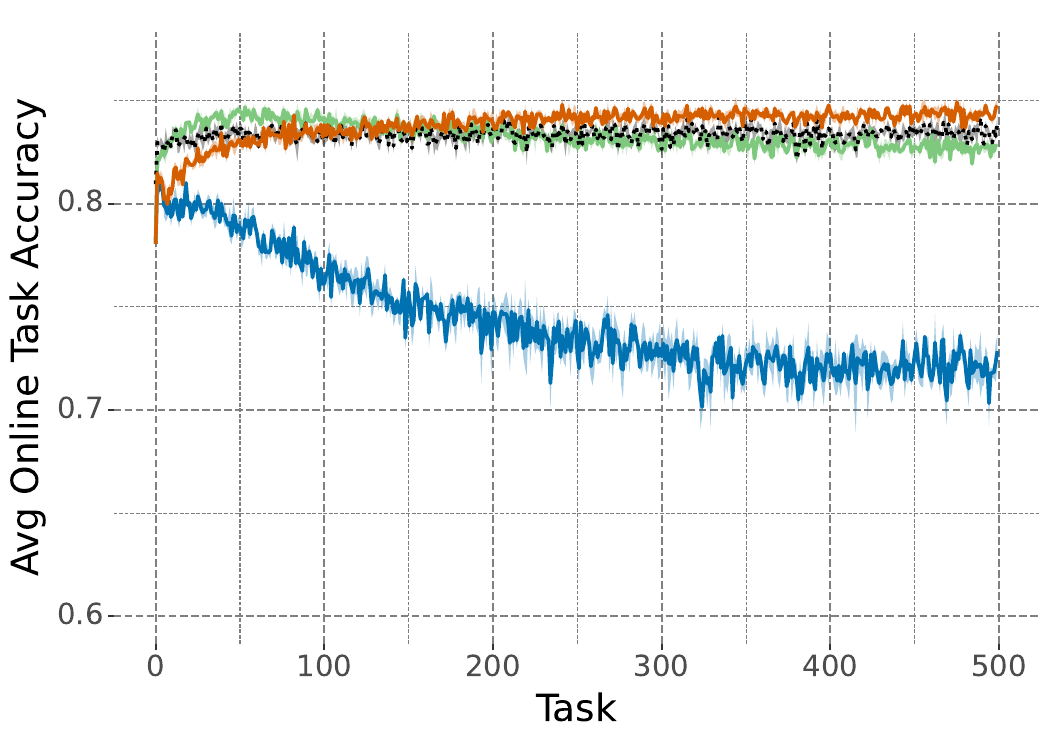}
                \vspace{-0.3cm} %
        \end{subfigure} &
        \begin{subfigure}{0.28\textwidth}
            \caption*{Random Label CIFAR} %
            \includegraphics[width=\linewidth]{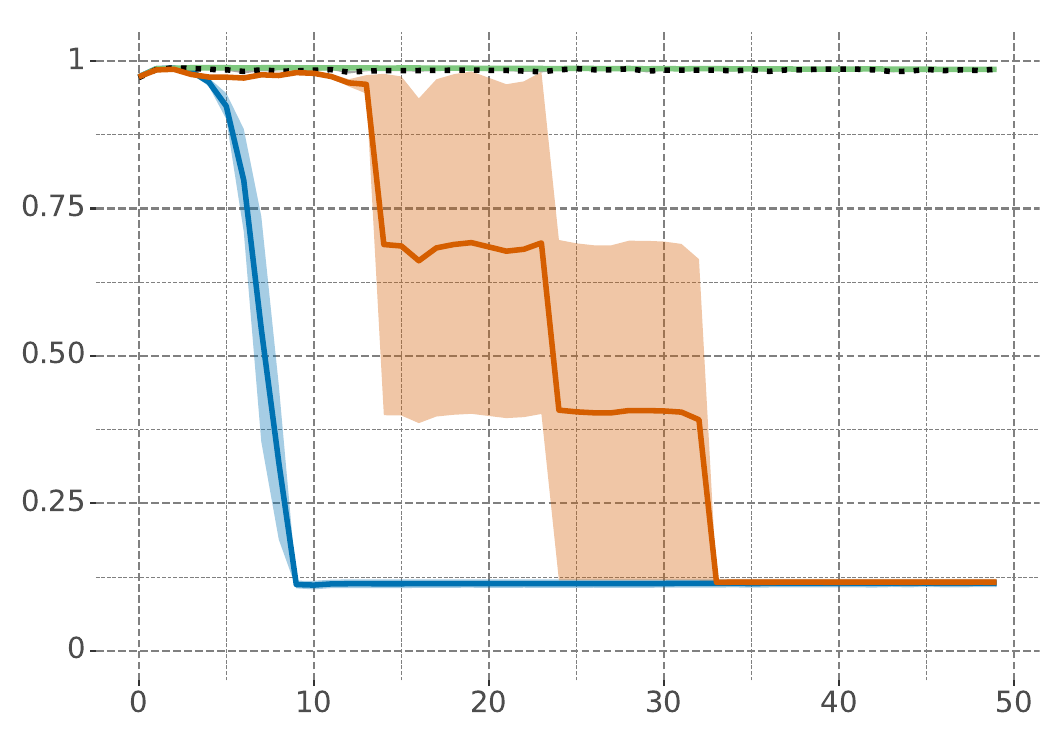}
                \vspace{-0.3cm} %
        \end{subfigure} &
        \begin{subfigure}{0.28\textwidth}
            \caption*{5+1 CIFAR} %
            \includegraphics[width=\linewidth]{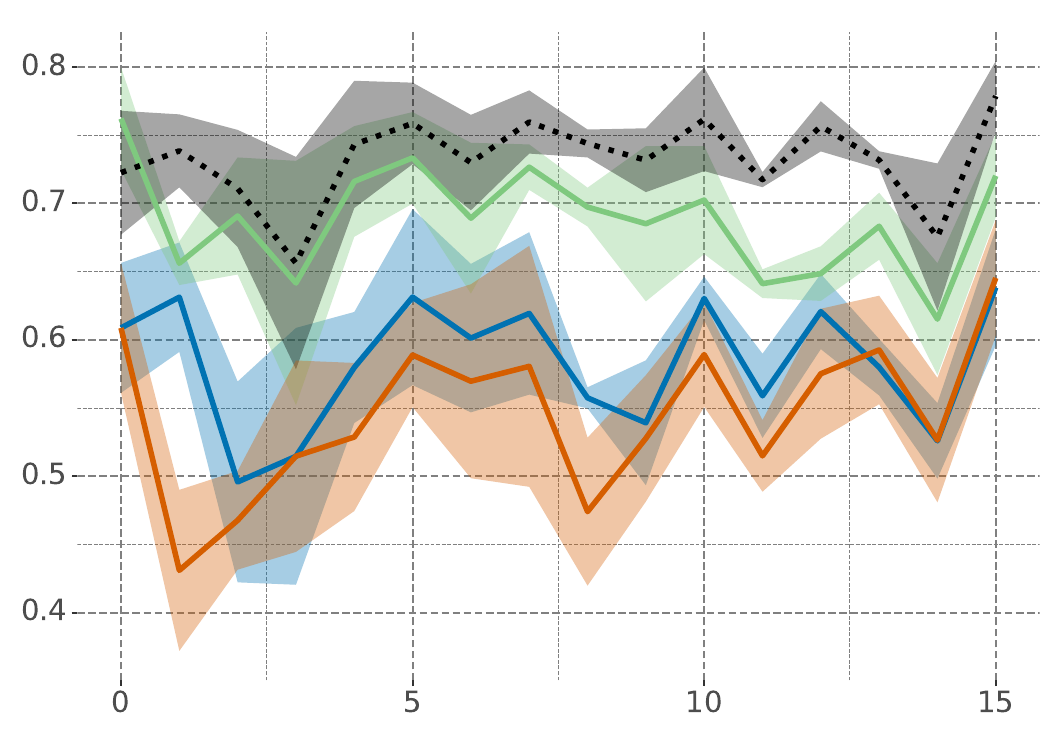}
                \vspace{-0.3cm} %
        \end{subfigure}
    \end{tabular}

\begin{tikzpicture}
    \draw [L2Init, thick, dotted, line width=2pt] (-2,0) -- (-1.5,0);
    \node[anchor=west] at (-1.5,0) {L2 Init};

    \draw [Baseline, thick, line width=2pt] (1,0) -- (1.5,0);
    \node[anchor=west] at (1.5,0) {Baseline};
    
    \draw [ConcatReLU, thick, line width=2pt] (4,0) -- (4.5,0);
    \node[anchor=west] at (4.5,0) {Concat ReLU};

    \draw [ReDO, thick, line width=2pt] (8,0) -- (8.5,0);
    \node[anchor=west] at (8.5,0) {ReDO};

\end{tikzpicture}
    \caption{Comparison of average online task accuracy on a subset of problems when using a wider network with the Adam optimizer. L2 Init consistently mitigates plasticity loss.}
    \label{fig:width}
\end{figure*}

\section{Conclusion}
Recently, multiple methods have been proposed for mitigating plasticity loss in continual learning. One common and quite successful category of methods is characterized by periodically re-initializing subsets of weights. However, resetting methods bring additional decisions to be made by the algorithm designer, such as which parameters to reinitialize and how often. In this paper, we propose a very simple alternative that we call \methodname. Concretely, we add a loss term that regularizes the parameters toward the initial parameters. This encourages parameters that have little influence on recent losses to drift toward initialization and therefore allows them to be recruited for future adaptation. This approach is similar to standard L2 regularization, but rather than regularizing toward the origin, we regularize toward the initial parameters, which ensures that the weight rank does not collapse. To evaluate \methodname, we perform an empirical study on continual supervised learning problems. 
Compared with other methods, \methodname most consistently maintains plasticity and performs similarly to Continual Backprop. 

We hope our method opens up avenues for future work on mitigating plasticity loss. 
In future work, it would be useful to evaluate \methodname on a broader set of problems \rebuttal{and more realistic environments}, including RL settings. It is possible that our method may need to be adjusted, for instance by using L1 instead of L2 regularization. Finally, this study has focused exclusively on maintaining plasticity, leaving aside the issue of forgetting. Designing methods that effectively balance the trade-off between maintaining plasticity and avoiding forgetting is an exciting avenue for future work.


\section*{Acknowledgements}
We thank Anmol Kagrecha and Wanqiao Xu for their feedback on an earlier version of this paper. Saurabh Kumar is supported by the Stanford Knight Hennessy Fellowship.

\bibliography{collas2024_conference}
\bibliographystyle{collas2024_conference}

\newpage
\appendix
\section{Appendix}
\subsection{Experiment Details}\label{appendix:experiments}

\subsubsection{\rebuttal{Compute Resources}}
\rebuttal{All experiments were run on a Google Cloud VM instance with 56 cores, allowing 56 training runs to be done in parallel. We did not use GPUs. A single training run on Permuted MNIST, Random Label CIFAR, Continual ImageNet, Random Label MNIST, and Random Label CIFAR took 10 minutes, 3 minutes, 20 minutes, 1 hour, and 2 hours to complete, respectively.}

\subsubsection{Problems}\label{appendix:environments}
Parameters for each of the five problems we consider are listed in Table~\ref{tab:problemparams}.

\begin{table}
\centering
\caption{Problem parameters.}
\begin{tabular}{ |p{5cm}||p{5cm}|  }
 \hline
 \multicolumn{2}{|c|}{Permuted MNIST} \\
 \hline
 Parameter & Value\\
 \hline
 dataset size per task  & 10,000 samples \\
 batch size & $16$ \\
 task duration &   $625$ timesteps ($1$ epoch) \\
 number of tasks & $500$ \\
 \hline
\end{tabular}
\\[10pt] %
\begin{tabular}{ |p{5cm}||p{5cm}|  }
 \hline
 \multicolumn{2}{|c|}{Random Label MNIST \& Random Label CIFAR} \\
 \hline
 Parameter & Value\\
 \hline
 dataset size per task  & $1200$ samples \\
 batch size & $16$ \\
 task duration &   30,000 timesteps ($400$ epochs) \\
 number of tasks & $50$ \\
 \hline
\end{tabular}
\\[10pt] %
\begin{tabular}{ |p{5cm}||p{5cm}|  }
 \hline
 \multicolumn{2}{|c|}{5+1 CIFAR} \\
 \hline
 Parameter & Value\\
 \hline
 dataset size per hard task   & $2500$ samples \\
 dataset size per easy task   & $500$ samples \\
 batch size & $32$ \\
 task duration &  780 timesteps \\
 number of tasks & $30$ ($15$ hard, $15$ easy) \\
 \hline
\end{tabular}
\\[10pt] %
\begin{tabular}{ |p{5cm}||p{5cm}|  }
 \hline
 \multicolumn{2}{|c|}{Continual ImageNet} \\
 \hline
 Parameter & Value\\
 \hline
 dataset size per task  & $1200$ samples \\
 batch size & $100$ \\
 task duration &   120 timesteps ($10$ epochs) \\
 number of tasks & $500$ \\
 \hline
\end{tabular}
\label{tab:problemparams}
\end{table}

\subsubsection{Agents}\label{appendix:agents}
\textbf{Neural network architectures.} For all agents, we used an MLP on Permuted MNIST and Random Label MNIST and a CNN on Random Label CIFAR, 5+1 CIFAR, and Continual ImageNet. We chose networks with small hidden layer width to study the setting in which plasticity loss is exacerbated due to capacity constraints. In particular, the neural network can achieve high average online task accuracy on a single task, or even a sequence of tasks, but when faced with a long sequence, plasticity loss occurs. The MLP and CNN architectures we use are as follows:
\begin{itemize}
    \item MLP: We use two hidden layers of width $100$ and ReLU activations.
    \item CNN: We use two convolutional layers followed by two fully-connected layers. The first convolutional layer uses kernel size $5 \times 5$ with $16$ output channels. This layer is followed by a max pool. The second also uses kernel size $5 \times 5$ with $16$ output channels and is also followed by a max pool. The fully-connected layers have widths $100$.
    \item All networks have a fully connected output layer at the end with $10$ outputs for Permuted MNIST, Random Label MNIST, and Random Label CIFAR, $100$ outputs for 5+1 CIFAR, and $2$ outputs for Continual ImageNet.
\end{itemize}

The exception to the above is Concat ReLU, for which we use a slightly smaller hidden size since otherwise Concat ReLU would have twice the number of parameters as all other agents. Specifically, we compute the smallest fraction of neurons to remove from each hidden layer such that the total number of parameters in the network is as least as large as the ones in the above architectures. These fractions are $0.09$ on Permuted MNIST and Random Label MNIST, $0.27$ on Random Label CIFAR and Continual ImageNet, and $0.31$ on 5+1 CIFAR.

\textbf{Hyper-parameters}
As described in Section $\ref{sec:experiments}$, for all agents on all problems, we performed a hyper-parameter sweep over $3$ seeds for each problem and optimizer combination. The optimal hyper-parameter configurations based on the total average online accuracy metric averaged across the $3$ seeds are listed in Tables~\ref{tab:agent-hyperparams1} and \ref{tab:agent-hyperparams2}.  We used these hyper-parameters with $10$ additional seeds to obtain all results.

\textit{Continual Backprop.} For Continual Backprop, we use the implementation in the public GitHub repository. We try two different methods for computing utility. The first one, called ``contribution," uses the inverse of the average weight magnitude as a measure of utility. The second one, ``adaptive-contribution," is the one proposed in~\citet{dohare2021continual} that also utilizes the activation magnitude multiplied by the outgoing weights. See ~\citet{dohare2021continual, dohare2023maintaining} (and the associated GitHub repository) for additional details. There was barely any difference in performance between the two utility types, so we present the results for the type presented in their paper. The other Continual Backprop hyper-parameter settings we use are those reported in~\citet{dohare2023maintaining}. In particular, we set the maturity threshold to be $100$ and the utility decay rate to be $0.99$. 

\subsection{Additional Results}\label{appendix:results}

\renewcommand\thesubfigure{\hspace{0.5cm}(\alph{subfigure})}
\captionsetup[subfigure]{labelformat=simple, labelsep=none}
\begin{figure*}[!htbp]

    \centering
        
        \begin{subfigure}{0.32\textwidth}
            \caption*{Permuted MNIST} %
            \includegraphics[width=\linewidth]{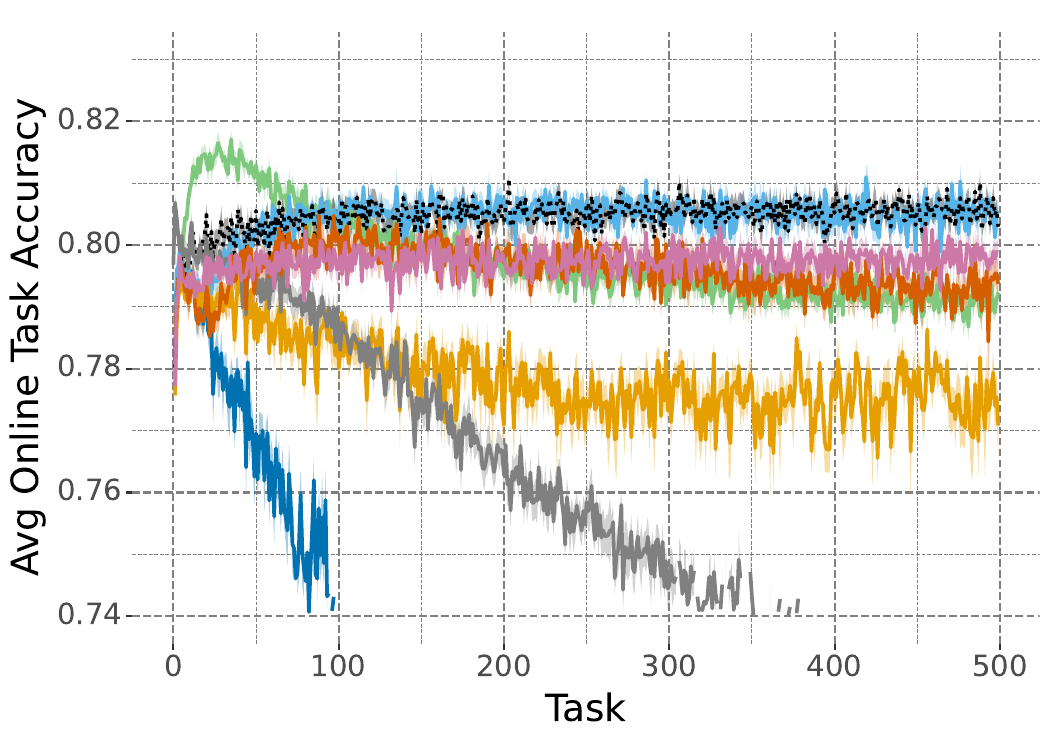}
            \vspace{-0.5cm} %
        \end{subfigure} 
        \begin{subfigure}{0.32\textwidth}
            \caption*{Random Label MNIST} %
            \includegraphics[width=\linewidth]{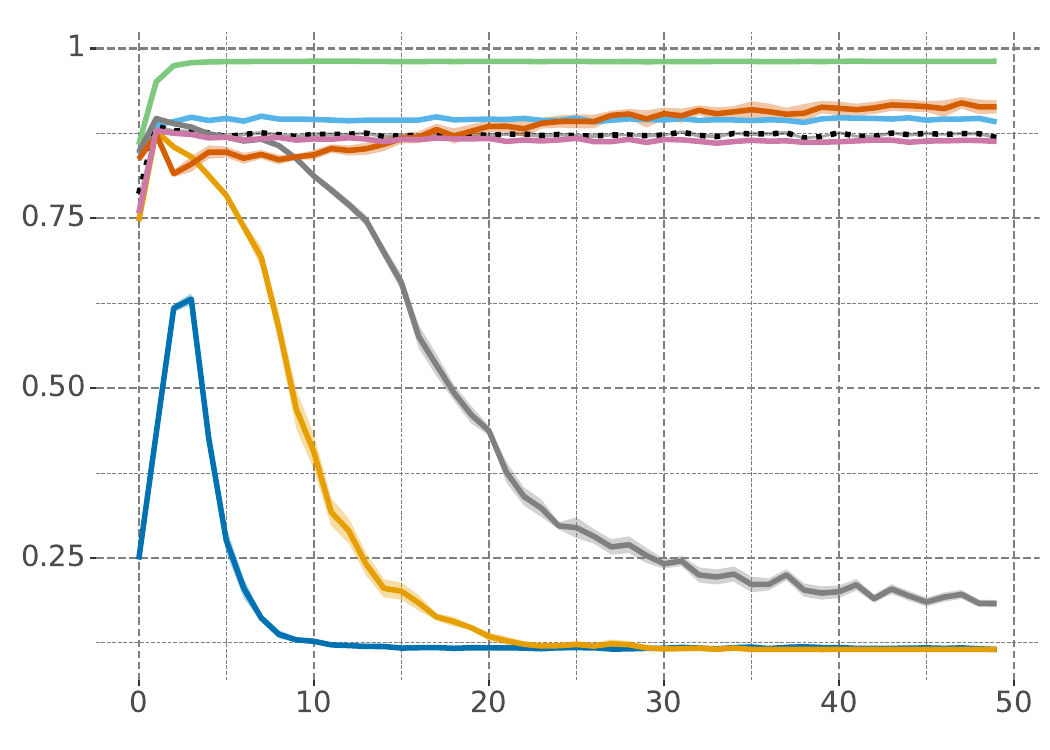}
            \vspace{-0.5cm} %
        \end{subfigure} 
        \begin{subfigure}{0.32\textwidth}
            \caption*{Random Label CIFAR} %
            \includegraphics[width=\linewidth]{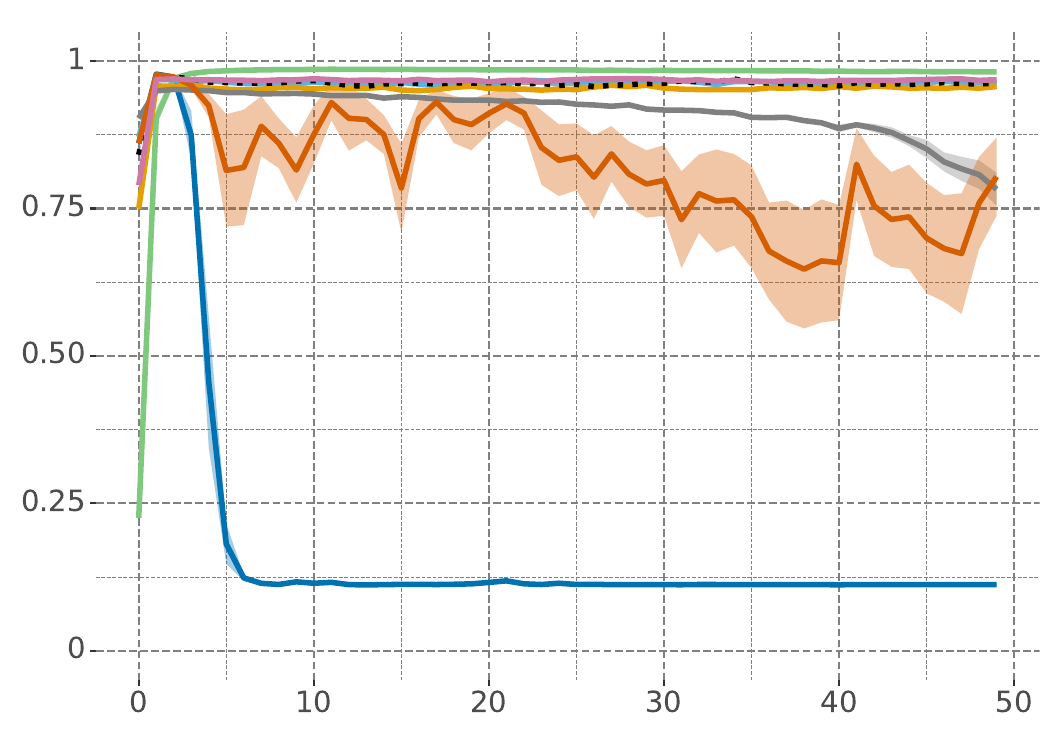}
            \vspace{-0.5cm} %
        \end{subfigure} \\

        \begin{subfigure}{0.32\textwidth}
            \caption*{5+1 CIFAR} %
            \includegraphics[width=\linewidth]{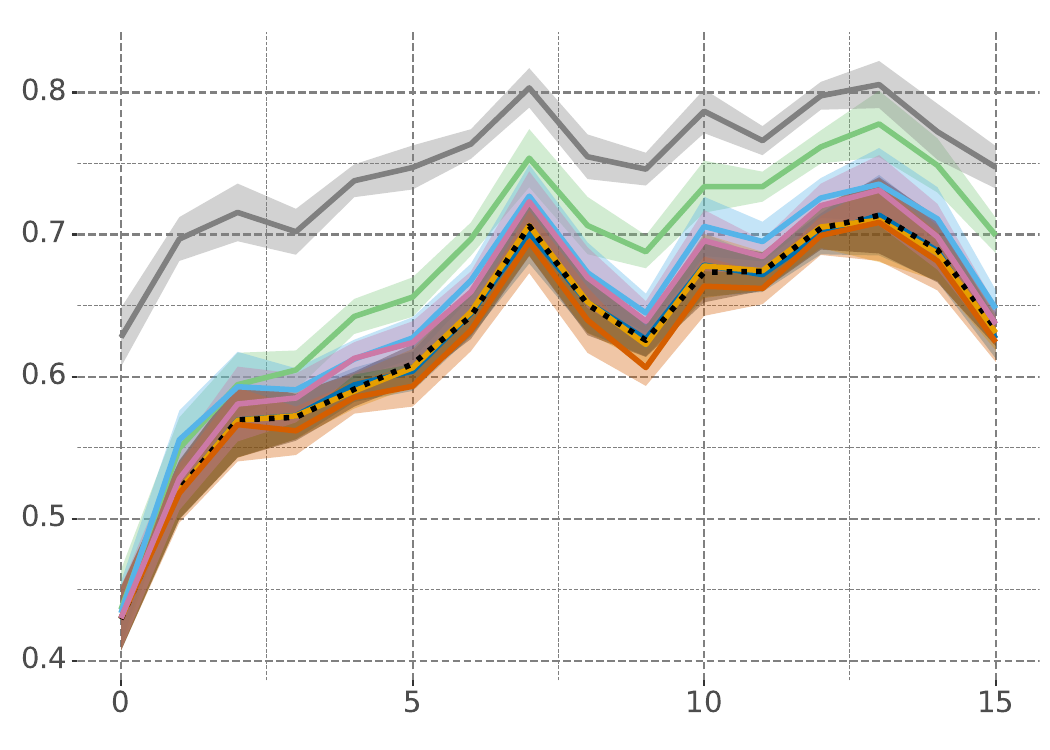}
            \vspace{-0.1cm} %
        \end{subfigure} 
        \begin{subfigure}{0.32\textwidth}
            \caption*{Continual ImageNet} %
            \includegraphics[width=\linewidth]{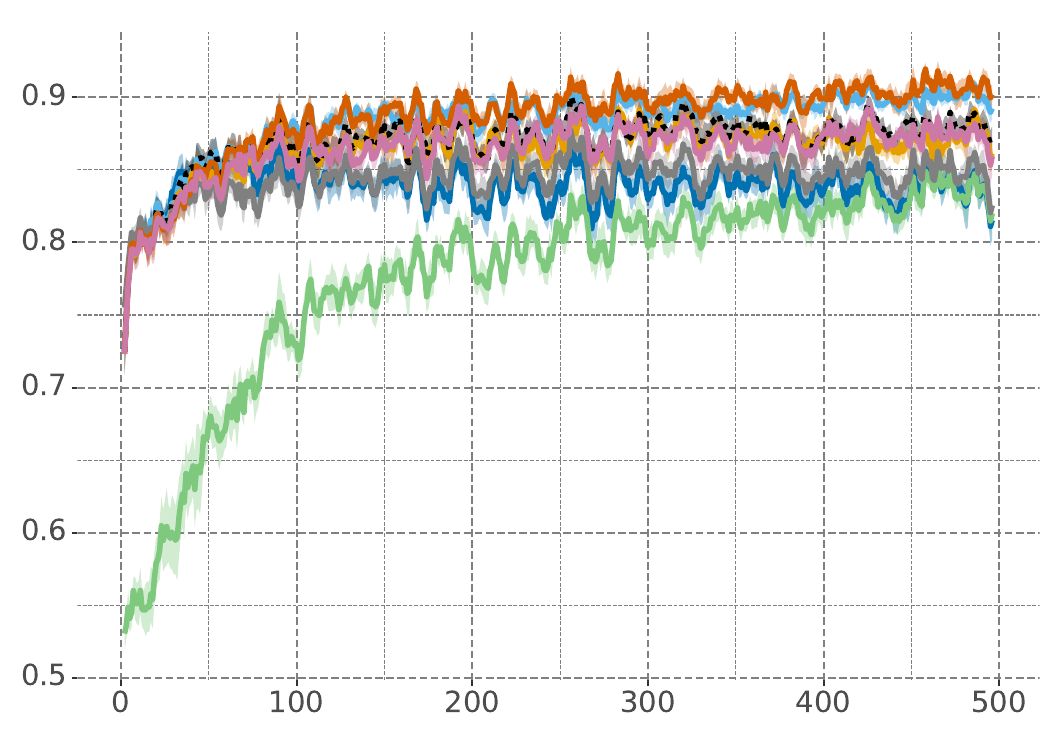}
            \vspace{-0.1cm} %
        \end{subfigure}

\begin{tikzpicture}

    \def\firstRowY{0.5}
    \draw [Baseline, thick, line width=2pt] (-2,\firstRowY) -- (-1.5,\firstRowY);
    \node[anchor=west] at (-1.5,\firstRowY) {Baseline};
    
    \draw [LayerNorm, thick, line width=2pt] (1,\firstRowY) -- (1.5,\firstRowY);
    \node[anchor=west] at (1.5,\firstRowY) {Layer Norm};

    \draw [ShrinkAndPerturb, thick, line width=2pt] (4,\firstRowY) -- (4.5,\firstRowY);
    \node[anchor=west] at (4.5,\firstRowY) {Shrink \& Perturb};

    \draw [ReDO, thick, line width=2pt] (8,\firstRowY) -- (8.5,\firstRowY);
    \node[anchor=west] at (8.5,\firstRowY) {ReDO};

    \draw [L2Init, thick, dotted, line width=2pt] (-2,0) -- (-1.5,0);
    \node[anchor=west] at (-1.5,0) {L2 Init};
    
    \draw [L2, thick, line width=2pt] (1,0) -- (1.5,0);
    \node[anchor=west] at (1.5,0) {L2};
    
    \draw [ContinualBackprop, thick, line width=2pt] (4,0) -- (4.5,0); 
    \node[anchor=west] at (4.5,0) {Continual Backprop};

    \draw [ConcatReLU, thick, line width=2pt] (8,0) -- (8.5,0);
    \node[anchor=west] at (8.5,0) {Concat ReLU};
\end{tikzpicture}
    
    \caption{Comparison of average online task accuracy across all five problems when using Vanilla SGD. \methodname consistently maintains plasticity, whereas L2 does not on Permuted MNIST and Random Label MNIST.}
    \label{fig:sgd-performance-comparison}
\end{figure*}

\subsubsection{Results with Vanilla SGD.}\label{appendix:sgd_results} 
When training agents with SGD, we sweep over $\alpha \in \{ 1\mathrm{e}{-2}, 1\mathrm{e}{-3} \}$. We additionally sweep over $\alpha=0.1$ on 5+1 CIFAR and Continual ImageNet. As a baseline agent, we run vanilla incremental SGD with constant stepsize.

Compared to when using Adam, there is less plasticity loss when using SGD, as shown in Figure~\ref{fig:sgd-performance-comparison}. \methodname performs similarly to Continual Backprop and consistently mitigates plasticity on problems on which it occurs. In contrast, L2 does not on Permuted MNIST and Random Label MNIST. \methodname also performs similarly to ReDO, although ReDO's performance has larger variation between seeds. Concat ReLU performs well across problems but loses plasticity on Permuted MNIST and has lower performance on Continual ImageNet. Unlike when using Adam, L2 Init does not outperform all methods on 5+1 CIFAR. Instead, Layer Norm performs the best on this problem.

\subsubsection{Test Accuracy}\label{appendix:test_accuracy}

\renewcommand\thesubfigure{\hspace{0.5cm}(\alph{subfigure})}
\captionsetup[subfigure]{labelformat=simple, labelsep=none}
\begin{figure*}

    \centering
        
        \begin{subfigure}{0.32\textwidth}
            \caption*{Permuted MNIST} %
            \includegraphics[width=\linewidth]{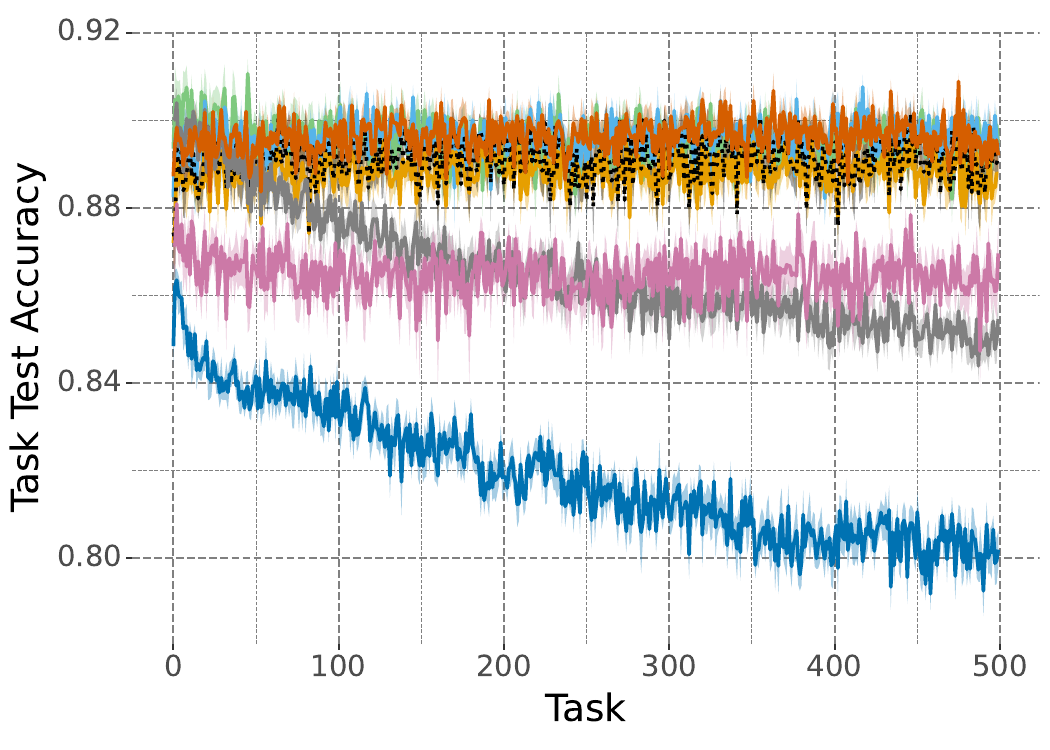}
            \vspace{-0.5cm} %
        \end{subfigure} 
        \begin{subfigure}{0.32\textwidth}
            \caption*{5+1 CIFAR} %
            \includegraphics[width=\linewidth]{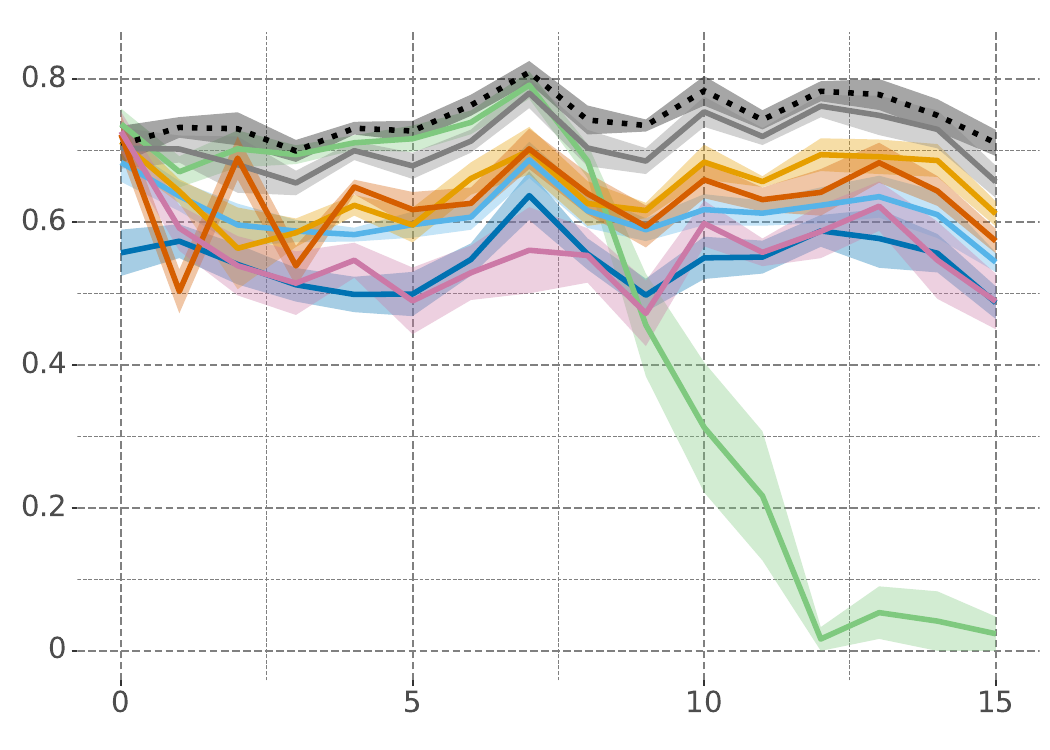}
            \vspace{-0.1cm} %
        \end{subfigure} 
        \begin{subfigure}{0.32\textwidth}
            \caption*{Continual ImageNet} %
            \includegraphics[width=\linewidth]{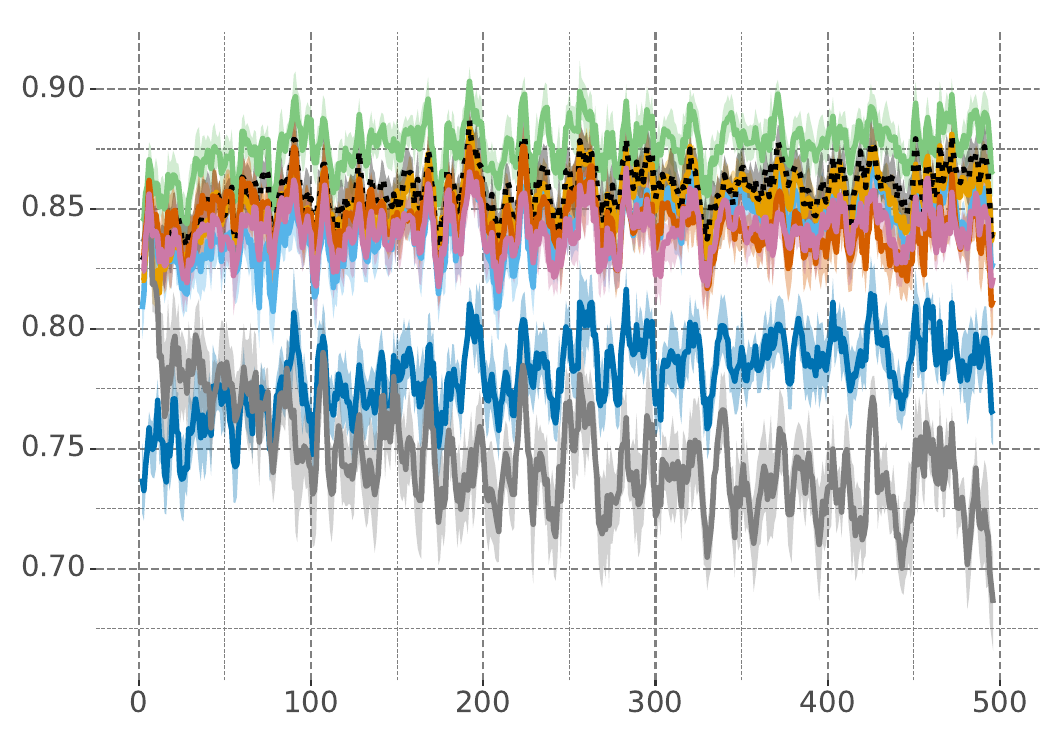}
            \vspace{-0.1cm} %
        \end{subfigure}

\begin{tikzpicture}

    \def\firstRowY{0.5}
    \draw [Baseline, thick, line width=2pt] (-2,\firstRowY) -- (-1.5,\firstRowY);
    \node[anchor=west] at (-1.5,\firstRowY) {Baseline};
    
    \draw [LayerNorm, thick, line width=2pt] (1,\firstRowY) -- (1.5,\firstRowY);
    \node[anchor=west] at (1.5,\firstRowY) {Layer Norm};

    \draw [ShrinkAndPerturb, thick, line width=2pt] (4,\firstRowY) -- (4.5,\firstRowY);
    \node[anchor=west] at (4.5,\firstRowY) {Shrink \& Perturb};

    \draw [ReDO, thick, line width=2pt] (8,\firstRowY) -- (8.5,\firstRowY);
    \node[anchor=west] at (8.5,\firstRowY) {ReDO};

    \draw [L2Init, thick, dotted, line width=2pt] (-2,0) -- (-1.5,0);
    \node[anchor=west] at (-1.5,0) {L2 Init};
    
    \draw [L2, thick, line width=2pt] (1,0) -- (1.5,0);
    \node[anchor=west] at (1.5,0) {L2};
    
    \draw [ContinualBackprop, thick, line width=2pt] (4,0) -- (4.5,0); 
    \node[anchor=west] at (4.5,0) {Continual Backprop};

    \draw [ConcatReLU, thick, line width=2pt] (8,0) -- (8.5,0);
    \node[anchor=west] at (8.5,0) {Concat ReLU};
\end{tikzpicture}
    
    \caption{Accuracy computed on held out task test data at the end of each task when training all agents with Adam. \methodname consistently maintains plasticity and performs similarly to the other resetting methods Continual Backprop and ReDO. Concat ReLU performs well on Continual ImageNet but poorly on 5+1 CIFAR.}
    \label{fig:adam-test-acc-comparison}
\end{figure*}

\renewcommand\thesubfigure{\hspace{0.5cm}(\alph{subfigure})}
\captionsetup[subfigure]{labelformat=simple, labelsep=none}
\begin{figure*}

    \centering
        
        \begin{subfigure}{0.32\textwidth}
            \caption*{Permuted MNIST} %
            \includegraphics[width=\linewidth]{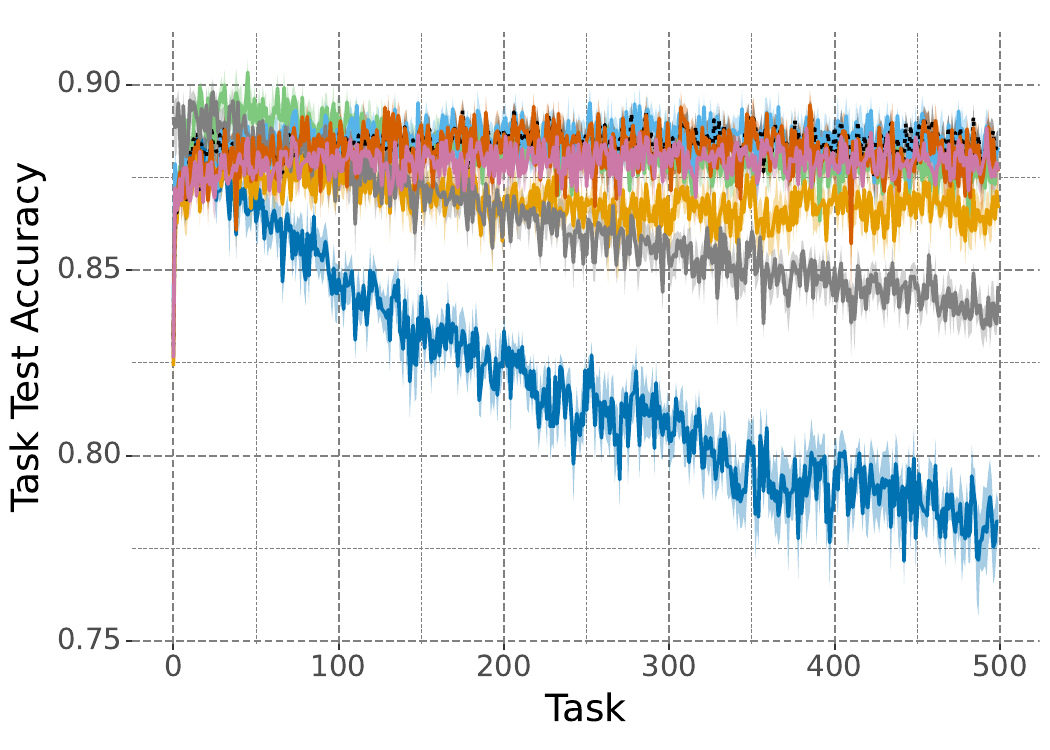}
            \vspace{-0.5cm} %
        \end{subfigure} 
        \begin{subfigure}{0.32\textwidth}
            \caption*{5+1 CIFAR} %
            \includegraphics[width=\linewidth]{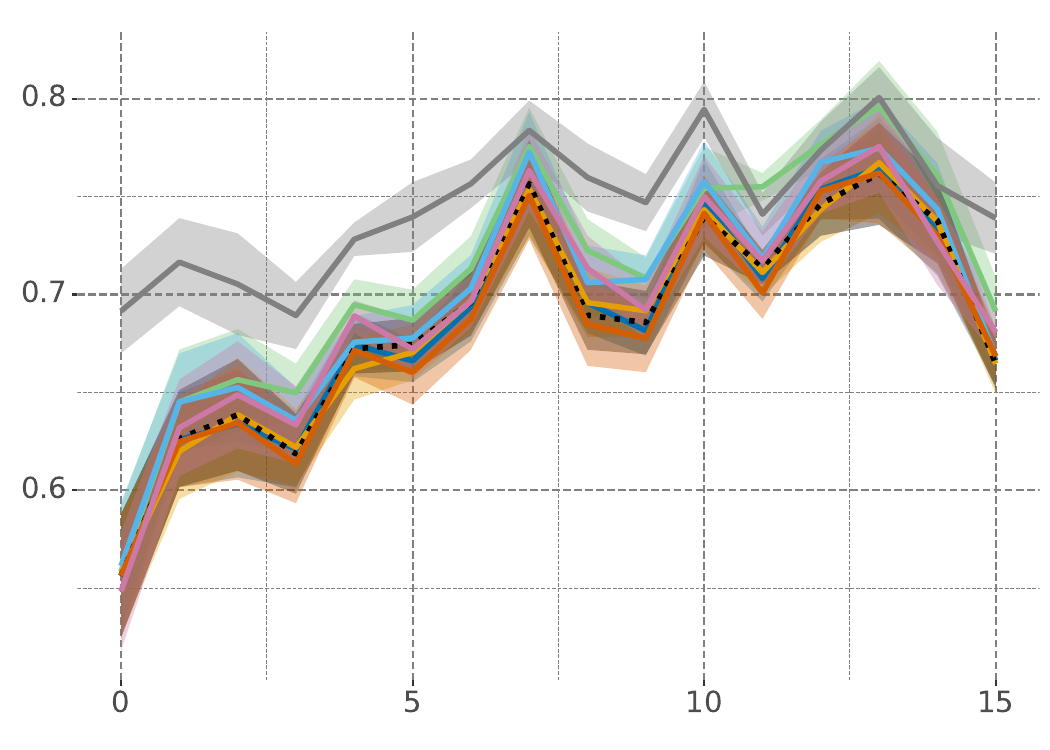}
            \vspace{-0.1cm} %
        \end{subfigure} 
        \begin{subfigure}{0.32\textwidth}
            \caption*{Continual ImageNet} %
            \includegraphics[width=\linewidth]{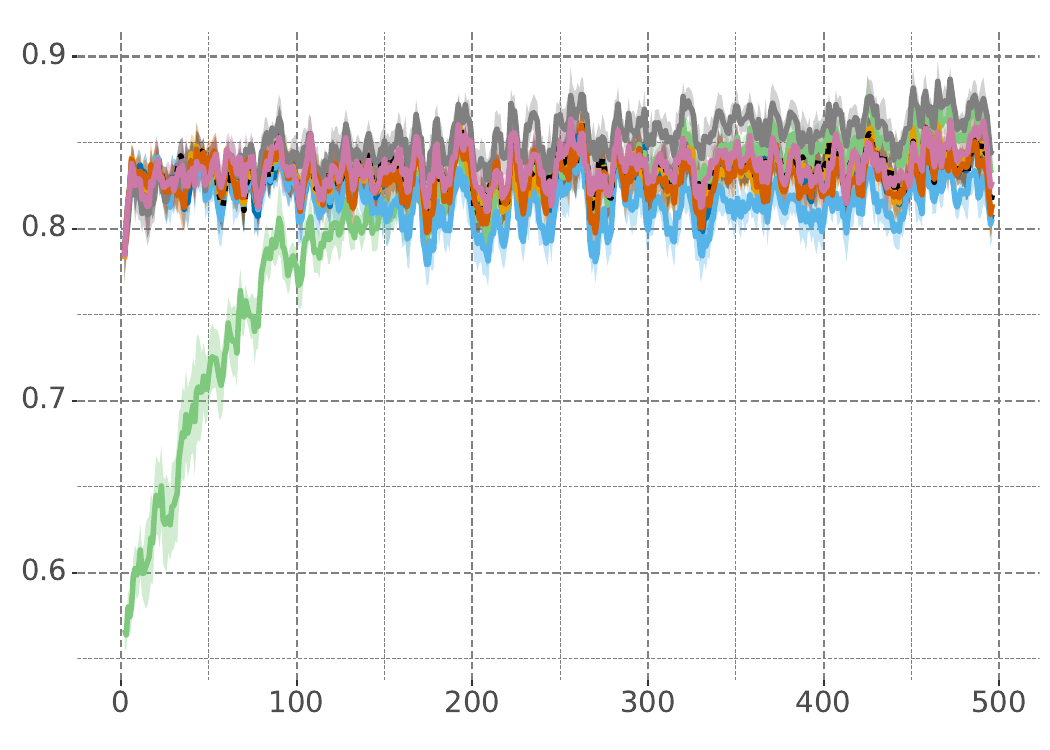}
            \vspace{-0.1cm} %
        \end{subfigure}

\begin{tikzpicture}

    \def\firstRowY{0.5}
    \draw [Baseline, thick, line width=2pt] (-2,\firstRowY) -- (-1.5,\firstRowY);
    \node[anchor=west] at (-1.5,\firstRowY) {Baseline};
    
    \draw [LayerNorm, thick, line width=2pt] (1,\firstRowY) -- (1.5,\firstRowY);
    \node[anchor=west] at (1.5,\firstRowY) {Layer Norm};

    \draw [ShrinkAndPerturb, thick, line width=2pt] (4,\firstRowY) -- (4.5,\firstRowY);
    \node[anchor=west] at (4.5,\firstRowY) {Shrink \& Perturb};

    \draw [ReDO, thick, line width=2pt] (8,\firstRowY) -- (8.5,\firstRowY);
    \node[anchor=west] at (8.5,\firstRowY) {ReDO};

    \draw [L2Init, thick, dotted, line width=2pt] (-2,0) -- (-1.5,0);
    \node[anchor=west] at (-1.5,0) {L2 Init};
    
    \draw [L2, thick, line width=2pt] (1,0) -- (1.5,0);
    \node[anchor=west] at (1.5,0) {L2};
    
    \draw [ContinualBackprop, thick, line width=2pt] (4,0) -- (4.5,0); 
    \node[anchor=west] at (4.5,0) {Continual Backprop};

    \draw [ConcatReLU, thick, line width=2pt] (8,0) -- (8.5,0);
    \node[anchor=west] at (8.5,0) {Concat ReLU};
\end{tikzpicture}
    
    \caption{Accuracy computed on held out task test data at the end of each task when training all agents with Vanilla SGD. While the results are mixed, L2 Init maintains good performance whereas L2 performs poorly on Permuted MNIST.}
    \label{fig:sgd-test-acc-comparison}
\end{figure*}

On problems which have test datasets (Permuted MNIST, 5+1 CIFAR, and Continual ImageNet), we additionally plot the test accuracy on each task in Figures \ref{fig:adam-test-acc-comparison} and \ref{fig:sgd-test-acc-comparison}. Specifically, at the end of each task, we compute the accuracy on the test data for that task. The generalization performance of L2 Init is consistently similar to that of the other resetting methods Continual Backprop and ReDO.

\subsubsection{Robustness to Network Depth}\label{appendix:depth}
\renewcommand\thesubfigure{\hspace{0.5cm}(\alph{subfigure})}
\captionsetup[subfigure]{labelformat=simple, labelsep=none, font=footnotesize}
\begin{figure*}

    \centering
    \setlength{\tabcolsep}{2pt}
    
\begin{tabular}{>{\centering\arraybackslash}m{0.2cm}>{\centering\arraybackslash}m{0.32\textwidth}>{\centering\arraybackslash}m{0.32\textwidth}>{\centering\arraybackslash}m{0.32\textwidth}} \\
         \rotatebox[origin=c]{90}{} \vspace{0.5cm} & 
        \begin{subfigure}{0.28\textwidth}
            \caption*{Permuted MNIST} %
            \includegraphics[width=\linewidth]{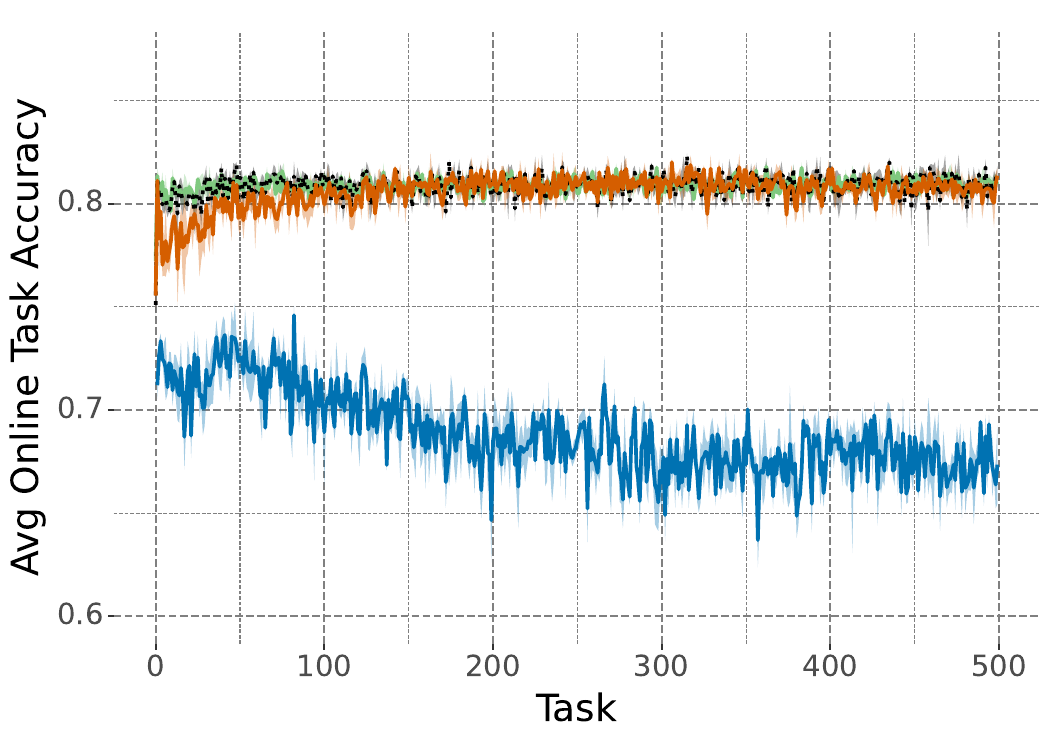}
                \vspace{-0.3cm} %
        \end{subfigure} &
        \begin{subfigure}{0.28\textwidth}
            \caption*{Random Label CIFAR} %
            \includegraphics[width=\linewidth]{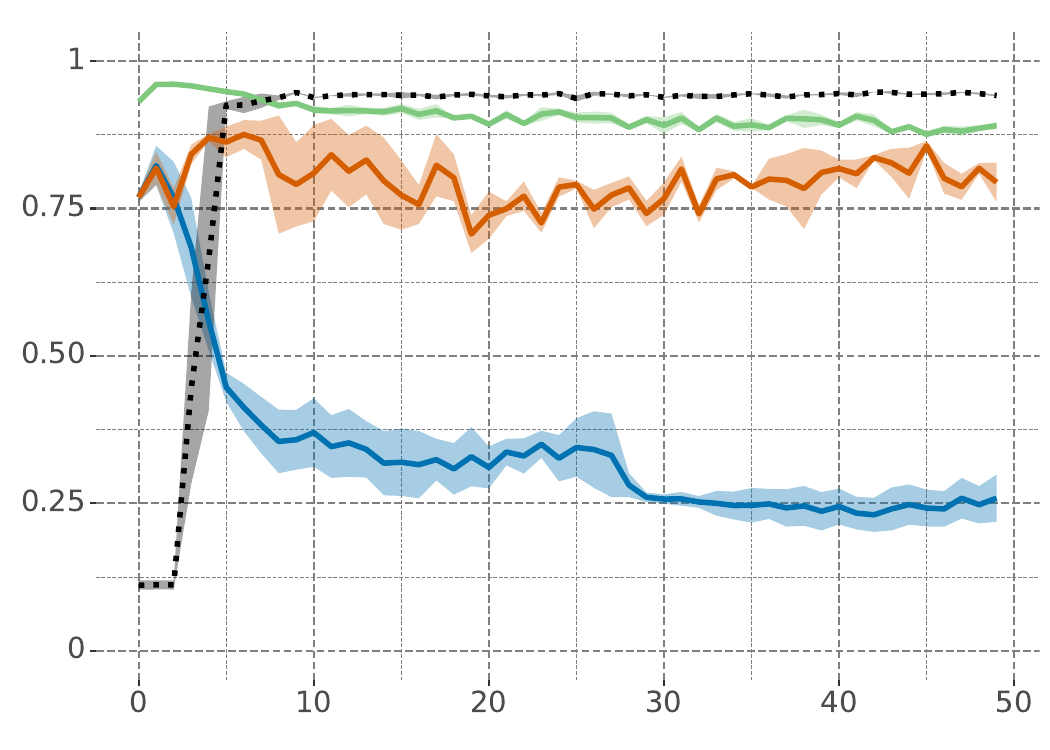}
                \vspace{-0.3cm} %
        \end{subfigure} &
        \begin{subfigure}{0.28\textwidth}
            \caption*{5+1 CIFAR} %
            \includegraphics[width=\linewidth]{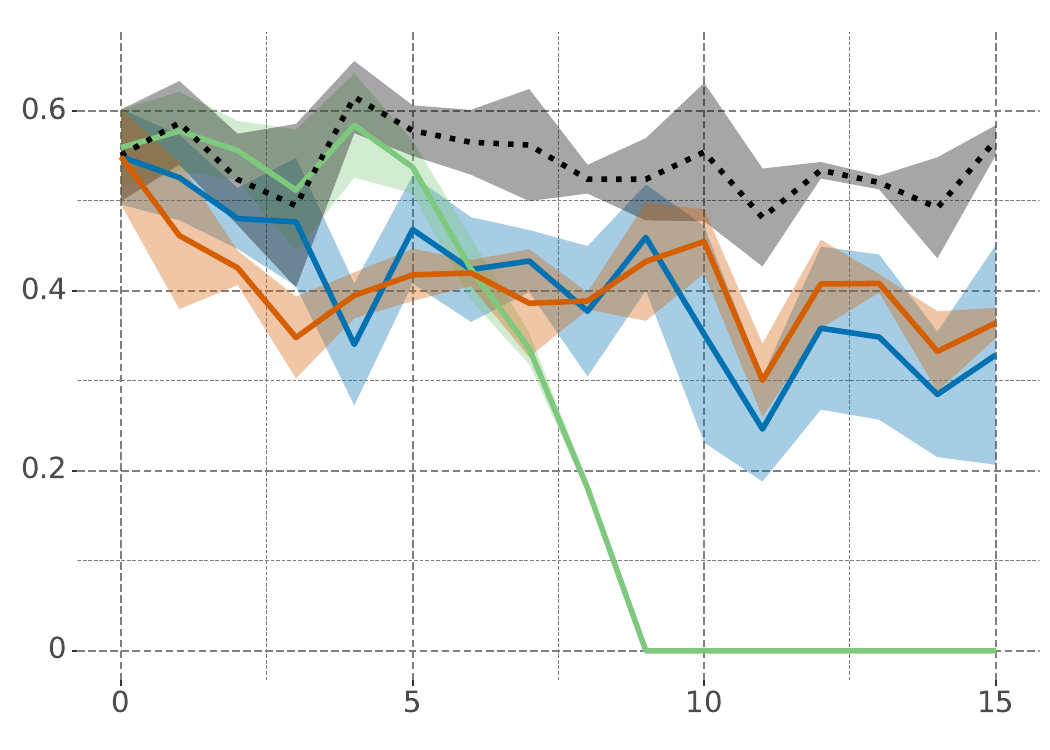}
                \vspace{-0.3cm} %
        \end{subfigure}
    \end{tabular}

\begin{tikzpicture}
    \draw [L2Init, thick, dotted, line width=2pt] (-2,0) -- (-1.5,0);
    \node[anchor=west] at (-1.5,0) {L2 Init};

    \draw [Baseline, thick, line width=2pt] (1,0) -- (1.5,0);
    \node[anchor=west] at (1.5,0) {Baseline};
    
    \draw [ConcatReLU, thick, line width=2pt] (4,0) -- (4.5,0);
    \node[anchor=west] at (4.5,0) {Concat ReLU};

    \draw [ReDO, thick, line width=2pt] (8,0) -- (8.5,0);
    \node[anchor=west] at (8.5,0) {ReDO};

\end{tikzpicture}
    
    \caption{Comparison of average online task accuracy on a subset of problems when using a deeper network with the Adam optimizer. As when using a shallower network, L2 Init consistently mitigates plasticity loss.}
    \label{fig:depth}
\end{figure*}

To determine whether L2 Init remains effective when using deeper networks, we evaluate L2 Init's performance on a subset of problems when using networks with two additional hidden layers.
We also evaluate Concat ReLU, ReDO, and the Baseline agent. We find that L2 Init's effectiveness is not diminished by increased network depth, and the other methods also perform similarly as with a shallower network, as shown in Figure~\ref{fig:depth}.

The deeper neural network architectures used in the experiments in Figure~\ref{fig:depth} are described below.
\begin{itemize}
    \item MLP: We use four hidden layers of width $100$ and ReLU activations. This adds two additional hidden layers to the previous MLP architecture used.
    \item CNN: We use three convolutional layers followed by three fully-connected layers. The first convolutional layer uses kernel size $5 \times 5$ with $16$ output channels. This layer is followed by a max pool. The second also uses kernel size $3 \times 3$ with $16$ output channels and is also followed by a max pool. The final convolutional layeer uses kernel size $3 \times 3$ with $16$ output channels. The fully-connected layers have widths $100$. Note that we use kernel size $3 \times 3$ for the second convolutional layer instead of $5 \times 5$ as in the previous CNN architecture. This is to prevent the feature map dimensions from becoming too small after the third convolutional layer is applied.
\end{itemize}

\subsubsection{\rebuttal{Sensitivity to Initialization Scheme}}

\rebuttal{In this section, we evaluate the sensitivity of L2 Init to initialization scheme. The PyTorch default initialization samples each weight and bias in layer $l$ from the uniform distribution $\mathcal{U}(\frac{-1}{\text{fan\_in}(l)}, \frac{1}{\text{fan\_in}(l)})$ where $\text{fan\_in}(l)$ is the input dimension for layer $l$.}

\rebuttal{We experiment with four additional initialization schemes:}
\begin{itemize}
    \item \rebuttal{Kaiming Uniform~\citep{he2015delving}: Samples each weight in layer $l$ from the uniform distribution $\mathcal{U}(-\sqrt{\frac{6}{\text{fan\_in}(l)}}, \sqrt{\frac{6}{\text{fan\_in}(l)}})$.}
    \item \rebuttal{Kaiming Normal~\citep{he2015delving}: Samples each weight in layer $l$ from the Normal distribution $\mathcal{N}(0, \sigma^2)$, where $\sigma = \sqrt{\frac{2}{\text{fan\_in}(l)}}$.}
    \item \rebuttal{Xavier Uniform~\citep{glorot2010understanding}: Samples each weight in layer $l$ from the uniform distribution $\mathcal{U}(-\sqrt{\frac{6}{\text{fan\_in}(l) + \text{fan\_out}(l)}}, \sqrt{\frac{6}{\text{fan\_in}(l) + \text{fan\_out}(l)}})$.}
    \item \rebuttal{Xavier Normal~\citep{glorot2010understanding}: Samples each weight in layer $l$ from the Normal distribution $\mathcal{N}(0, \sigma^2)$, where $\sigma = \sqrt{\frac{2}{\text{fan\_in}(l) + \text{fan\_out}(l)}}$.}
\end{itemize}
\rebuttal{These are all standard options in PyTorch for initializing neural network weights. For the above four initialization schemes, we initialize all biases to be $0$.}

\begin{figure*}

    \centering
    \setlength{\tabcolsep}{2pt}
    
\begin{tabular}{>{\centering\arraybackslash}m{0.2cm}>{\centering\arraybackslash}m{0.32\textwidth}>{\centering\arraybackslash}m{0.32\textwidth}>{\centering\arraybackslash}m{0.32\textwidth}} \\
         \rotatebox[origin=c]{90}{} \vspace{0.5cm} & 
        \begin{subfigure}{0.28\textwidth}
            \caption*{Permuted MNIST} %
            \includegraphics[width=\linewidth]{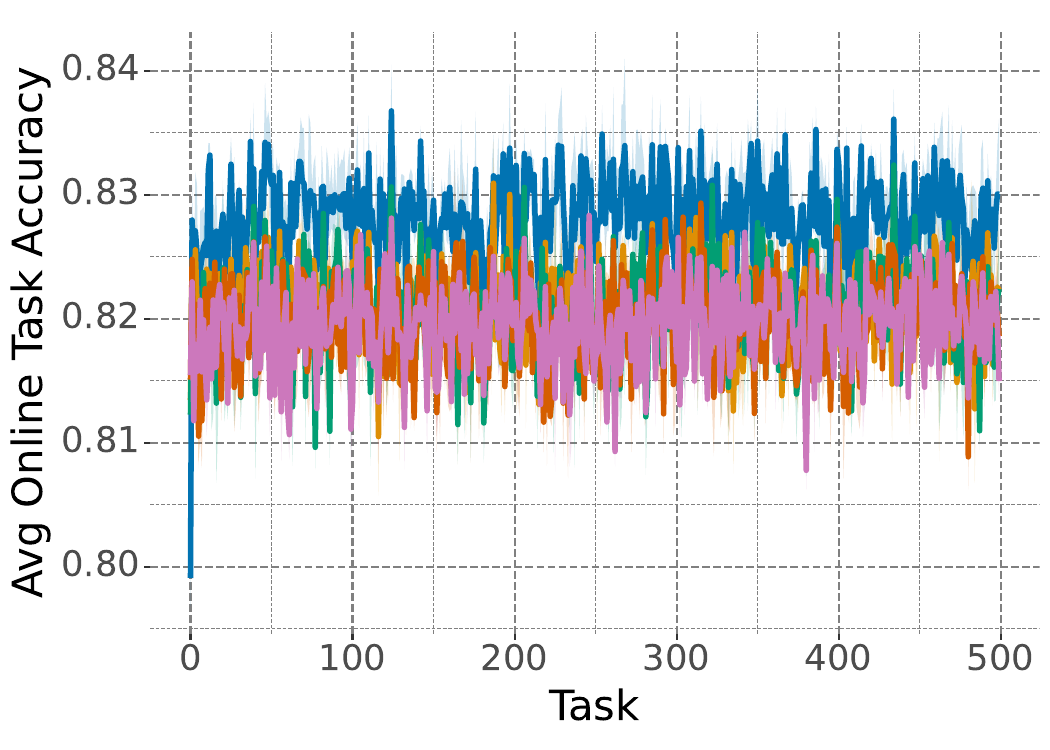}
                \vspace{-0.3cm} %
        \end{subfigure} &
        \begin{subfigure}{0.28\textwidth}
            \caption*{Random Label CIFAR} %
            \includegraphics[width=\linewidth]{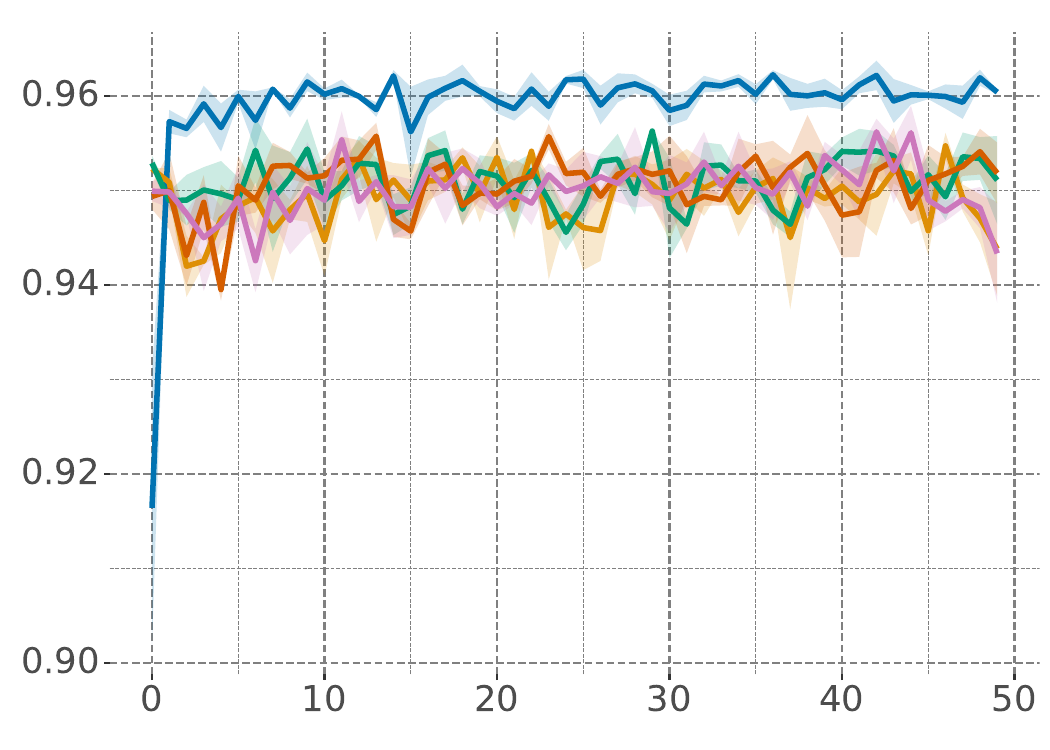}
                \vspace{-0.3cm} %
        \end{subfigure} &
        \begin{subfigure}{0.28\textwidth}
            \caption*{5+1 CIFAR} %
            \includegraphics[width=\linewidth]{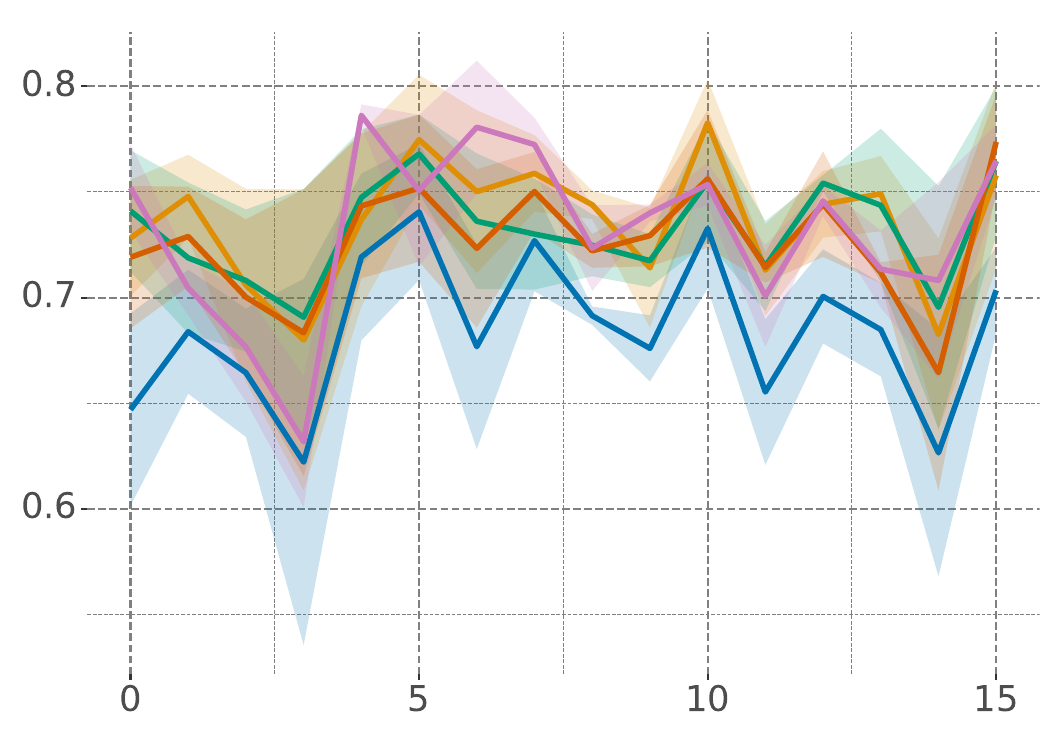}
                \vspace{-0.3cm} %
        \end{subfigure}
    \end{tabular}

\begin{tikzpicture}
    \draw [PyTorch_Default, thick, line width=2pt] (-2,0) -- (-1.5,0);
    \node[anchor=west] at (-1.5,0) {PyTorch Default};

    \draw [Kaiming_Normal, thick, line width=2pt] (1.5,0) -- (2,0);
    \node[anchor=west] at (2,0) {Kaiming Normal};
    
    \draw [Kaiming_Uniform, thick, line width=2pt] (5,0) -- (5.5,0);
    \node[anchor=west] at (5.5,0) {Kaiming Uniform};

    \draw [Xavier_Normal, thick, line width=2pt] (8.5,0) -- (9,0);
    \node[anchor=west] at (9,0) {Xavier Normal};

    \draw [Xavier_Uniform, thick, line width=2pt] (11.5,0) -- (12,0);
    \node[anchor=west] at (12,0) {Xavier Uniform};

\end{tikzpicture}
    
    \caption{\rebuttal{Sensitivity of L2 Init to different initialization schemes when using the Adam optimizer. The performance of L2 Init is better with the PyTorch Default initialization relative to other initialization schemes. However, L2 Init consistently mitigates plasticity loss regardless of initialization scheme.}}
    \label{fig:l2-init-sensitivity}
\end{figure*}

\begin{figure*}

    \centering
    \setlength{\tabcolsep}{2pt}
    
\begin{tabular}{>{\centering\arraybackslash}m{0.2cm}>{\centering\arraybackslash}m{0.32\textwidth}>{\centering\arraybackslash}m{0.32\textwidth}>{\centering\arraybackslash}m{0.32\textwidth}} \\
         \rotatebox[origin=c]{90}{} \vspace{0.5cm} & 
        \begin{subfigure}{0.28\textwidth}
            \caption*{Permuted MNIST} %
            \includegraphics[width=\linewidth]{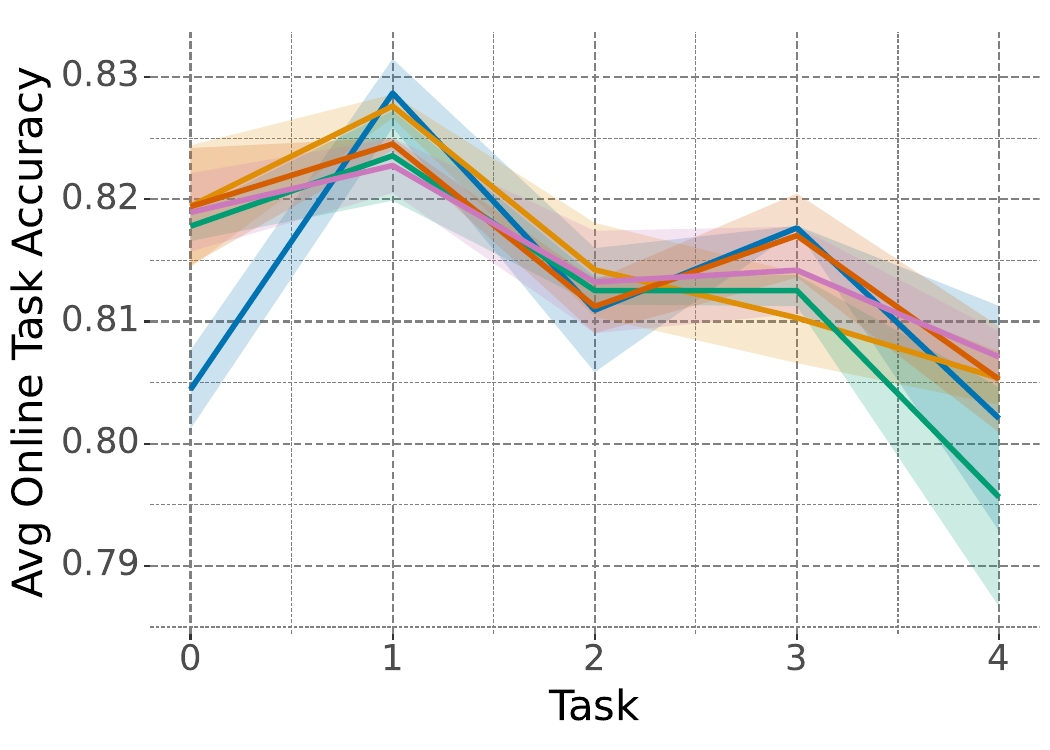}
                \vspace{-0.3cm} %
        \end{subfigure} &
        \begin{subfigure}{0.28\textwidth}
            \caption*{Random Label CIFAR} %
            \includegraphics[width=\linewidth]{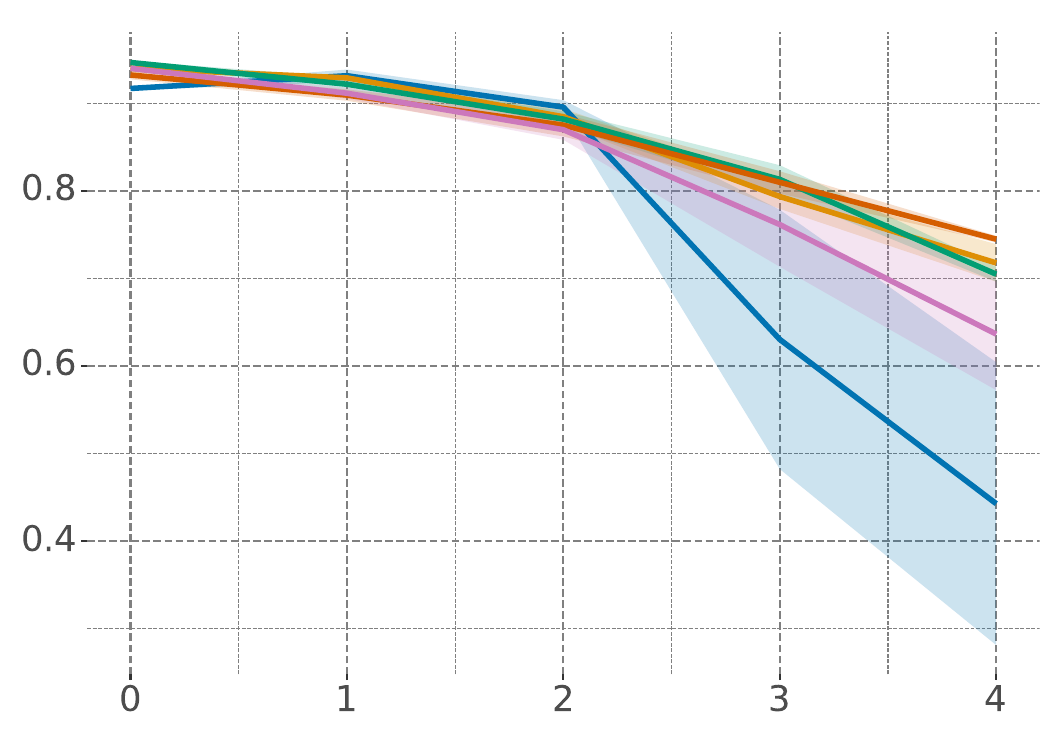}
                \vspace{-0.3cm} %
        \end{subfigure} &
        \begin{subfigure}{0.28\textwidth}
            \caption*{5+1 CIFAR} %
            \includegraphics[width=\linewidth]{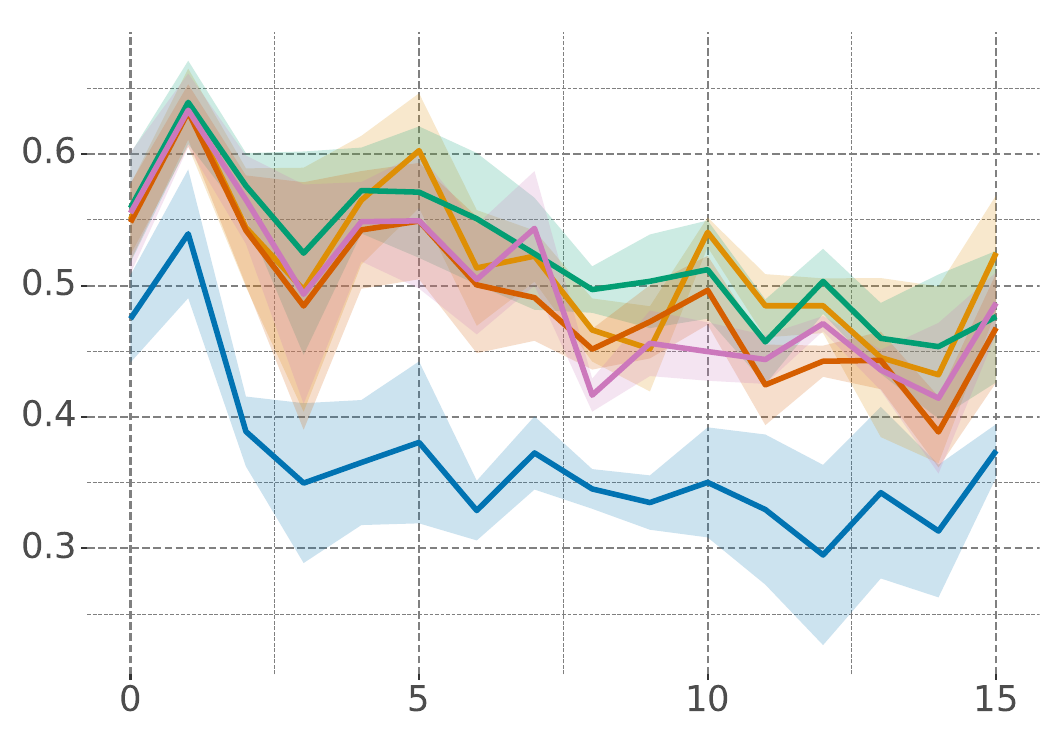}
                \vspace{-0.3cm} %
        \end{subfigure}
    \end{tabular}

\begin{tikzpicture}
    \draw [PyTorch_Default, thick, line width=2pt] (-2,0) -- (-1.5,0);
    \node[anchor=west] at (-1.5,0) {PyTorch Default};

    \draw [Kaiming_Normal, thick, line width=2pt] (1.5,0) -- (2,0);
    \node[anchor=west] at (2,0) {Kaiming Normal};
    
    \draw [Kaiming_Uniform, thick, line width=2pt] (5,0) -- (5.5,0);
    \node[anchor=west] at (5.5,0) {Kaiming Uniform};

    \draw [Xavier_Normal, thick, line width=2pt] (8.5,0) -- (9,0);
    \node[anchor=west] at (9,0) {Xavier Normal};

    \draw [Xavier_Uniform, thick, line width=2pt] (11.5,0) -- (12,0);
    \node[anchor=west] at (12,0) {Xavier Uniform};

\end{tikzpicture}
    
    \caption{\rebuttal{Sensitivity of the Baseline (Adam optimizer) to different initialization schemes. We evaluate on the first 5 tasks of Permuted MNIST and Random Label CIFAR. The performance with PyTorch default initialization on Random Label CIFAR and 5+1 CIFAR is worse than with other initialization schemes.}}
    \label{fig:baseline-init-sensitivity}
\end{figure*}

\rebuttal{In Figure~\ref{fig:l2-init-sensitivity}, we compare the performance of L2 Init with different initialization schemes. We do so using the Adam optimizer and run experiments on Permuted MNIST, Random Label CIFAR, and 5+1 CIFAR. Across all initialization schemes, L2 Init mitigates plasticity loss. However, we find that the performance of L2 Init is better when using the PyTorch default initialization as compared to when using the other four initialization schemes. To better understand what drives this difference, in Figure~\ref{fig:baseline-init-sensitivity} we also examine the performance of the Baseline agent (just using the Adam optimizer) over the initial set of tasks on Permuted MNIST and Random Label CIFAR before significant plasticity loss occurs. While the results are not conclusive, we find that the training dynamics are different when using the PyTorch default initialization scheme versus when using the other four initialization schemes.}

\renewcommand\thesubfigure{\hspace{0.5cm}(\alph{subfigure})}
\captionsetup[subfigure]{labelformat=simple, labelsep=none, font=footnotesize}

\renewcommand\thesubfigure{\hspace{0.5cm}(\alph{subfigure})}
\captionsetup[subfigure]{labelformat=simple, labelsep=none, font=footnotesize}

\subsubsection{\rebuttal{Sensitivity to Regularization Strength}}
\renewcommand\thesubfigure{\hspace{0.5cm}(\alph{subfigure})}
\captionsetup[subfigure]{labelformat=simple, labelsep=none, font=footnotesize}
\begin{figure*}

    \centering
    \setlength{\tabcolsep}{2pt}
    
\begin{tabular}{>{\centering\arraybackslash}m{0.2cm}>{\centering\arraybackslash}m{0.3\textwidth}>{\centering\arraybackslash}m{0.3\textwidth}>{\centering\arraybackslash}m{0.3\textwidth}} \\

         \rotatebox[origin=c]{90}{} \vspace{0.5cm} & 
        \begin{subfigure}{0.31\textwidth}
            \caption*{Permuted MNIST} %
            \includegraphics[width=\linewidth]{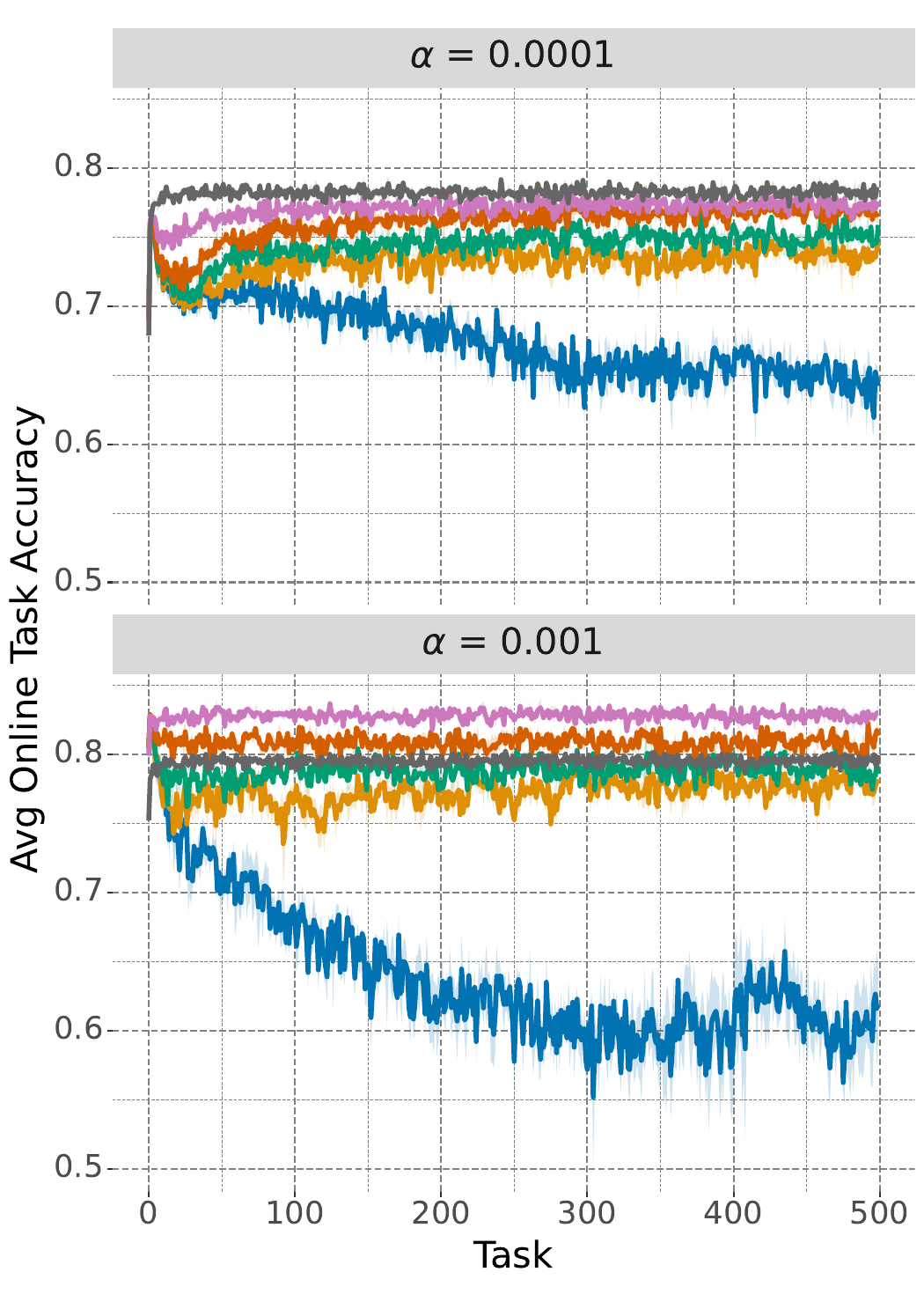}
                \vspace{-0.3cm} %
        \end{subfigure} &
        \begin{subfigure}{0.30\textwidth}
            \caption*{Random Label MNIST} %
            \includegraphics[width=\linewidth]{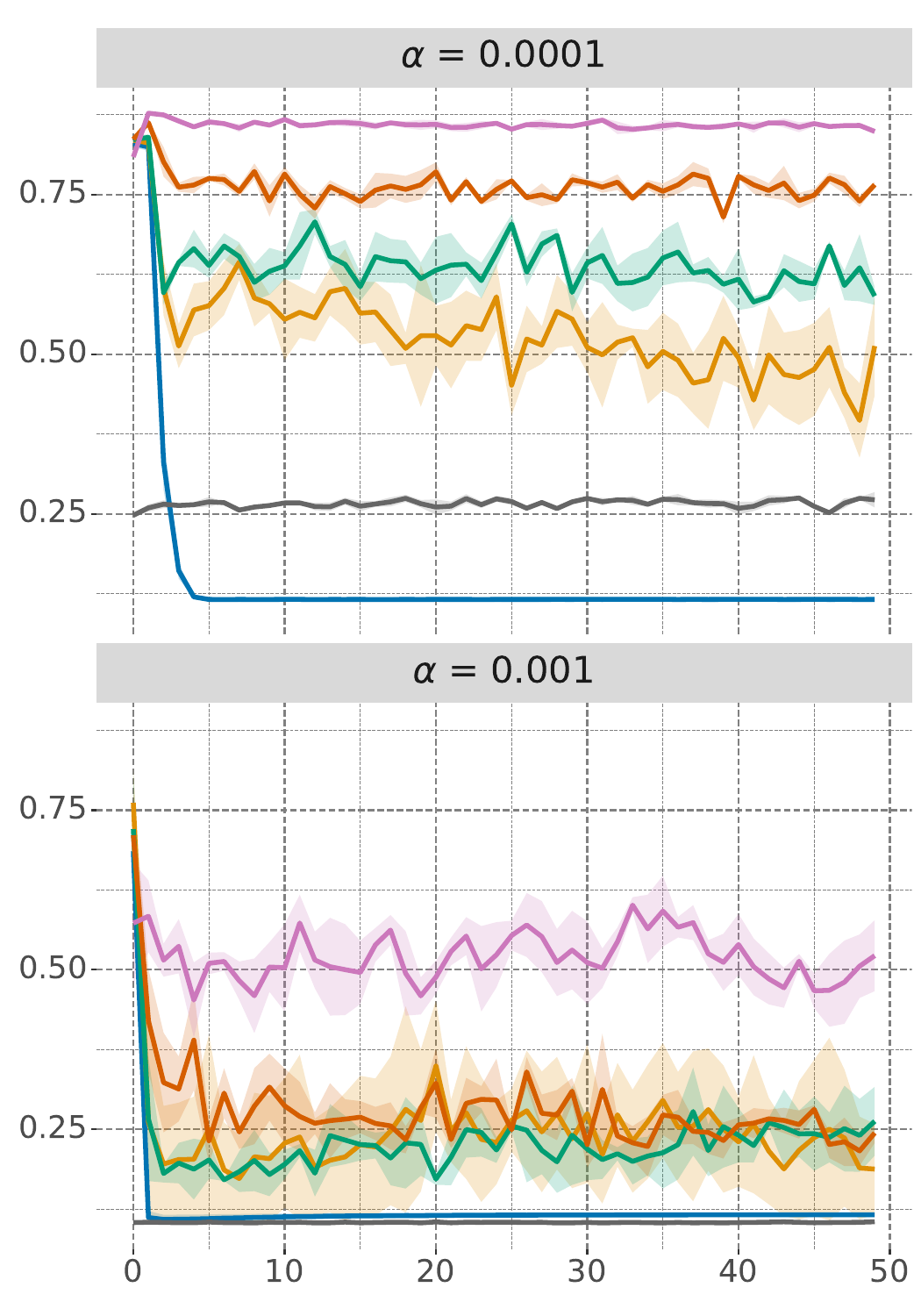}
                \vspace{-0.3cm} %
        \end{subfigure} &
        \begin{subfigure}{0.30\textwidth}
            \caption*{5+1 CIFAR} %
            \includegraphics[width=\linewidth]{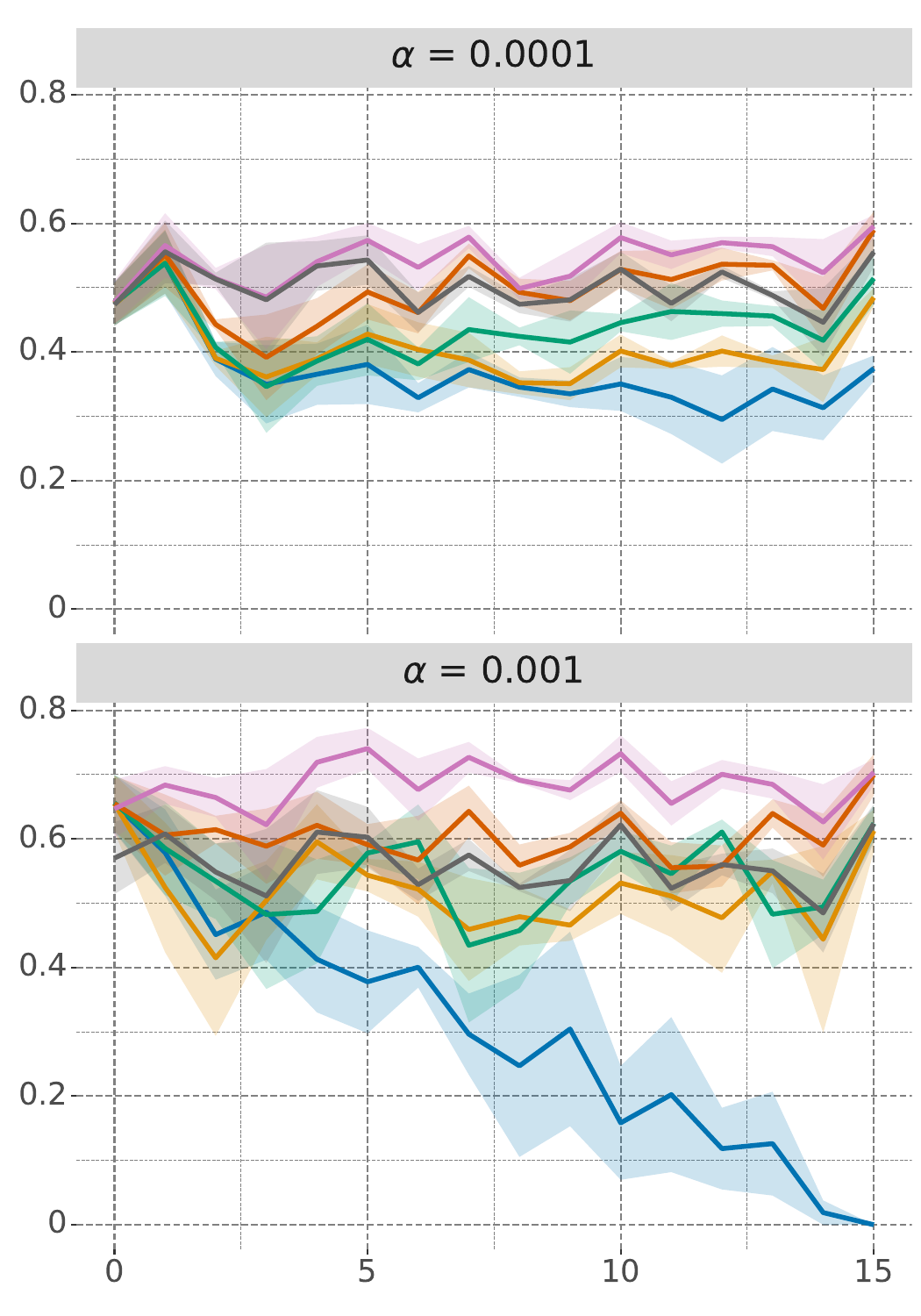}
                \vspace{-0.3cm} %
        \end{subfigure}
    \end{tabular}

\begin{tikzpicture}
    \def\firstRowY{0.5}
    \draw [lambda_zero, thick, line width=2pt] (-2,\firstRowY) -- (-1.5,\firstRowY);
    \node[anchor=west] at (-1.5,\firstRowY) {$\lambda = 0$};
    
    \draw [lambda_1e-5, thick, line width=2pt] (1,\firstRowY) -- (1.5,\firstRowY);
    \node[anchor=west] at (1.5,\firstRowY) {$\lambda = 1\mathrm{e}{-5}$};

    \draw [lambda_1e-4, thick, line width=2pt] (4,\firstRowY) -- (4.5,\firstRowY);
    \node[anchor=west] at (4.5,\firstRowY) {$\lambda = 1\mathrm{e}{-4}$};

    \draw [lambda_1e-3, thick, line width=2pt] (-2,0) -- (-1.5,0);
    \node[anchor=west] at (-1.5,0) {$\lambda = 1\mathrm{e}{-3}$};
    
    \draw [lambda_1e-2, thick, line width=2pt] (1,0) -- (1.5,0);
    \node[anchor=west] at (1.5,0) {$\lambda = 1\mathrm{e}{-2}$};
    
    \draw [lambda_1e-1, thick, line width=2pt] (4,0) -- (4.5,0); 
    \node[anchor=west] at (4.5,0) {$\lambda = 1\mathrm{e}{-1}$};

\end{tikzpicture}
    
    \caption{\rebuttal{Sensitivity to regularization strength. We vary $\lambda$ when using the Adam optimizer with two different learning rates, $\alpha = 0.0001$ and $\alpha = 0.001$. L2 Init is sensitive to $\lambda$ on each problem. However, a single value $\lambda = 0.01$ achieves the best performance across problems.}}
    \label{fig:lambda-sensitivity}
\end{figure*}

\rebuttal{To investigate how sensitive L2 Init is to regularization strength, we vary the regularization strength $\lambda \in {1\mathrm{e}{-5}, 1\mathrm{e}{-4}, 1\mathrm{e}{-3}, 1\mathrm{e}{-2}, 1\mathrm{e}{-1}}$ on a subset of problems. As shown in Figure~\ref{fig:lambda-sensitivity}, we find that L2 Init is sensitive to the choice of $\lambda$ on each problem. However, a single value, $\lambda = 1\mathrm{e}{-2}$, achieves the best performance across problems.}

\subsubsection{\rebuttal{Effect on Forgetting}}

\begin{table}[ht]
\centering
\begin{tabular}{lccc}
\hline
\textbf{Method} & \textbf{BWT} & \textbf{One-Step BWT} & \textbf{Total Online Avg Acc} \\
\hline
L2 Init, $\lambda = 1e-3$ & $-63.4\%$ & $-11.3\%$ & $\textbf{81.1\%}$ \\
EWC, $\beta = 10$ & $\textbf{-39.0\%}$ & $-7.8\%$ & $75.8\%$ \\
L2 Init + EWC, $\lambda = 1e-3, \beta = 10$ & $-51.1\%$ & $\textbf{-7.8\%}$ & $80.0\%$ \\
\hline
\end{tabular}
\caption{\rebuttal{Comparision of L2 Init and EWC on Permuted MNIST with $20$ tasks. We find that adding EWC to L2 Init significantly reduces forgetting while having little impact on plasticity. However, EWC on its own, while having relatively poor plasticity, has less forgetting than L2 Init + EWC. Adding L2 Init to EWC does increase forgetting although the One-Step BWT metric is not affected.}}
\label{table:forgetting_1}
\end{table}


\rebuttal{In this section, we study the interaction of L2 Init with Elastic Weight Consolidation (EWC), a regularization approach designed to mitigate forgetting on past tasks~\cite{kirkpatrick2017overcoming}. We run experiments on Permuted MNIST with $20$ tasks. To study both forgetting and plasiticy, we compute three metrics: backward transfer, one-step backwards transfer, and the total online average accuracy metric described in Section~\ref{sec:continual_supervised_learning}.}

\rebuttal{To measure forgetting, we use the backwards transfer (BWT) metric as computed in \citet{lopez2017gradient}: $$\text{BWT} = \frac{1}{K - 1} \sum_{m = 1}^{K - 1} A_{K, m} - A_{m, m}$$}

\rebuttal{where $m$ is a task index and $K$ is the total number of tasks. An agent is trained on tasks $1, 2, 3, ..., K$ in a sequence. $A_{m, m}$ is the accuracy on the $m$th task's test data immediately after training on task $m$. $A_{K, m}$ is the accuracy on the $m$th task's test data after training on the final task $K$.}


\rebuttal{To better understand the effect of L2 Init on forgetting, we additionally measure how much, on average, training on a task affects performance on the previous task. We call this  ``one step backwards transfer," and compute it as: $$\text{One Step BWT} = \frac{1}{K - 1} \sum_{m = 1}^{K - 1} A_{m+1, m} - A_{m, m}$$ For each task, this metric computes the change in performance on a task, due to training on the subsequent task.}

\rebuttal{BTW can be less informative when there is significant plasticity loss. For instance, a method that fails to learn anything on a task will also not forget anything on that task, achieving a higher BTW score relative to methods which learn and forget. Therefore, we also compute the total online average accuracy metric so that we capture plasticity loss as well.}

\rebuttal{For our experiments, we use learning rate $\alpha = 1e-3$ with the Adam optimizer. We do not perform any hyper-paramter tuning: for L2 Init, we use $\lambda = 1e-3$, and for EWC we use the regularization strength $\beta = 10$ for the EWC regularization term. We average results over $5$ seeds. In addition to evaluating EWC and L2 Init in isolation, we also evaluate combining the two by adding both regularization terms. We call the resulting method L2 Init + EWC. Our results and their interpretation are in Table~\ref{table:forgetting_1}.}



\subsubsection{\rebuttal{Robustness to Batch Size}}

\renewcommand\thesubfigure{\hspace{0.5cm}(\alph{subfigure})}
\captionsetup[subfigure]{labelformat=simple, labelsep=none}
\begin{figure*}

    \centering
        
        \begin{subfigure}{0.32\textwidth}
            \caption*{Batch Size = 1} %
            \includegraphics[width=\linewidth]{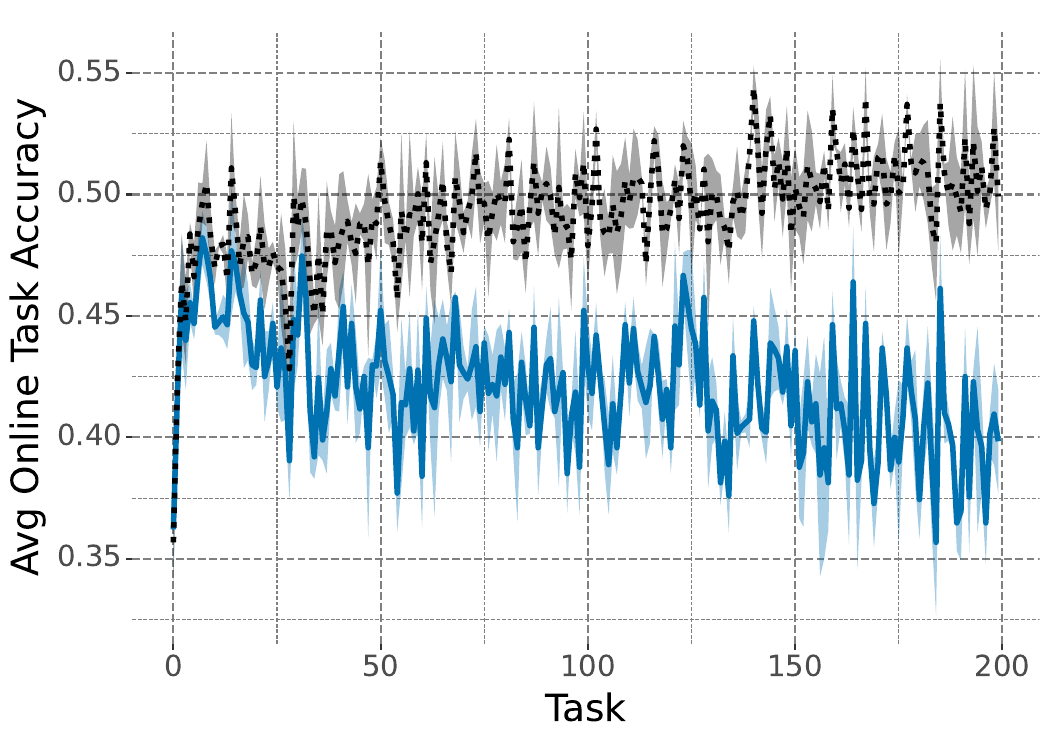}
            \vspace{-0.5cm} %
        \end{subfigure} 
        \begin{subfigure}{0.32\textwidth}
            \caption*{Batch Size = 32} %
            \includegraphics[width=\linewidth]{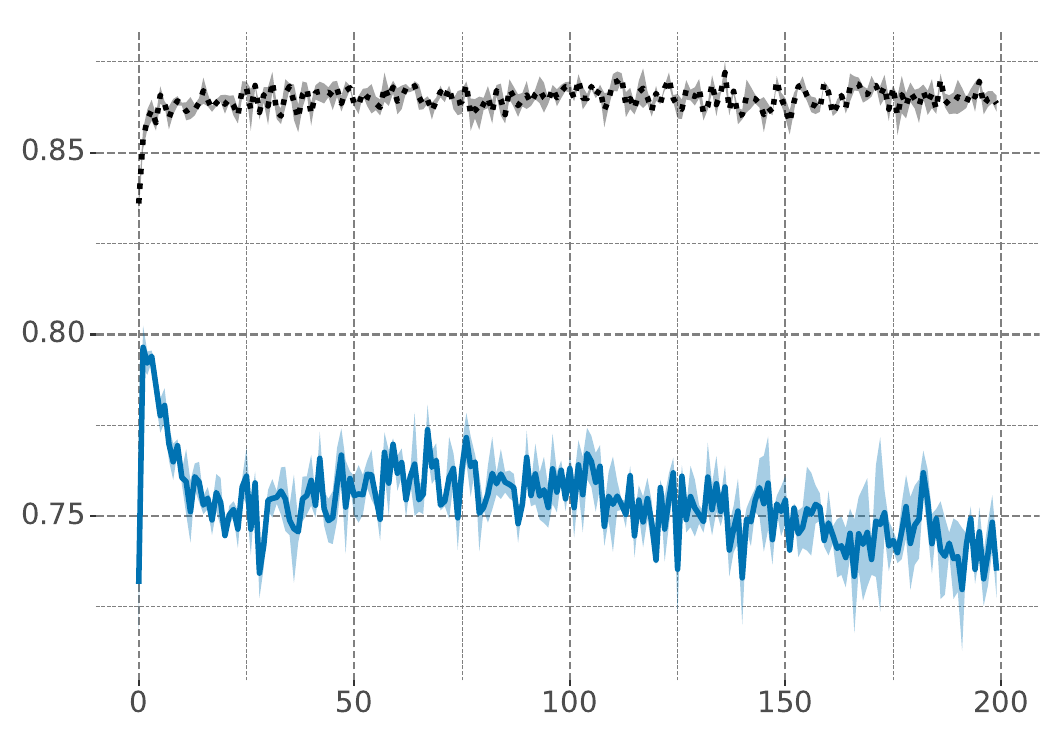}
            \vspace{-0.1cm} %
        \end{subfigure} 
        \begin{subfigure}{0.32\textwidth}
            \caption*{Batch Size = 128} %
            \includegraphics[width=\linewidth]{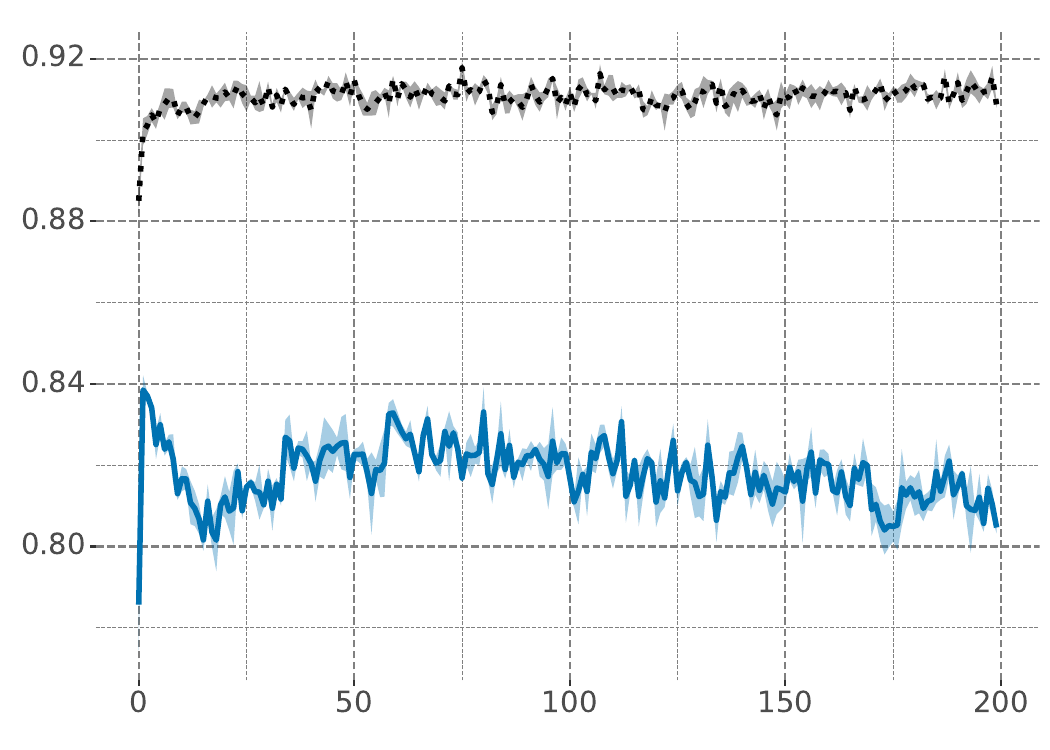}
            \vspace{-0.1cm} %
        \end{subfigure}

\begin{tikzpicture}
    \draw [Baseline, thick, line width=2pt] (-3,0) -- (-2.5,0);
    \node[anchor=west] at (-2.5,0) {Baseline};

    \draw [L2Init, thick, dotted, line width=2pt] (0,0) -- (0.5,0);
    \node[anchor=west] at (0.5,0) {L2 Init};
\end{tikzpicture}
    
    \caption{\rebuttal{Comparison of average online task accuracy on Permuted MNIST when using $3$ different batch sizes. We compare L2 Init with the Baseline (Adam) and find that across different batch sizes, L2 Init mitigates plasticity loss.}}
    \label{fig:robustness-to-batch-size}
\end{figure*}

\rebuttal{In this section, we investigate the robustness of L2 Init to different batch sizes. On Permuted MNIST, we compare of performance of L2 Init with the Baseline (Adam) when using different batch sizes. We sweep over learning rates $\alpha \in \{1e-5, 1e-4, 1e-3, 1e-2\}$. We do not perform a hyper-parameter sweep for L2 Init and use a fixed $\lambda = 0.01$ for all experiments. Our results in Figure~\ref{fig:robustness-to-batch-size} indicate that L2 Init mitigates plasticity loss regardless of batch size.}

\subsection{Connection to Shrink and Perturb}
\label{appendix:shrink-and-perturb-connection}
In \citet{ash2020warm}, the Shrink and Perturb method was proposed to mitigate loss of plasticity. Every time a task switches, Shrink and Perturb multiplies neural network parameters by a shrinkage factor $p < 1$ and then perturbs them by a small noise vector $\epsilon$. The Shrink and Perturb procedure is applied to the neural network when a task switches but can in principle be applied after every gradient step with a larger value of $p$. The update applied to the parameters $\theta_t$ at timestep $t$ is
\begin{align*}
    \theta_{t+1} = \underbrace{p}_{\text{Shrink}} (\underbrace{\theta_t - \alpha \nabla \mathcal{L}_{\text{train}}(\theta_t)}_{\text{SGD update}}) +  \underbrace{\sigma \epsilon}_{\text{Perturb}} 
\end{align*}
where $\epsilon$ is a noise vector and $\sigma$ is a scaling factor of the noise. 

\citet{ash2020warm} suggest sampling $\epsilon$ from the same distribution that the neural network parameters were sampled from at initialization and then scaling with $\sigma$ which is a hyperparameter. This is to ensure that the noise magnitude scales appropriately with the width and type of the neural network layer corresponding to each individual parameter. 

Before making the connection to our method, we will rewrite the Shrink and Perturb update rule further:
\begin{align*}
    \theta_{t+1} = \underbrace{p \theta_t}_{\text{Shrink}} + \underbrace{\sigma \epsilon}_{\text{Perturb}} - \alpha \underbrace{p}_{\text{Shrink}} \nabla \mathcal{L}_{\text{train}} (\theta_t)
\end{align*}
where we instead shrink both $\theta_t$ and shrink the gradient.

When using SGD with a constant stepsize $\alpha$, our method can be written on a form that is quite similar to this. Specifically, when applying \methodname, we can write the update to the parameters $\theta_t$ at timestep $t$ as

$$
\theta_{t+1} = \underbrace{(1 - \alpha \lambda)\theta_t}_{\text{Shrink}} + \underbrace{\alpha \lambda \theta_0}_{\text{Perturb}}  - \alpha \nabla \mathcal{L}_{\text{train}}(\theta_t)
$$   
where $\theta_0$ are the initial parameters at time step $0$, rather than random noise, and where the gradient is not shrunk. This form can be derived by taking the gradient of the \methodname augmented loss function, plugging it into the SGD update rule, and factoring out $\theta_t$. 

There are four seemingly small, but important, differences between \methodname and Shrink and Perturb. First, our method has only one hyperparameter $\lambda$ rather than two. That is because the shrinkage and noise scaling factors are tied to $\lambda$: $p = (1 - \alpha \lambda)$ and $\sigma = \alpha \lambda$. Further, both the shrinkage and noise scale parameters are tied to the step size. Second, our method regularizes toward the initial parameters, rather than toward a random sample from the initial distribution. Third, the gradient is not shrunk. Finally, when using Adam, the above connection between the two methods no longer holds for the same reason that L$2$ regularization and weight decay are not equivalent when using Adam.

\begin{table}
\centering
\caption{Agent optimal hyper-parameters on Permuted MNIST, Random Label MNIST, and Random Label CIFAR.}
\begin{tabular}{ |p{3.1cm}||p{1.6cm}||p{8.4cm}|  }
 \hline
 \multicolumn{3}{|c|}{Optimal Hyper-parameters on Permuted MNIST} \\
 \hline
 Agent & Optimizer & Optimal Hyper-parameters\\
 \hline
 Baseline  & SGD & $\alpha=1\mathrm{e}{-2}$ \\
 Layer Norm  & SGD & $\alpha=1\mathrm{e}{-2}$ \\
 \methodname & SGD & $\alpha=1\mathrm{e}{-2}$, $\lambda=1\mathrm{e}{-2}$ \\
 L2 & SGD & $\alpha=1\mathrm{e}{-2}$, $\lambda=1\mathrm{e}{-2}$ \\
 Shrink \& Perturb & SGD & $\alpha=1\mathrm{e}{-2}$, $p=1-1\mathrm{e}{-4}$, $\sigma=1\mathrm{e}{-2}$ \\
 Continual Backprop & SGD & $\alpha=1\mathrm{e}{-2}$, $r = 1\mathrm{e}{-3}$ \\
 Concat ReLU & SGD & $\alpha=1\mathrm{e}{-2}$ \\
 ReDO & SGD & $\alpha=1\mathrm{e}{-2}$, recycle period = 625, recycle threshold = 0 \\
 \hdashline
 Baseline  & Adam & $\alpha=1\mathrm{e}{-4}$ \\
 Layer Norm  & Adam & $\alpha=1\mathrm{e}{-3}$ \\
 \methodname & Adam & $\alpha=1\mathrm{e}{-3}$, $\lambda=1\mathrm{e}{-2}$ \\
 L2 Origin & Adam & $\alpha=1\mathrm{e}{-3}$, $\lambda=1\mathrm{e}{-2}$ \\
 Shrink \& Perturb & Adam & $\alpha=1\mathrm{e}{-3}$, $p=1-1\mathrm{e}{-3}$, $\sigma=1\mathrm{e}{-2}$ \\
 Continual Backprop & Adam & $\alpha=1\mathrm{e}{-3}$, $r=1\mathrm{e}{-3}$ \\
 Concat ReLU & Adam & $\alpha=1\mathrm{e}{-3}$ \\
 ReDO & Adam & $\alpha=1\mathrm{e}{-3}$, recycle period = 625, recycle threshold = 0  \\
 \hline
\end{tabular}
\\[10pt] %
\begin{tabular}{ |p{3.1cm}||p{1.6cm}||p{8.8cm}|  }
 \hline
 \multicolumn{3}{|c|}{Optimal Hyper-parameters on Random Label MNIST} \\
 \hline
 Agent & Optimizer & Optimal Hyper-parameters\\
 \hline
 Baseline  & SGD & $\alpha=1\mathrm{e}{-3}$ \\
 Layer Norm  & SGD & $\alpha=1\mathrm{e}{-3}$ \\
 \methodname & SGD & $\alpha=1\mathrm{e}{-2}$, $\lambda=1\mathrm{e}{-2}$ \\
 L2 & SGD & $\alpha=1\mathrm{e}{-2}$, $\lambda=1\mathrm{e}{-2}$ \\
 Shrink and Perturb & SGD & $\alpha=1\mathrm{e}{-2}$, $p=1-1\mathrm{e}-4$, $\sigma=1\mathrm{e}{-2}$ \\
 Continual Backprop & SGD & $\alpha=1\mathrm{e}{-2}$, $r = 1\mathrm{e}{-3}$ \\
 Concat ReLU & SGD & $\alpha=1\mathrm{e}{-2}$ \\
 ReDO & SGD & $\alpha=1\mathrm{e}{-2}$, recycle period = 30000, recycle threshold = 0.1 \\
 \hdashline
 Baseline  & Adam & $\alpha=1\mathrm{e}{-4}$ \\
 Layer Norm  & Adam & $\alpha=1\mathrm{e}{-4}$ \\
 \methodname & Adam & $\alpha=1\mathrm{e}{-4}$, $\lambda=1\mathrm{e}{-2}$ \\
 L2 & Adam & $\alpha=1\mathrm{e}{-4}$, $\lambda=1\mathrm{e}{-2}$ \\
 Shrink and Perturb & Adam & $\alpha=1\mathrm{e}{-4}$, $p=1-1\mathrm{e}{-4}$, $\sigma=1\mathrm{e}{-2}$ \\
 Continual Backprop & Adam & $\alpha=1\mathrm{e}{-3}$, $r = 1\mathrm{e}{-3}$ \\
 Concat ReLU & Adam & $\alpha=1\mathrm{e}{-3}$ \\
 ReDO & Adam & $\alpha=1\mathrm{e}{-3}$, recycle period = 30000, recycle threshold = 0.1 \\
 \hline
\end{tabular}
\\[10pt] %
\begin{tabular}{ |p{3.1cm}||p{1.6cm}||p{8.8cm}|  }
 \hline
 \multicolumn{3}{|c|}{Optimal Hyper-parameters on Random Label CIFAR} \\
 \hline
 Agent & Optimizer & Optimal Hyper-parameters\\
 \hline
 Baseline  & SGD & $\alpha=1\mathrm{e}{-2}$ \\
 Layer Norm  & SGD & $\alpha=1\mathrm{e}{-2}$ \\
 \methodname & SGD & $\alpha=1\mathrm{e}{-2}$, $\lambda=1\mathrm{e}{-2}$ \\
 L2 & SGD & $\alpha=1\mathrm{e}{-2}$, $\lambda=1\mathrm{e}{-2}$ \\
 Shrink \& Perturb & SGD & $\alpha=1\mathrm{e}{-2}$, $p=1-1\mathrm{e}{-4}$, $\sigma=1\mathrm{e}{-2}$ \\
 Continual Backprop & SGD & $\alpha=1\mathrm{e}{-2}$, $r = 1\mathrm{e}{-3}$ \\
 Concat ReLU & SGD & $\alpha=1\mathrm{e}{-3}$ \\
 ReDO & SGD & $\alpha=1\mathrm{e}{-2}$, recycle period = 30000, recycle threshold = 0.1 \\
 \hdashline
 Baseline  & Adam & $\alpha=1\mathrm{e}{-3}$ \\
 Layer Norm  & Adam & $\alpha=1\mathrm{e}{-3}$ \\
 \methodname & Adam & $\alpha=1\mathrm{e}{-3}$, $\lambda=1\mathrm{e}{-2}$ \\
 L2 & Adam & $\alpha=1\mathrm{e}{-4}$, $\lambda=1\mathrm{e}{-2}$ \\
 Shrink \& Perturb & Adam & $\alpha=1\mathrm{e}{-3}$, $p=1-1\mathrm{e}{-4}$, $\sigma=1\mathrm{e}{-2}$ \\
 Continual Backprop & Adam & $\alpha=1\mathrm{e}{-3}, r = 1\mathrm{e}{-4}$ \\
 Concat ReLU & Adam & $\alpha=1\mathrm{e}{-3}$ \\
 ReDO & Adam & $\alpha=1\mathrm{e}{-4}$, recycle period = 30000, recycle threshold = 0.1 \\
 \hline
\end{tabular}
\label{tab:agent-hyperparams1}
\end{table}

\begin{table}
\caption{Agent optimal hyper-parameters on 5+1 CIFAR and Continual ImageNet.}
\centering
\begin{tabular}{ |p{3.1cm}||p{1.6cm}||p{8.6cm}|  }
 \hline
 \multicolumn{3}{|c|}{Optimal Hyper-parameters on 5+1 CIFAR} \\
 \hline
 Agent & Optimizer & Optimal Hyper-parameters\\
 \hline
 Baseline  & SGD & $\alpha=1\mathrm{e}{-2}$ \\
 Layer Norm  & SGD & $\alpha=0.1$ \\
 \methodname & SGD & $\alpha=1\mathrm{e}{-2}$, $\lambda=1\mathrm{e}{-5}$ \\
 L2 & SGD & $\alpha=1\mathrm{e}{-2}$, $\lambda=1\mathrm{e}{-4}$ \\
 Shrink \& Perturb & SGD & $\alpha=1\mathrm{e}{-2}$, $p=1-1\mathrm{e}{-5}$, $\sigma=1\mathrm{e}{-2}$ \\
 Continual Backprop & SGD & $\alpha=1\mathrm{e}{-2}$, $r = 1\mathrm{e}{-3}$ \\
 Concat ReLU & SGD & $\alpha=1\mathrm{e}{-2}$ \\
 ReDO & SGD & $\alpha=1\mathrm{e}{-2}$, recycle period = 1560, recycle threshold = 0 \\
 \hdashline
 Baseline  & Adam & $\alpha=1\mathrm{e}{-4}$ \\
 Layer Norm  & Adam & $\alpha=1\mathrm{e}{-4}$ \\
 \methodname & Adam & $\alpha=1\mathrm{e}{-3}$, $\lambda=1\mathrm{e}{-2}$ \\
 L2 Origin & Adam & $\alpha=1\mathrm{e}{-3}$, $\lambda=1\mathrm{e}{-3}$ \\
 Shrink \& Perturb & Adam & $\alpha=1\mathrm{e}{-3}$, $p=1-1\mathrm{e}{-4}$, $\sigma=1\mathrm{e}{-2}$ \\
 Continual Backprop & Adam & $\alpha=1\mathrm{e}{-3}$, $r=1\mathrm{e}{-4}$ \\
 Concat ReLU & Adam & $\alpha=1\mathrm{e}{-3}$ \\
 ReDO & Adam & $\alpha=1\mathrm{e}{-3}$, recycle period = 1560, recycle threshold = 0 \\
 \hline
\end{tabular}
\\[10pt] %
\begin{tabular}{ |p{3.1cm}||p{1.6cm}||p{8.6cm}|  }
 \hline
 \multicolumn{3}{|c|}{Optimal Hyper-parameters on Continual ImageNet} \\
 \hline
 Agent & Optimizer & Optimal Hyper-parameters\\
 \hline
 Baseline  & SGD & $\alpha=0.1$ \\
 Layer Norm  & SGD & $\alpha=0.1$ \\
 \methodname & SGD & $0.1$, $\lambda=1\mathrm{e}{-3}$ \\
 L2 & SGD & $0.1$, $\lambda=1\mathrm{e}{-3}$ \\
 Shrink and Perturb & SGD & $\alpha=0.1$, $p=1-1\mathrm{e}{-4}$, $\sigma=1\mathrm{e}{-4}$ \\
 Continual Backprop & SGD & $\alpha=0.1$, \rebuttal{$r = 1\mathrm{e}{-3}$} \\
 Concat ReLU & SGD & $\alpha=1\mathrm{e}{-2}$ \\
 ReDO & SGD & $\alpha=0.1$, recycle period = 600, recycle threshold = 0.1 \\
 \hdashline
 Baseline  & Adam & $\alpha=1\mathrm{e}{-4}$ \\
 Layer Norm  & Adam & $\alpha=1\mathrm{e}{-3}$ \\
 \methodname & Adam & $\alpha=1\mathrm{e}{-3}$, $\lambda=1\mathrm{e}{-3}$ \\
 L2 & Adam & $\alpha=1\mathrm{e}{-3}$, $\lambda=1\mathrm{e}{-3}$ \\
 Shrink and Perturb & Adam & $\alpha=1\mathrm{e}{-3}$, $p=1-1\mathrm{e}{-4}$, $\sigma=1\mathrm{e}{-2}$ \\
 Continual Backprop & Adam & $\alpha=1\mathrm{e}{-3}, r = 1\mathrm{e}{-4}$ \\
 Concat ReLU & Adam & $\alpha=1\mathrm{e}{-3}$ \\
 ReDO & Adam & $\alpha=1\mathrm{e}{-3}$, recycle period = 120, recycle threshold = 0 \\
 \hline
\end{tabular}
\\[10pt] %
\label{tab:agent-hyperparams2}
\end{table}

\end{document}